\PassOptionsToPackage{numbers, compress}{natbib}
\documentclass[]{antgroup}
\usepackage{antgroup}
\newif\ifshowrevisions
\showrevisionsfalse
\DeclareRobustCommand{\revised}[1]{%
  \ifshowrevisions{\textcolor{red}{#1}}\else{#1}\fi}

\newcommand{\methodname}{UniSpark}
\newcommand{\modelscale}{9B}
\newcommand{\methoddescription}{Exploring Hybrid Attention for Multimodal Understanding and Generation}
\newcommand{\modelname}{\methodname-\modelscale}
\newcommand{\papertitle}{\methodname: \methoddescription}

\usepackage{graphicx}
\usepackage{xspace}
\usepackage{tikz}
\usepackage{todonotes}
\usepackage{multirow}
\usepackage{amsmath}
\usepackage{cleveref}
\usepackage{subcaption}
\usepackage{multicol}
\usepackage{amssymb}
\usepackage{booktabs}
\usepackage{array}
\usepackage{bm}
\usepackage{enumitem}
\usepackage{CJKutf8}
\usepackage{algorithm}
\usepackage{algpseudocode}
\usepackage{tabularx}
\usepackage{bbm}
\usepackage{makecell}
\usepackage{threeparttable}
\usepackage{float}
\usepackage{wrapfig}
\usepackage[normalem]{ulem}
\useunder{\uline}{\ul}{}

\usepackage{amsmath,amsfonts,bm}

\def\figref#1{figure~\ref{#1}}

\def\secref#1{section~\ref{#1}}

\def\eqref#1{equation~\ref{#1}}

\def\algref#1{algorithm~\ref{#1}}

\def\1{\bm{1}}

\DeclareMathAlphabet{\mathsfit}{\encodingdefault}{\sfdefault}{m}{sl}
\SetMathAlphabet{\mathsfit}{bold}{\encodingdefault}{\sfdefault}{bx}{n}

\newcommand*\justify{%
  \fontdimen2\font=0.4em%
  \fontdimen3\font=0.2em%
  \fontdimen4\font=0.1em%
  \fontdimen7\font=0.1em%
  \hyphenchar\font=`\-%
}

\renewcommand{\texttt}[1]{%
  \begingroup
  \ttfamily
  \begingroup\lccode`~=`/\lowercase{\endgroup\def~}{/\discretionary{}{}{}}%
  \begingroup\lccode`~=`[\lowercase{\endgroup\def~}{[\discretionary{}{}{}}%
  \begingroup\lccode`~=`.\lowercase{\endgroup\def~}{.\discretionary{}{}{}}%
  \catcode`/=\active\catcode`[=\active\catcode`.=\active
  \justify\scantokens{#1\noexpand}%
  \endgroup
}

\usepackage{amsmath}
\usepackage{amssymb}
\usepackage{amsfonts}                               %
\usepackage{amsthm}
\usepackage[mathcal]{eucal}
\usepackage{mathrsfs}
\usepackage{bm}                                     %
\usepackage{blkarray}                               %
\usepackage{nicefrac}                               %

\usepackage{wrapfig}
\usepackage{graphicx}                               %
\usepackage{caption}
\usepackage{cleveref}

\usepackage{tikz}                                          
\usepackage{circuitikz}
\usetikzlibrary{patterns}
\usetikzlibrary{positioning,calc,fit,decorations.pathmorphing,shapes.geometric, shapes.gates.logic.US, calc}
\usetikzlibrary{arrows,arrows.meta,decorations.markings,shapes,shapes.arrows}
\usetikzlibrary{decorations,decorations.pathreplacing}
\usetikzlibrary{backgrounds}
\usepackage{pgfplots}
\usepackage{pgfplotstable}
\usepgfplotslibrary{groupplots}
\usepackage{scalefnt}
\pgfplotsset{compat=newest}

\makeatletter
\patchcmd{\@addtotoporbot}
  {\@midlist\@botlist}
  {\@toplist\@midlist\@botlist}
  {}{\PackageError{manuscript}{Cannot enforce one top float per type}{}}
\makeatother

\definecolor{firstcolor}{HTML}{C3423F}
\definecolor{secondcolor}{HTML}{2A4B8C}
\definecolor{myblue}{HTML}{D9E6FF}
\definecolor{statusgreen}{HTML}{26734D}
\definecolor{statusorange}{HTML}{CC7000}

\usepackage{tocloft}
\newenvironment{talign*}
  {\csname align*\endcsname}
  {\endalign}

\definecolor{darkred}{RGB}{150,0,0}
\providecommand{\darkredtext}[1]{\textcolor{darkred}{#1}}

\renewcommand{\eqref}[1]{%
  \hyperref[#1]{\darkredtext{\textup{Eq.~(\ref*{#1})}}}}
\renewcommand{\secref}[1]{%
  \hyperref[#1]{\darkredtext{Sec.~\ref*{#1}}}}
\renewcommand{\figref}[1]{%
  \hyperref[#1]{\darkredtext{Fig.~\ref*{#1}}}}
\renewcommand{\algref}[1]{%
  \hyperref[#1]{\darkredtext{Alg.~\ref*{#1}}}}

\providecommand{\thmref}[1]{\hyperref[#1]{\darkredtext{Thm.~\ref*{#1}}}}
\providecommand{\defref}[1]{\hyperref[#1]{\darkredtext{Def.~\ref*{#1}}}}
\providecommand{\propref}[1]{\hyperref[#1]{\darkredtext{Prop.~\ref*{#1}}}}
\providecommand{\assumpref}[1]{\hyperref[#1]{\darkredtext{Assump.~\ref*{#1}}}}
\providecommand{\remarkref}[1]{\hyperref[#1]{\darkredtext{Rem.~\ref*{#1}}}}
\providecommand{\hypref}[1]{\hyperref[#1]{\darkredtext{Hyp.~\ref*{#1}}}}
\providecommand{\conjref}[1]{\hyperref[#1]{\darkredtext{Conj.~\ref*{#1}}}}
\providecommand{\lemref}[1]{\hyperref[#1]{\darkredtext{Lem.~\ref*{#1}}}}
\providecommand{\corref}[1]{\hyperref[#1]{\darkredtext{Cor.~\ref*{#1}}}}
\providecommand{\noteref}[1]{\hyperref[#1]{\darkredtext{Nota.~\ref*{#1}}}}
\providecommand{\claimref}[1]{\hyperref[#1]{\darkredtext{Clm.~\ref*{#1}}}}
\providecommand{\obsref}[1]{\hyperref[#1]{\darkredtext{Obs.~\ref*{#1}}}}
\providecommand{\tabref}[1]{%
  \hyperref[#1]{\darkredtext{Tab.~\ref*{#1}}}}
\providecommand{\appref}[1]{%
  \hyperref[#1]{\darkredtext{App.~\ref*{#1}}}}

\theoremstyle{plain}

\theoremstyle{definition}

\theoremstyle{remark}

\tcbset{
  llada theorem box/.style={
    enhanced,
    breakable,
    sharp corners,
    boxrule=0pt,
    leftrule=2.5pt,
    colback=title_blue!4!white,
    colframe=title_blue,
    left=6pt,
    right=6pt,
    top=3pt,
    bottom=3pt,
    before skip=7pt,
    after skip=7pt
  },
  llada remark box/.style={
    llada theorem box,
    colback=statusorange!5!white,
    colframe=statusorange
  }
}

\tcolorboxenvironment{theorem}{llada theorem box}
\tcolorboxenvironment{proposition}{llada theorem box}
\tcolorboxenvironment{lemma}{llada theorem box}
\tcolorboxenvironment{corollary}{llada theorem box}
\tcolorboxenvironment{conjecture}{llada theorem box}
\tcolorboxenvironment{definition}{llada theorem box}
\tcolorboxenvironment{assumption}{llada theorem box}
\tcolorboxenvironment{hypothesis}{llada theorem box}
\tcolorboxenvironment{notation}{llada theorem box}
\tcolorboxenvironment{claim}{llada theorem box}
\tcolorboxenvironment{problem}{llada theorem box}
\tcolorboxenvironment{observation}{llada theorem box}
\tcolorboxenvironment{remark}{llada remark box}

\title{\papertitle}

\author{Jiancheng Pan$^{*}$, Yuzheng Liu$^{*}$, Hengyuan Cao$^{*}$, Qiyuan Zhang$^{*}$,\\
Yangyan Li, Biao Gong$^{\dagger}$, Zheng Zhang, Ka Leong Cheng, Xinyang Wang, Yujun Shen}
\contribution[*]{Equal Contribution\qquad}
\contribution[\dagger]{Project Lead}

\begingroup
\endgroup

\begin{document}
\maketitle

\begin{abstract}

We present \textbf{\modelname{}}, a unified multimodal model that investigates hybrid attention for visual understanding and generation. Built on a Mixture-of-Transformer-Experts backbone with approximately 9B parameters in each branch, \methodname{} combines full attention for explicit token interactions with GatedDeltaNet linear attention for compact context propagation. Autoregressive text prediction and visual flow matching jointly support visual question answering, text-to-image generation, instruction-guided editing, and interleaved text–image prediction. In interleaved generation, noisy visual targets are isolated from subsequent predictions, while clean visual representations provide context for later outputs. A blockwise protocol enforces this distinction through attention masks in full-attention layers and recurrent-state retention rules in linear-attention layers. 
The selected training sources contain 159.7M generation samples, 213.4M understanding samples, and 624.8K interleaved sequences. A four-stage curriculum jointly trains multimodal understanding and
visual generation, refining answer accuracy and instruction following
while extending image synthesis to 2K.
\methodname{} scores 2470.3 on MME-S and 0.83 on GenEval. Qualitative examples further illustrate its capabilities in interleaved image generation, video synthesis, and video editing. 
Together, these results demonstrate that hybrid attention provides an effective backbone for unified multimodal understanding and generation.

\par\vspace{6pt}
\begingroup
\normalsize
\definecolor{unisparkpink}{HTML}{FF3158}
\noindent\makebox[\linewidth][c]{%
  \begin{tabular}{
    @{}m{12pt}@{\hspace{6pt}}
    m{64pt}@{\hspace{10pt}}l@{}
  }
    \tikz[x=0.25pt,y=-0.25pt]{%
      \path[use as bounding box] (0,0) rectangle (48,48);
      \fill[unisparkpink]
        (24,2)--(29,18)--(46,24)--(29,30)--
        (24,46)--(18,30)--(2,24)--(18,18)--cycle;
      \fill[white]
        (24,15)--(27,21)--(33,24)--(27,27)--
        (24,33)--(21,27)--(15,24)--(21,21)--cycle;
    }
    & \textbf{Project Page}
    & \href{https://lucaria-academy.github.io/UniSpark/}{%
        \nolinkurl{https://lucaria-academy.github.io/UniSpark/}}
  \end{tabular}%
}
\endgroup
\end{abstract}

\par\vspace{1mm}
\begin{center}
\includegraphics[width=\linewidth,height=7cm,keepaspectratio]{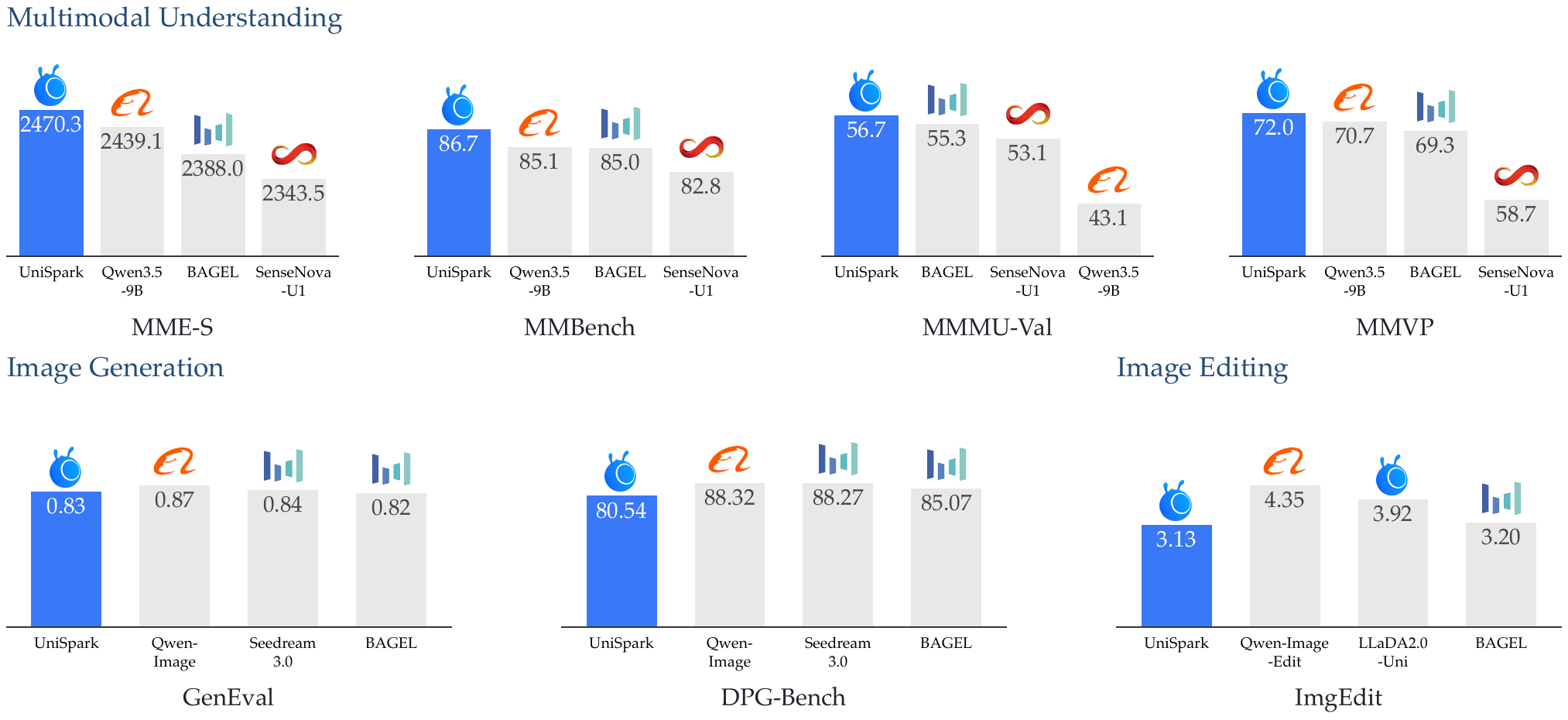}
\captionof{figure}{\textbf{Multimodal understanding, generation, and editing performance.}}
\label{fig:benchmark-preview}
\end{center}

\newgeometry{top=1.05cm,bottom=2.5cm,left=1.2cm,right=1.2cm,footskip=30pt}
\begingroup
\setlength{\parskip}{0pt}\setlength{\parindent}{0pt}
\captionsetup{skip=6pt}
\newcommand{\showcaseheading}[3]{\phantomsection\label{#1}\thispagestyle{plain}}

\newsavebox{\generationmosaic}
\showcaseheading{vis:t2i-1}{Multimodal Generation}{}
\sbox{\generationmosaic}{\vbox{\offinterlineskip\hbox{\begin{minipage}[t]{.33333333\linewidth}\offinterlineskip
 \includegraphics[width=\linewidth]{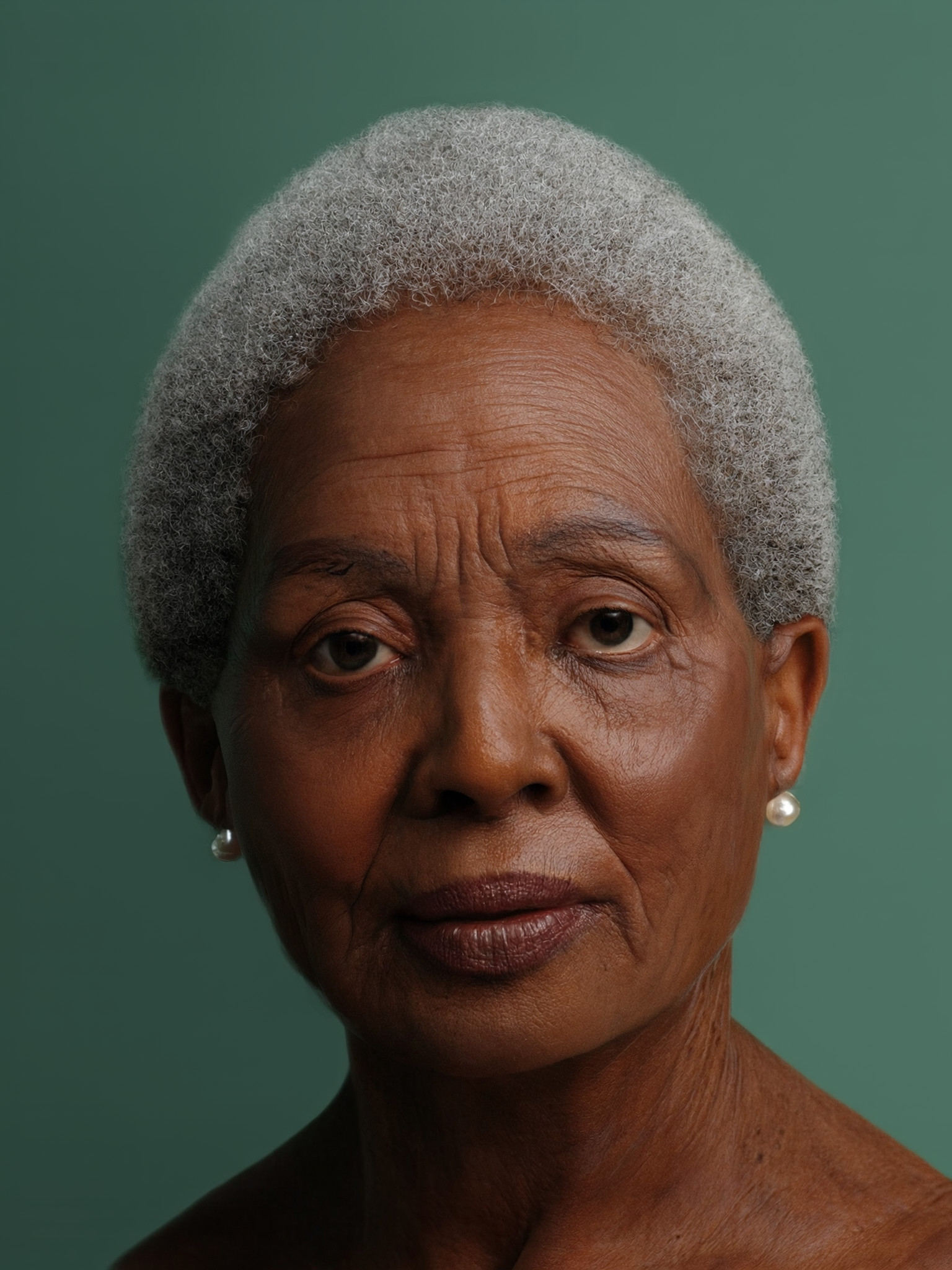}\par
 \includegraphics[width=\linewidth]{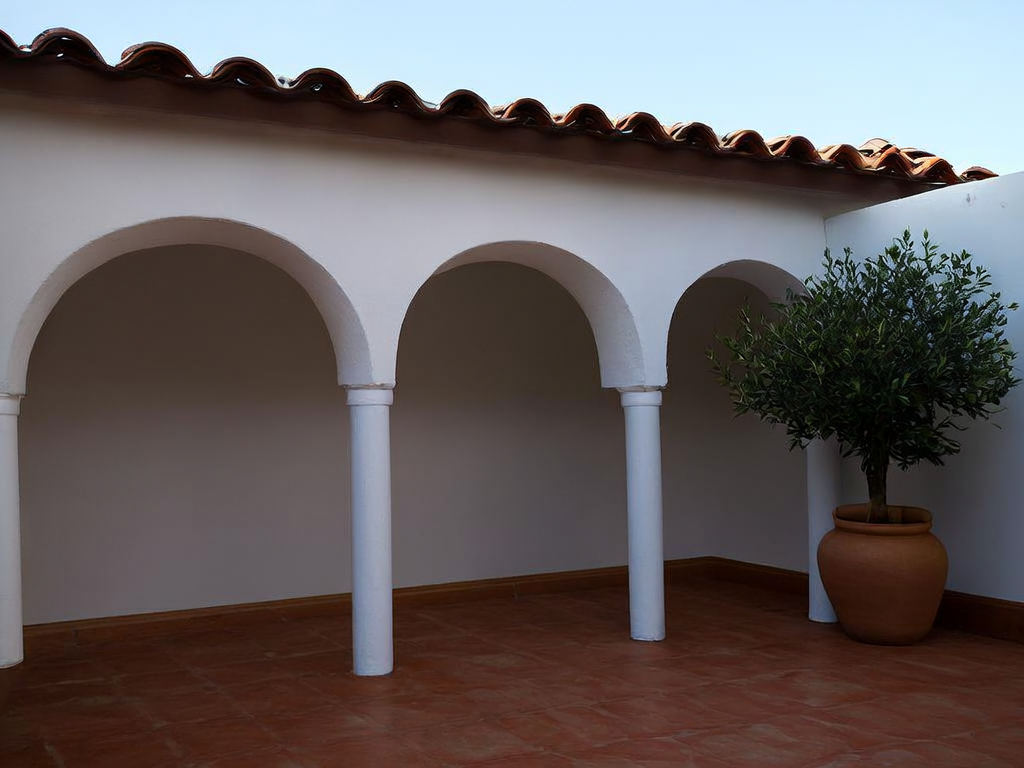}\par
 \includegraphics[width=\linewidth]{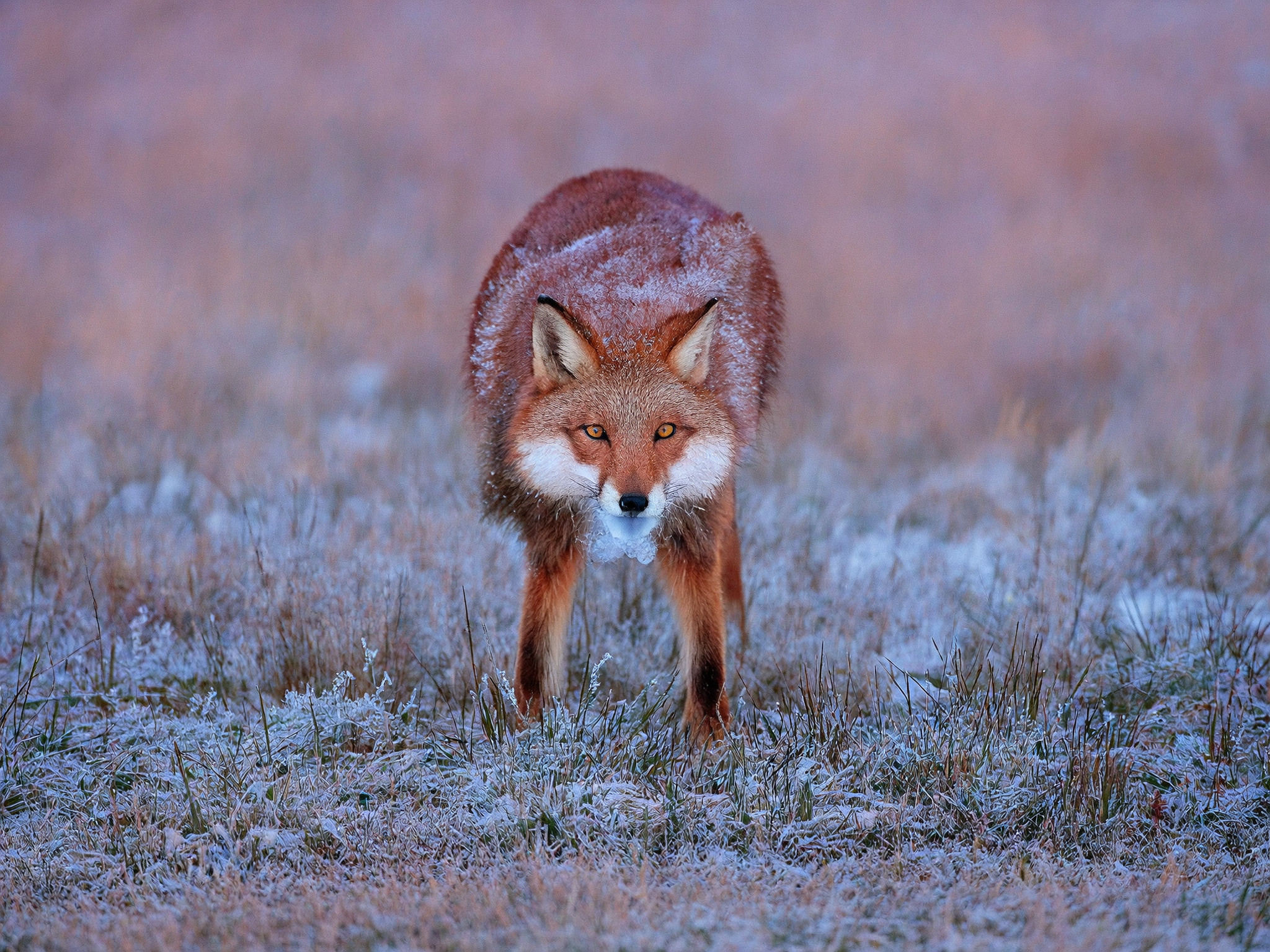}\par
 \includegraphics[width=\linewidth]{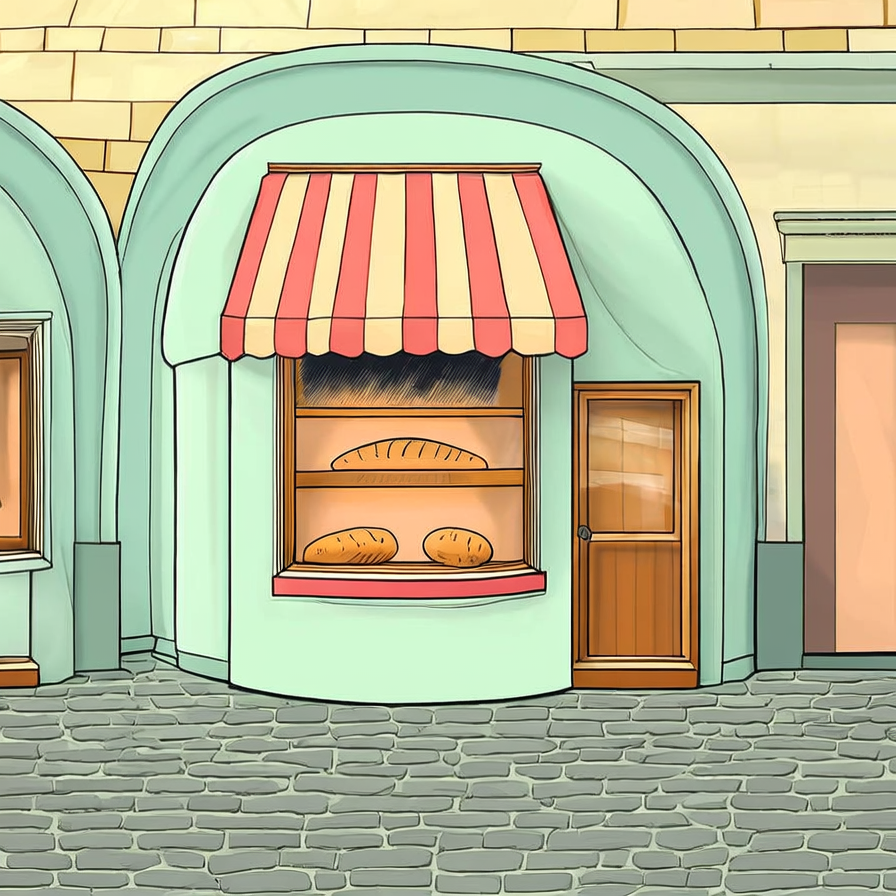}\par
 \end{minipage}\begin{minipage}[t]{.33333333\linewidth}\offinterlineskip
 \includegraphics[width=\linewidth]{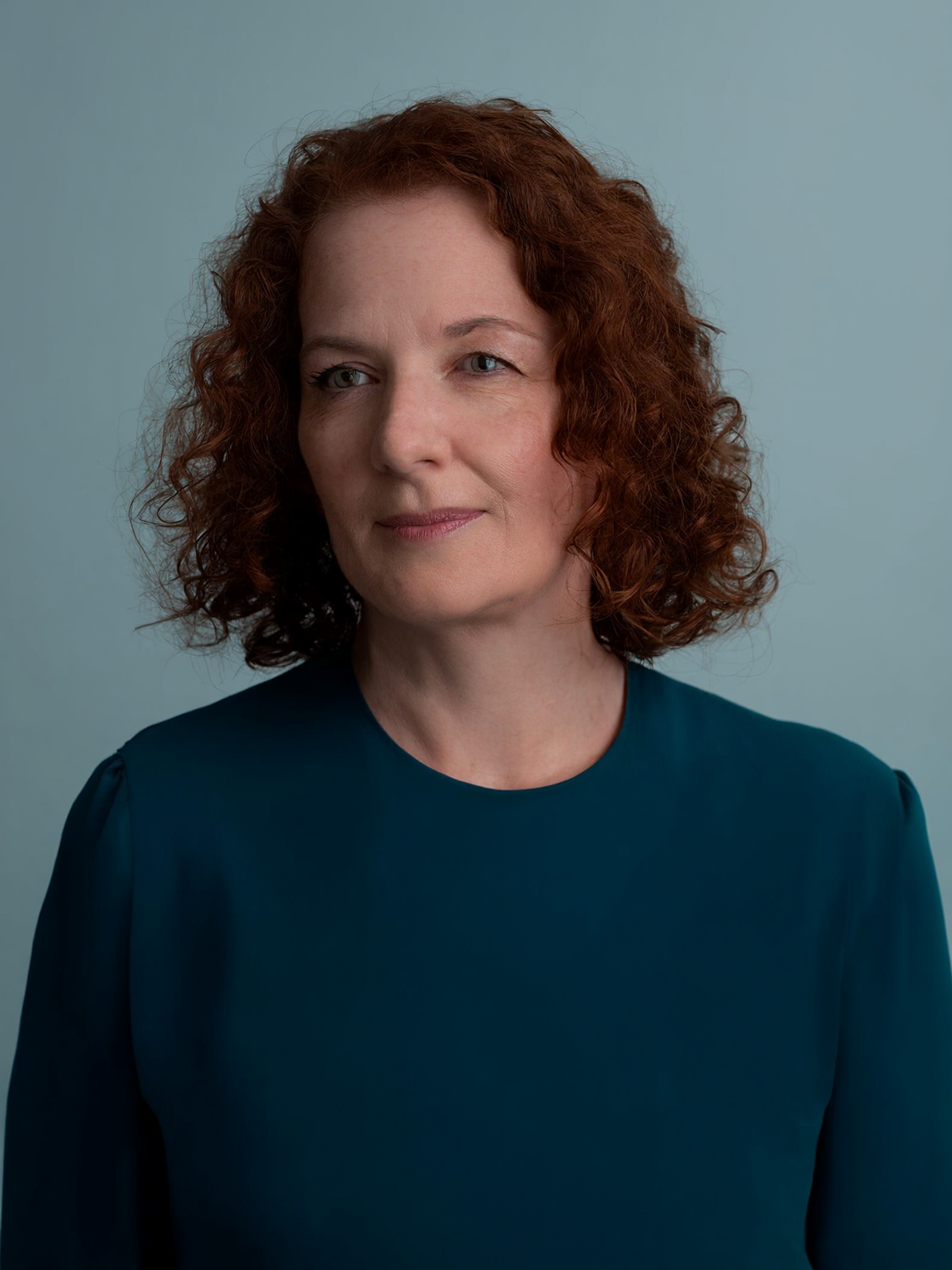}\par
 \includegraphics[width=\linewidth]{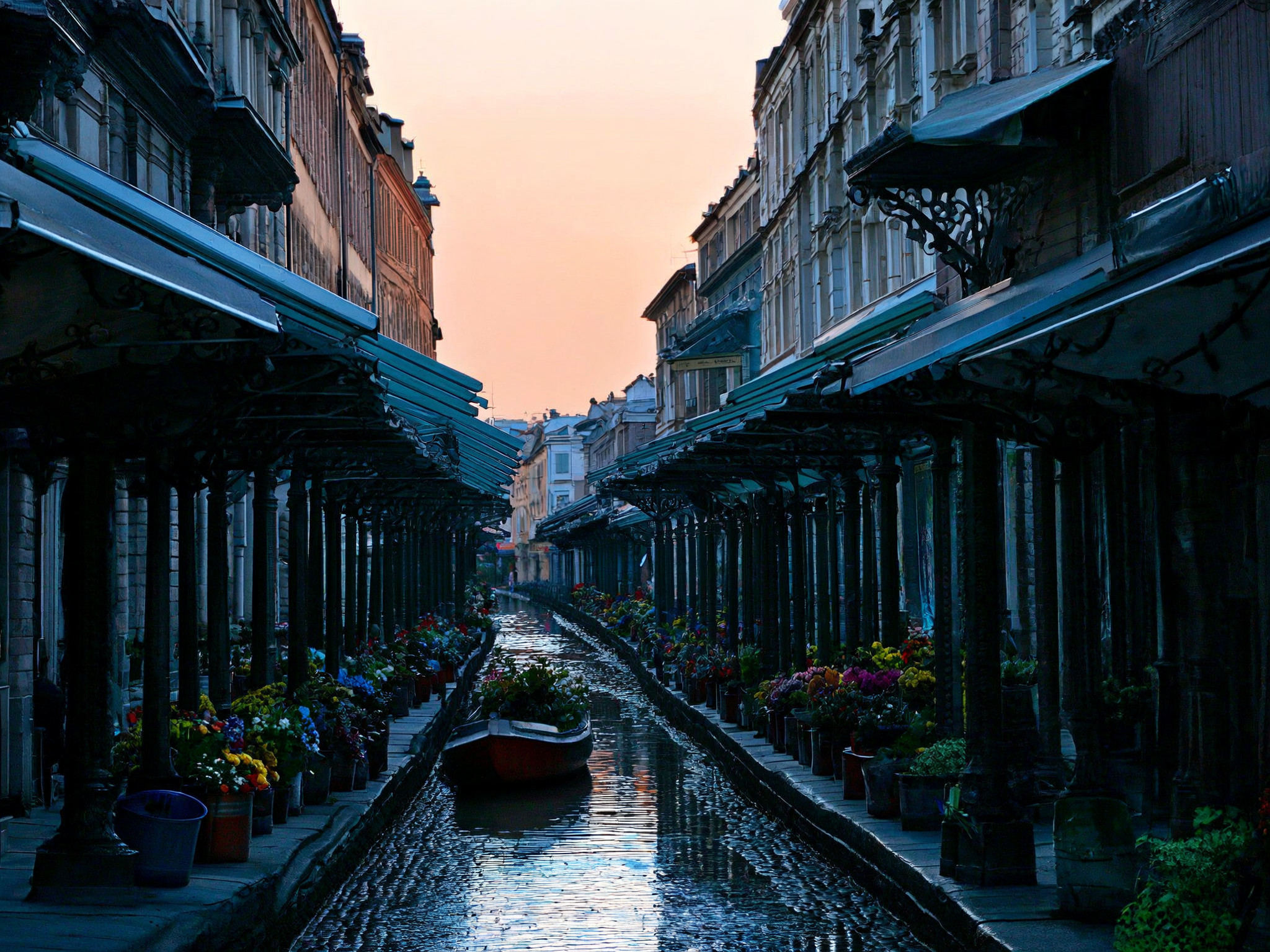}\par
 \includegraphics[width=\linewidth]{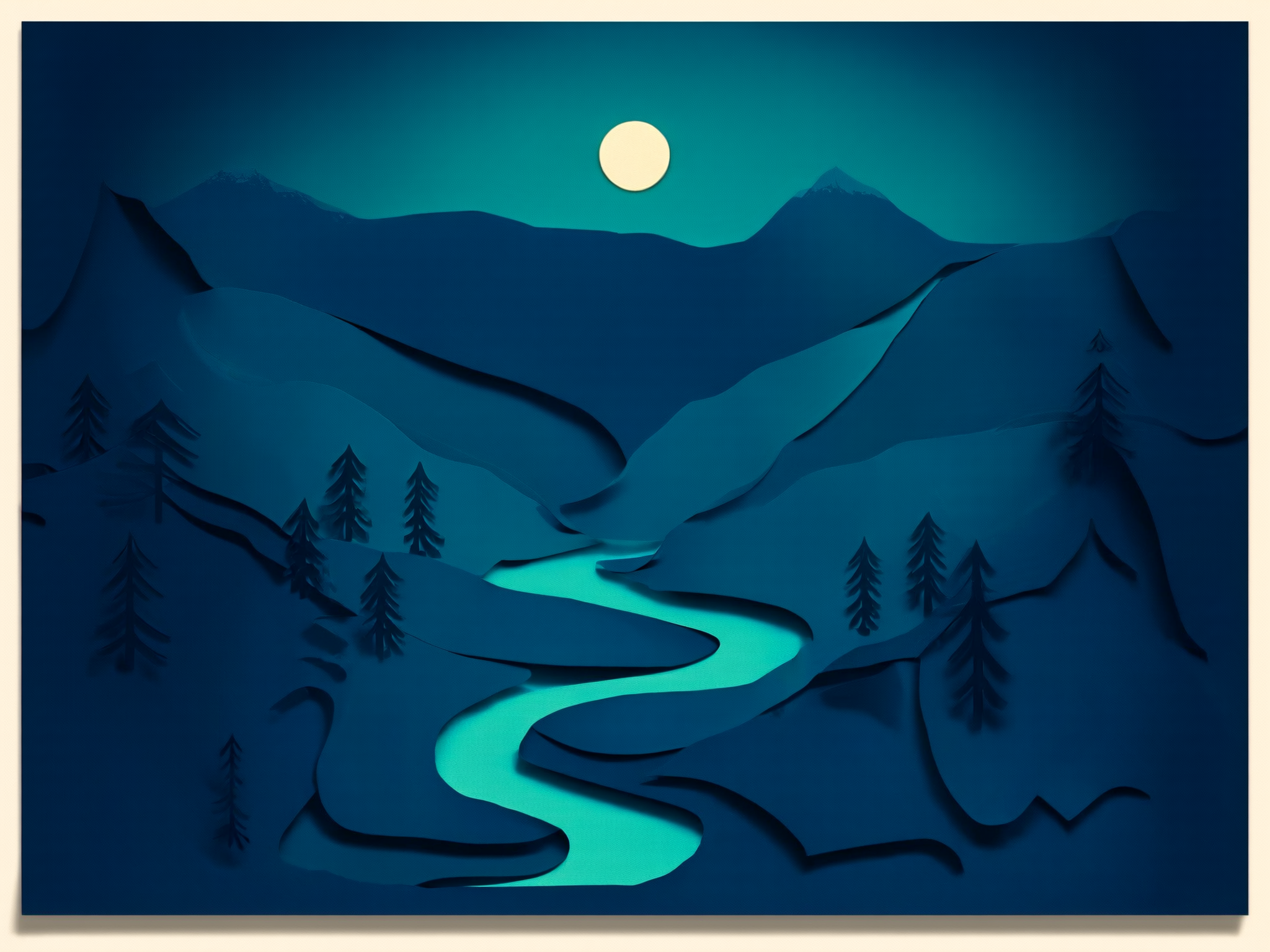}\par
 \includegraphics[width=\linewidth]{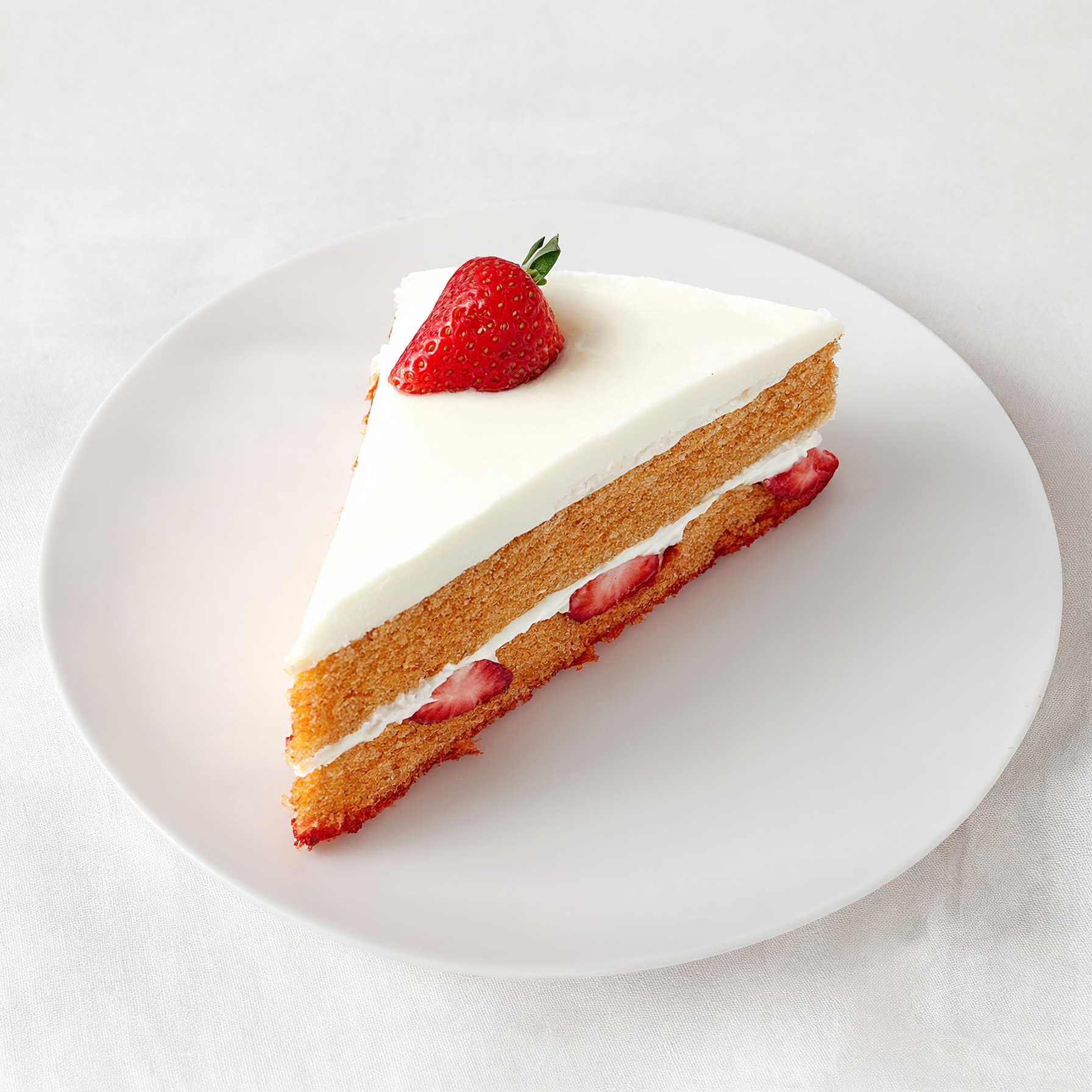}\par
 \end{minipage}\begin{minipage}[t]{.33333333\linewidth}\offinterlineskip
 \includegraphics[width=\linewidth]{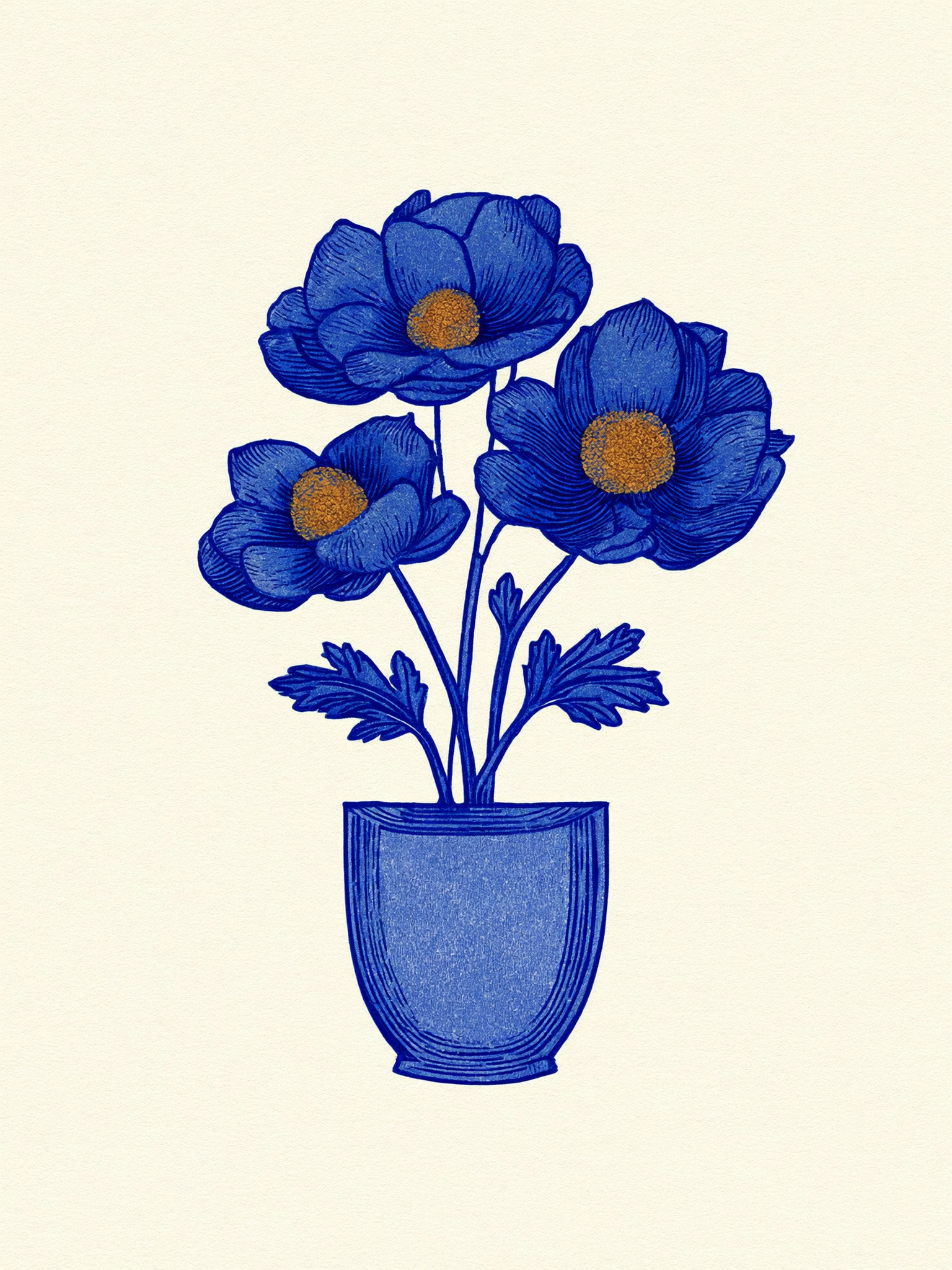}\par
 \includegraphics[width=\linewidth]{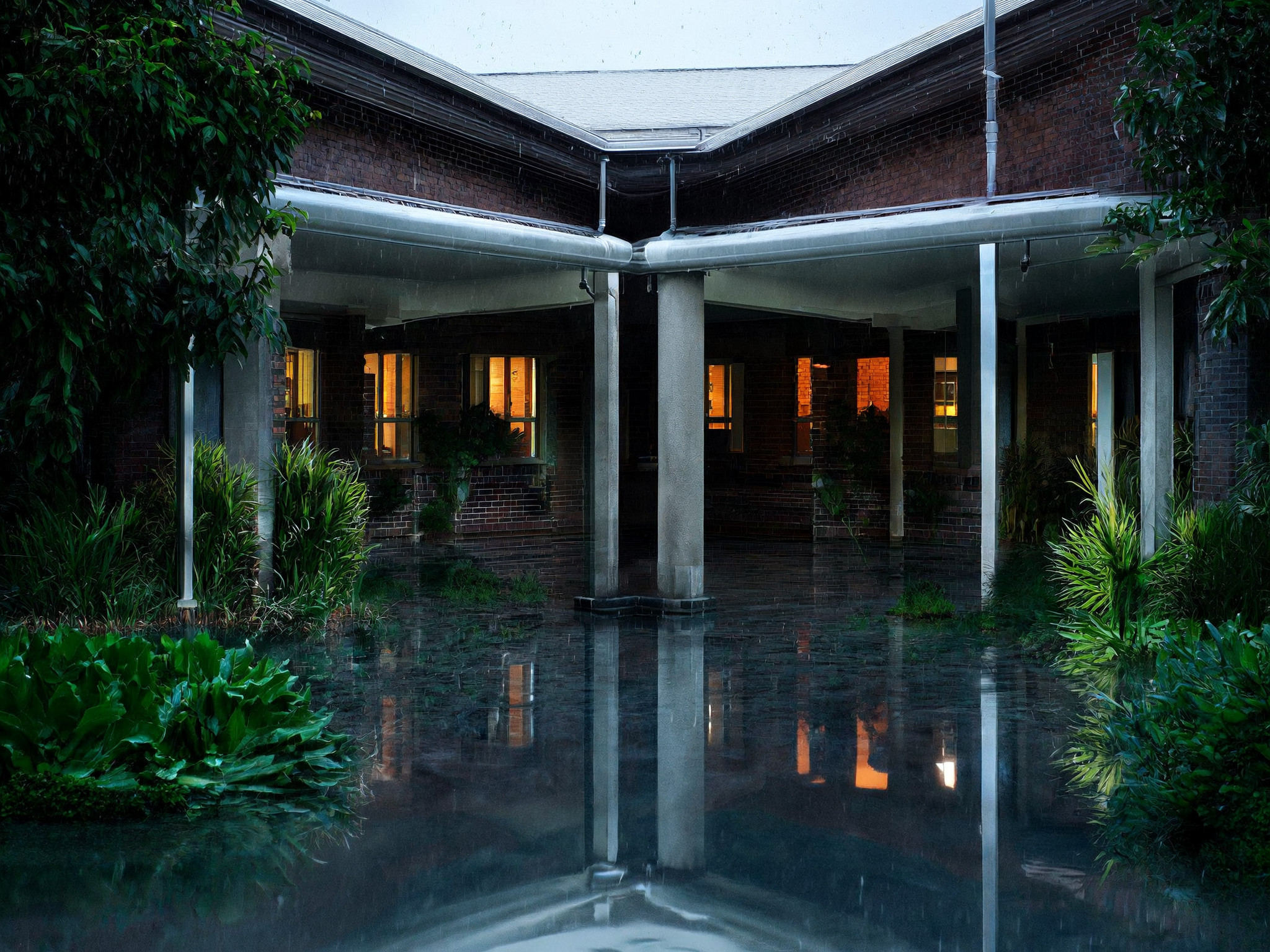}\par
 \includegraphics[width=\linewidth]{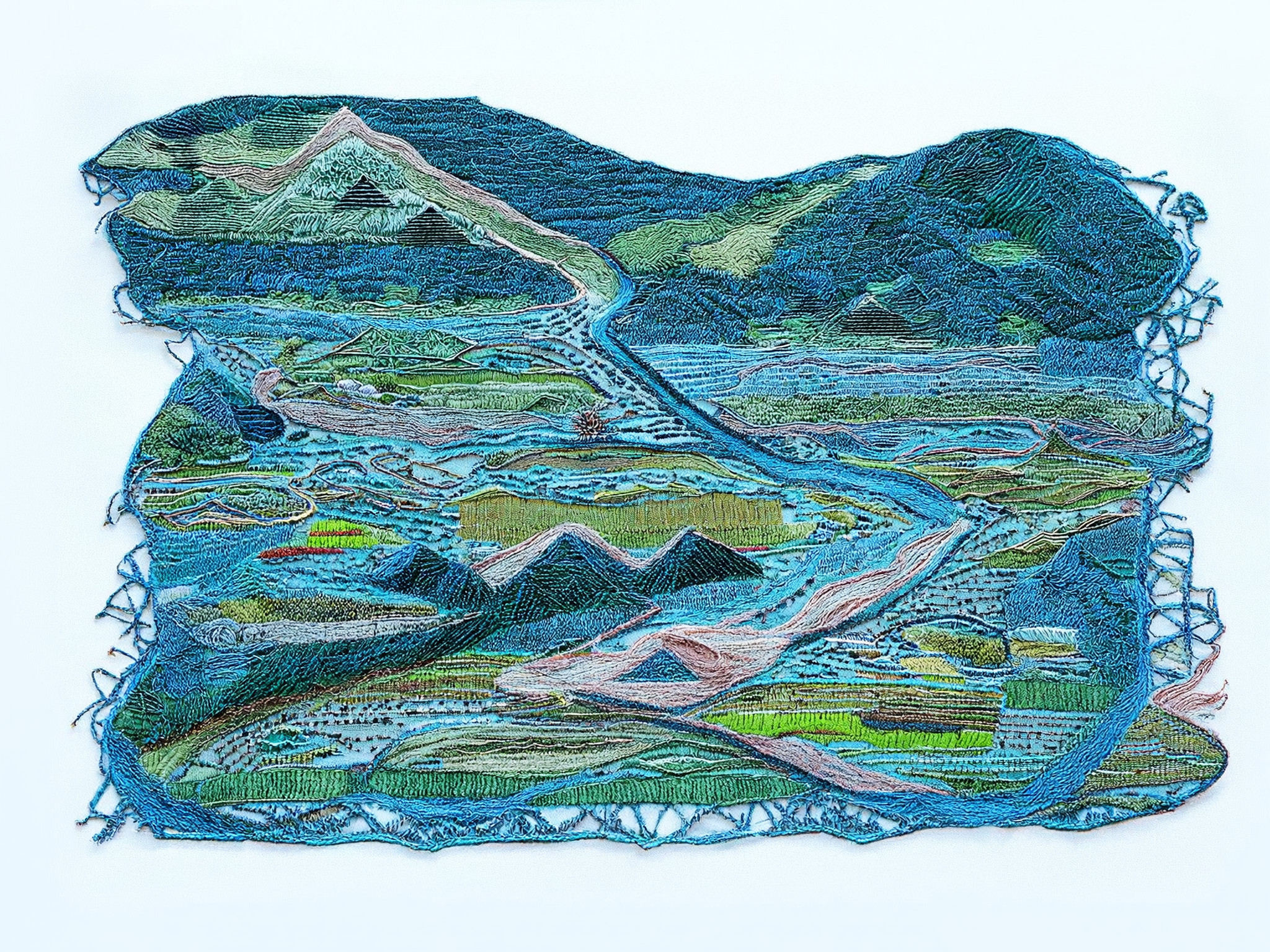}\par
 \includegraphics[width=\linewidth]{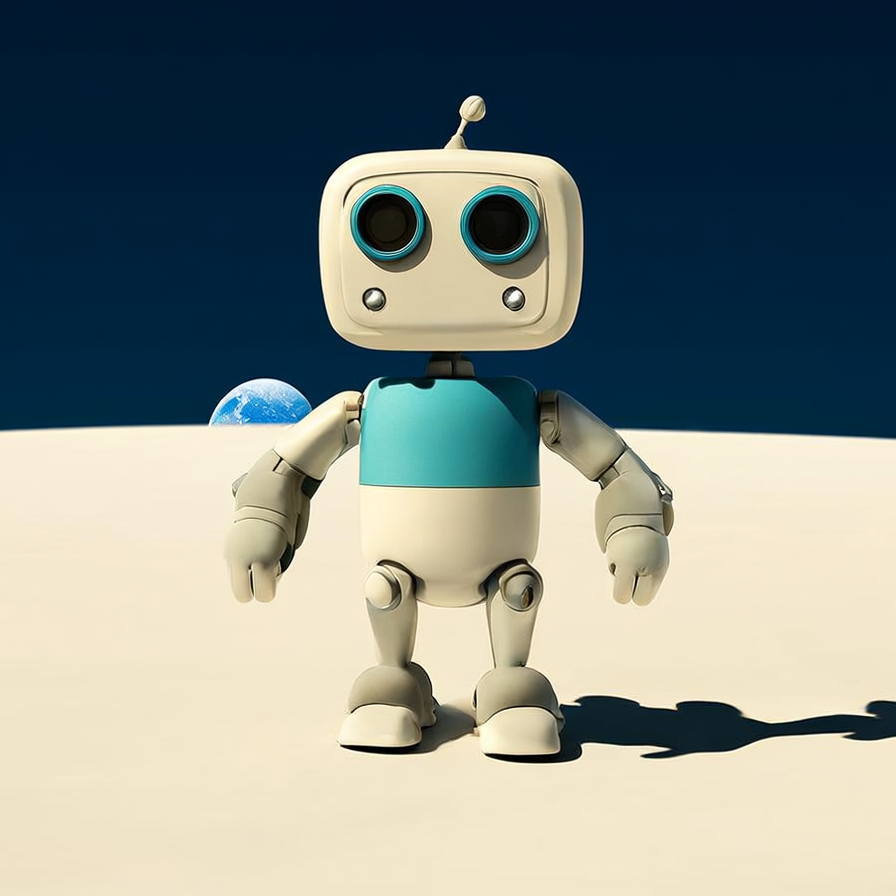}\par
 \end{minipage}}\kern0pt}}
\noindent\makebox[\linewidth][c]{\resizebox{!}{\dimexpr\textheight-24pt\relax}{\usebox{\generationmosaic}}}\par
\captionof{figure}{\textbf{Text-to-image generation.}}\label{fig:showcase-generation}
\clearpage
\showcaseheading{vis:edit-person}{}{}
\phantomsection\label{vis:edit-scene}
\noindent\hfill\begin{minipage}{.90\linewidth}
\noindent\vbox{\offinterlineskip
\hbox{\includegraphics[width=.25\linewidth]{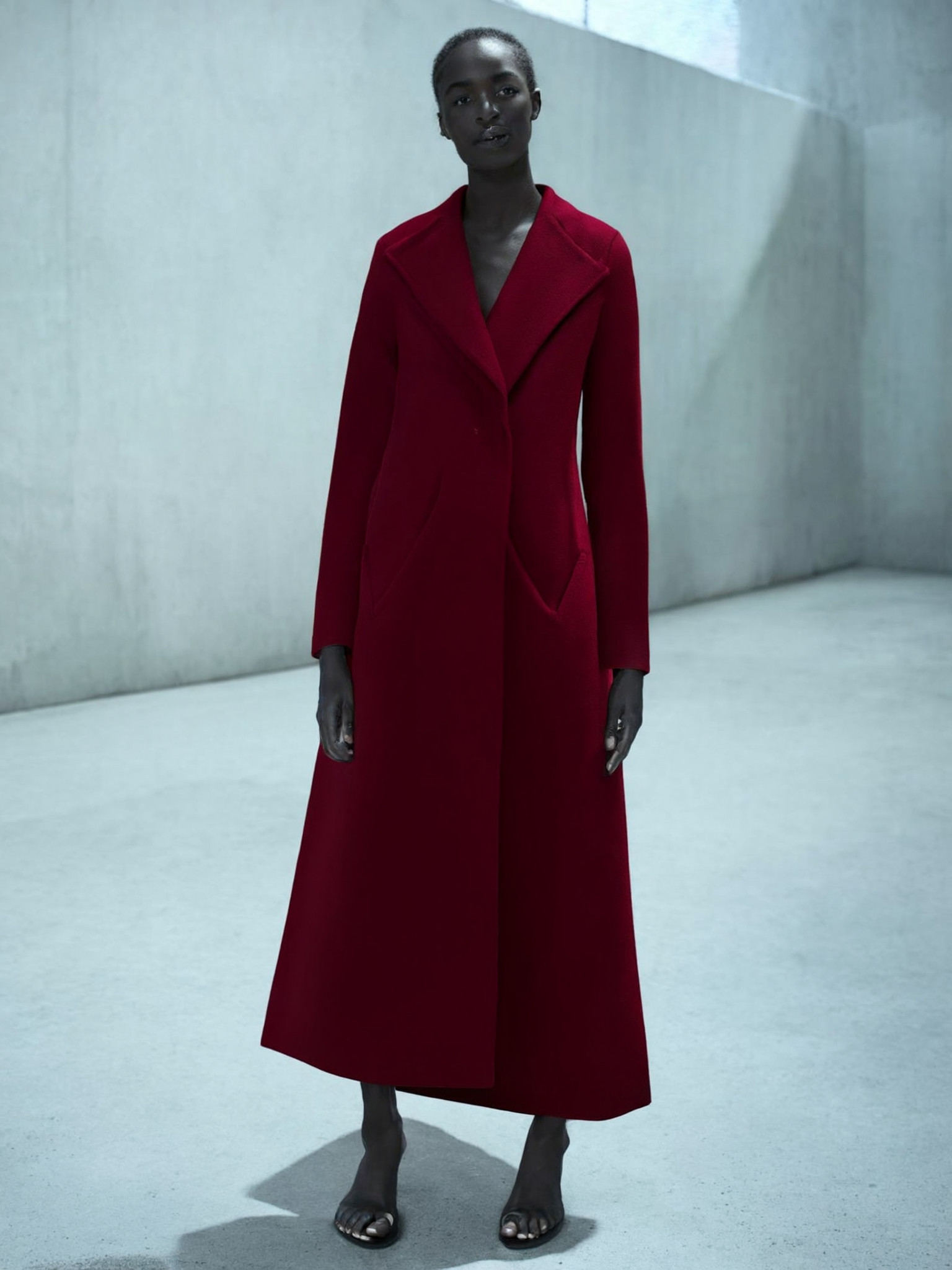}\includegraphics[width=.25\linewidth]{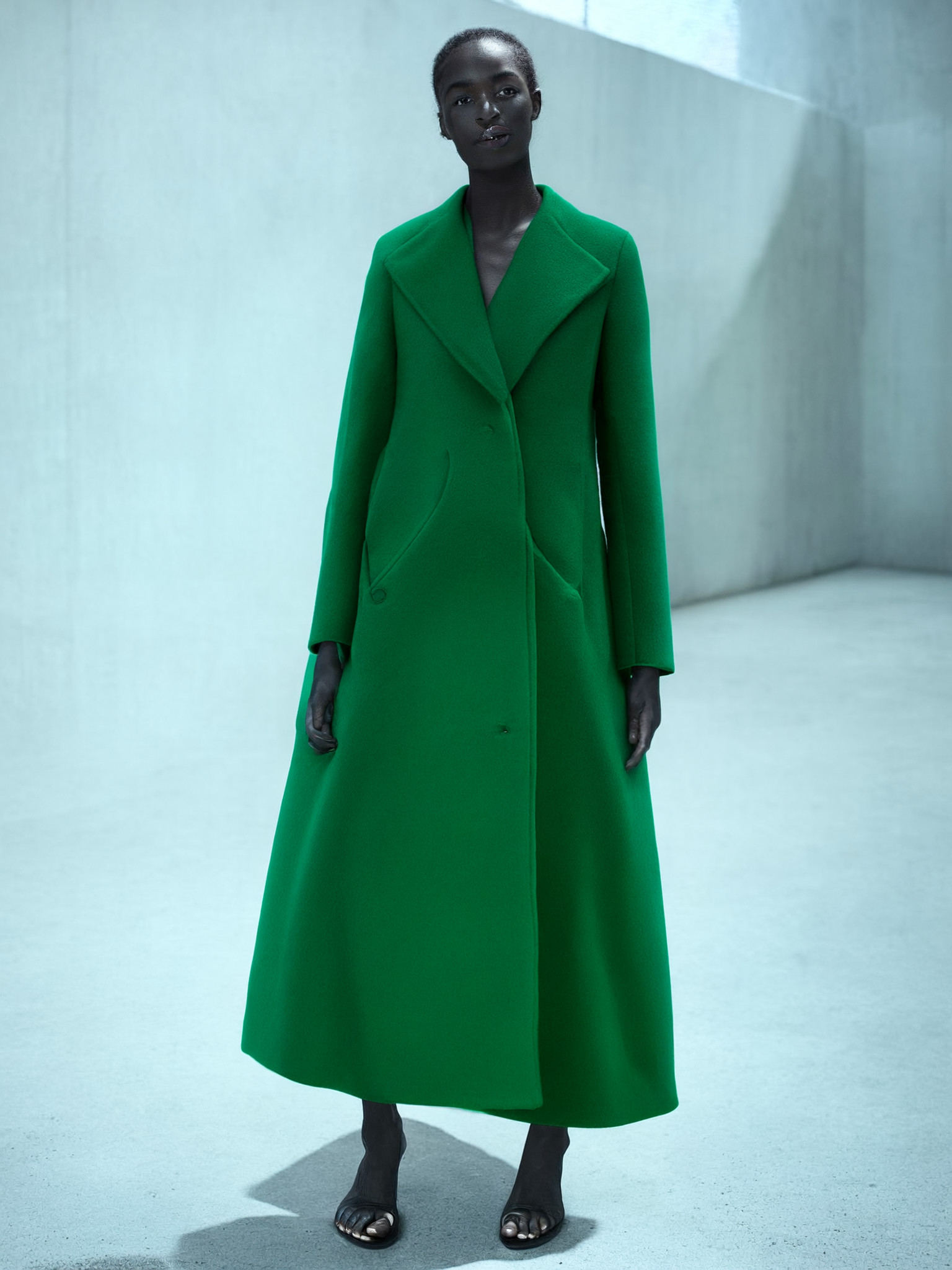}\includegraphics[width=.25\linewidth]{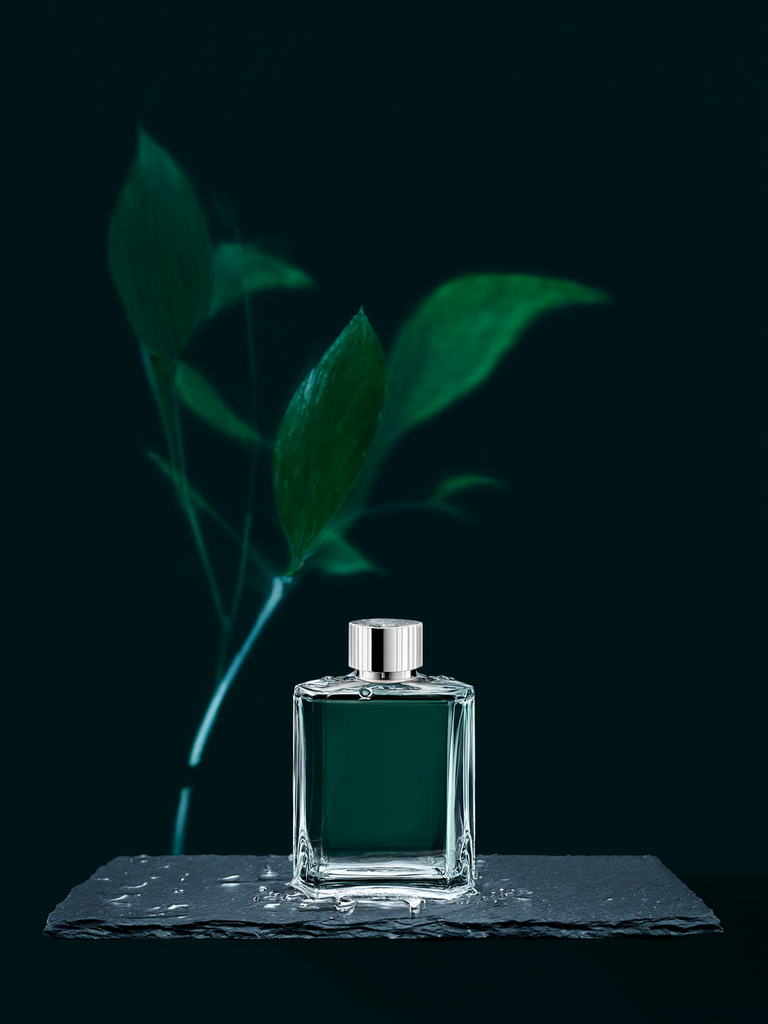}\includegraphics[width=.25\linewidth]{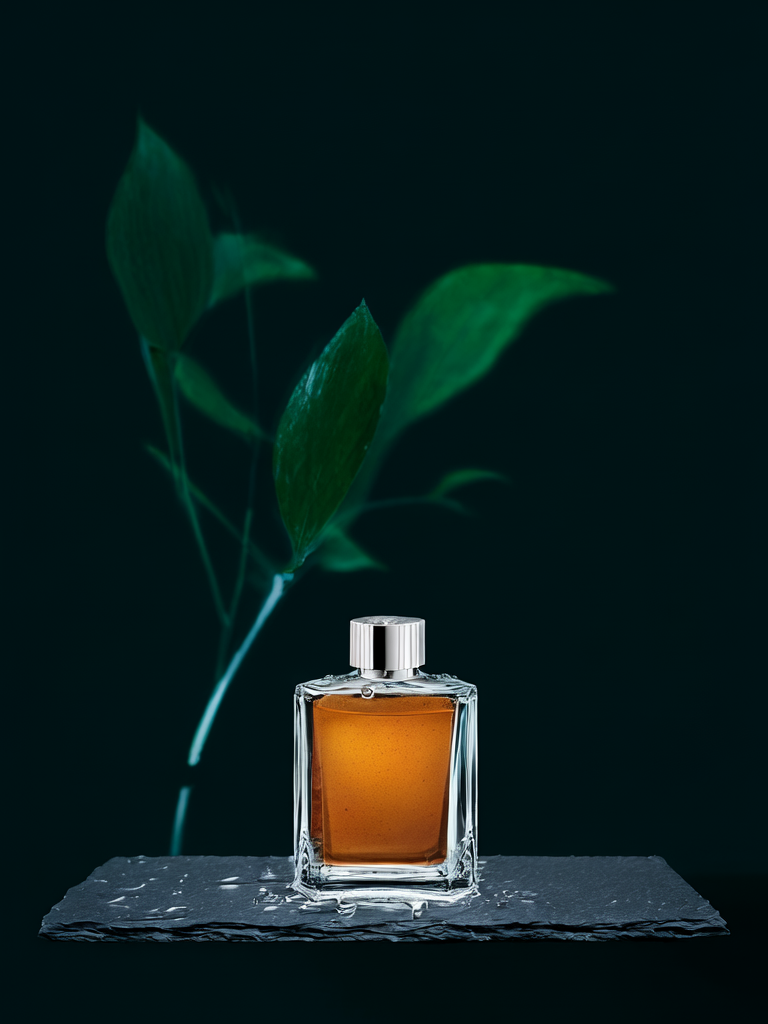}}\hbox{\includegraphics[width=.25\linewidth]{image025.png}\includegraphics[width=.25\linewidth]{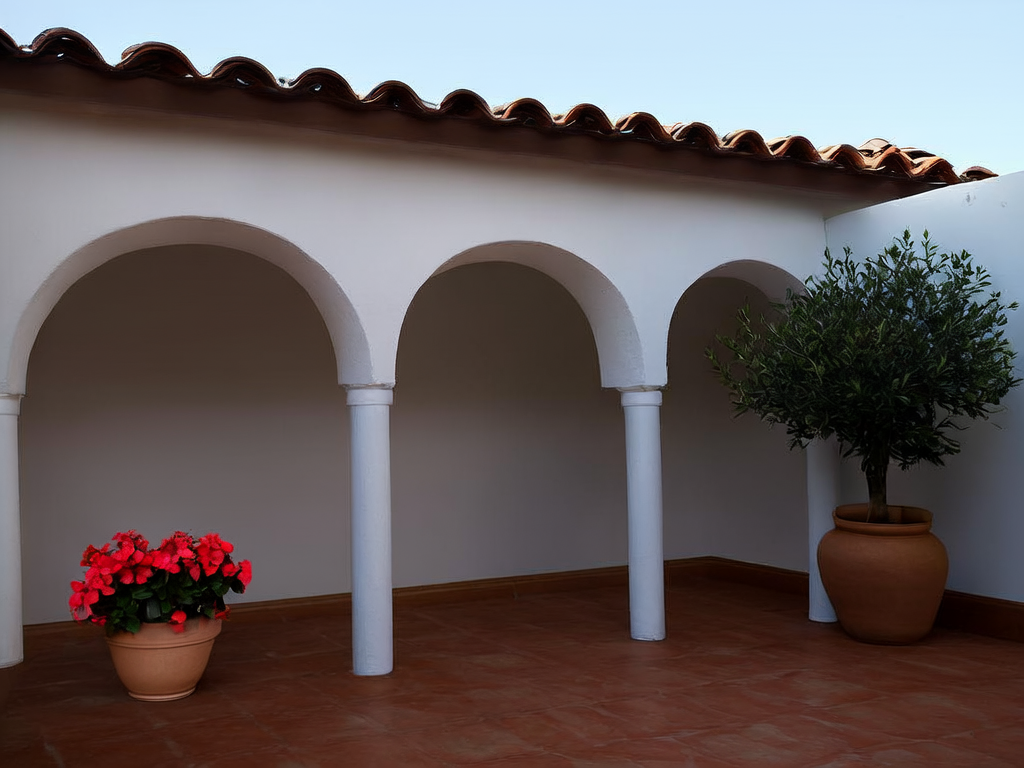}\includegraphics[width=.25\linewidth]{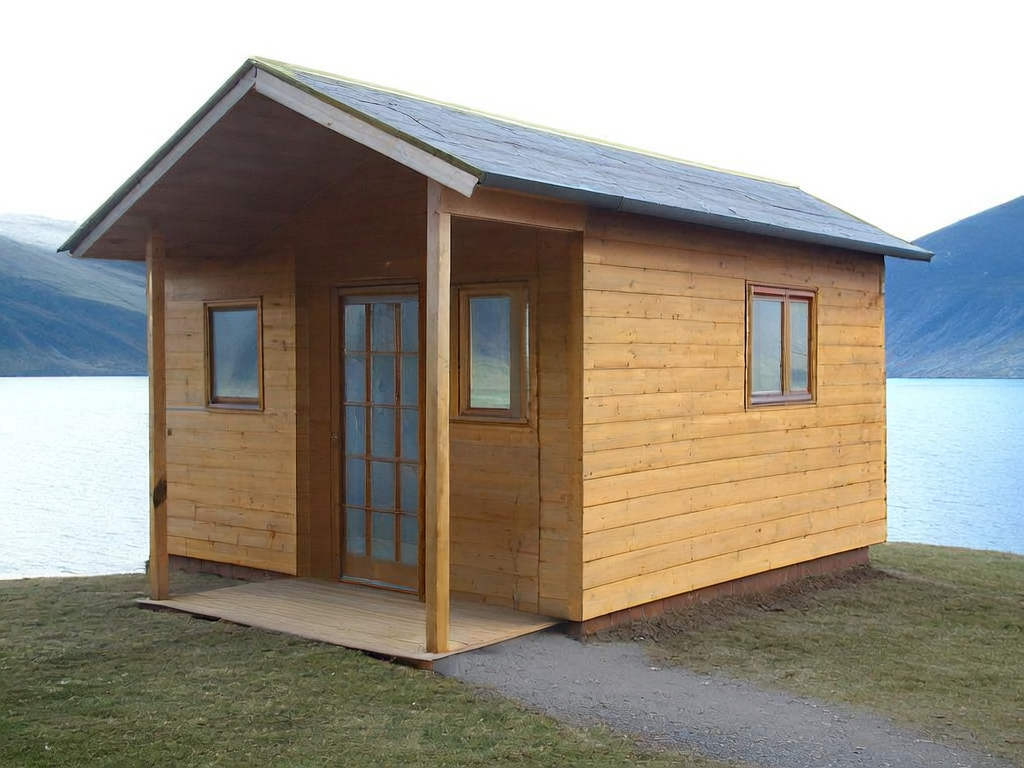}\includegraphics[width=.25\linewidth]{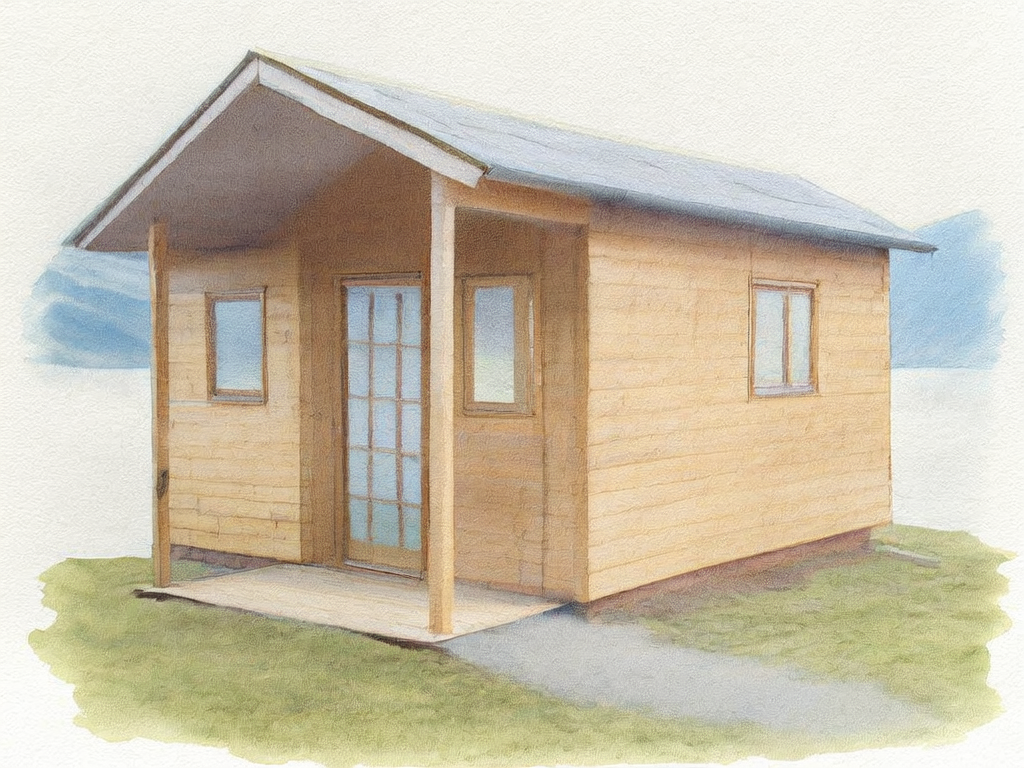}}\hbox{\includegraphics[width=.25\linewidth]{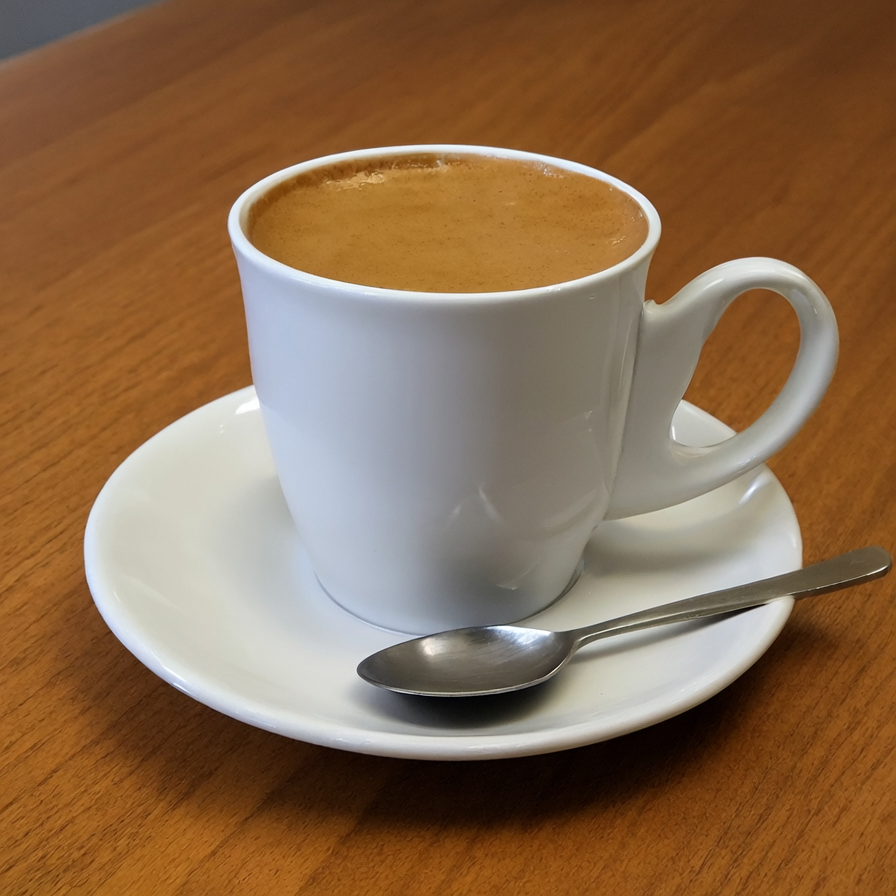}\includegraphics[width=.25\linewidth]{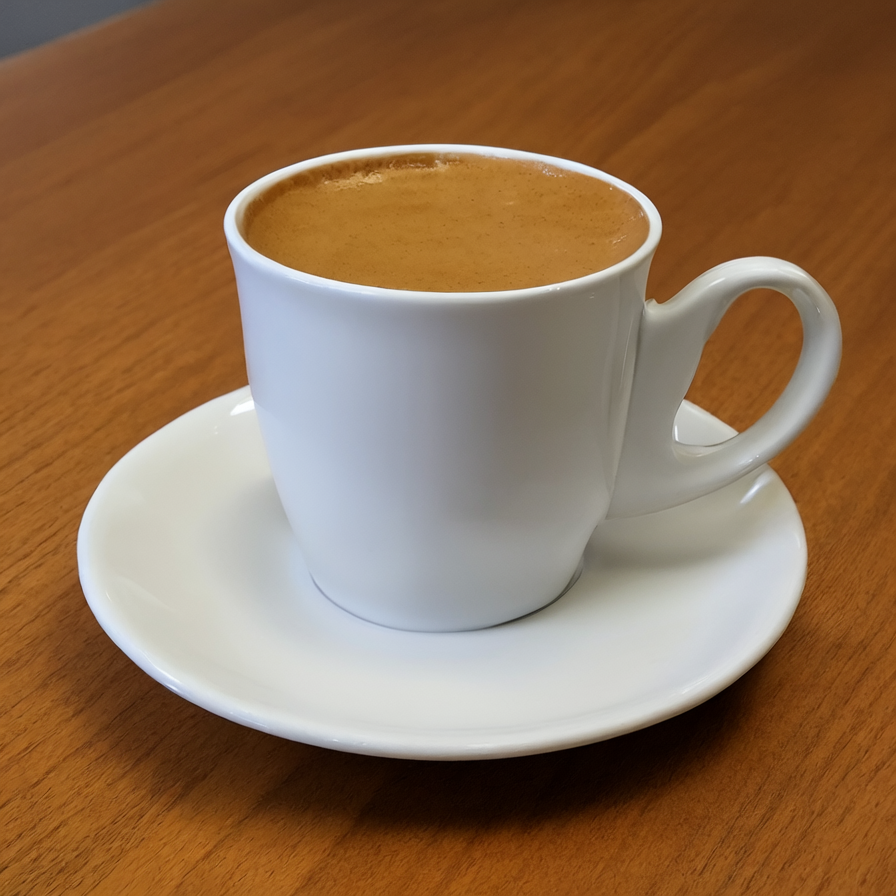}\includegraphics[width=.25\linewidth]{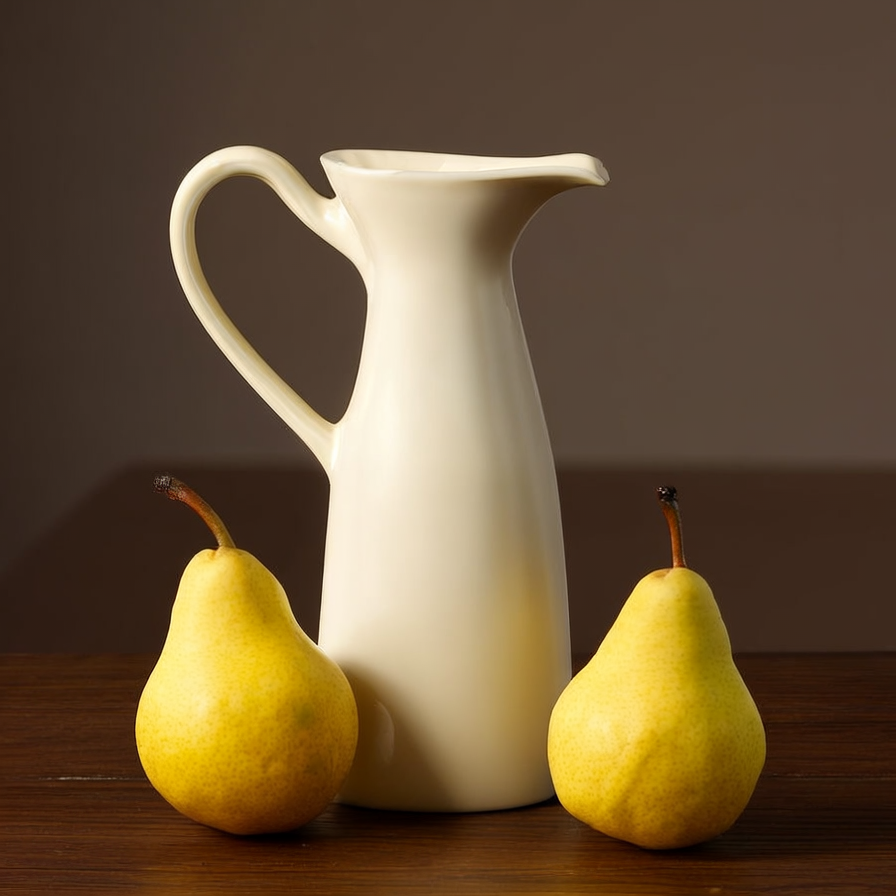}\includegraphics[width=.25\linewidth]{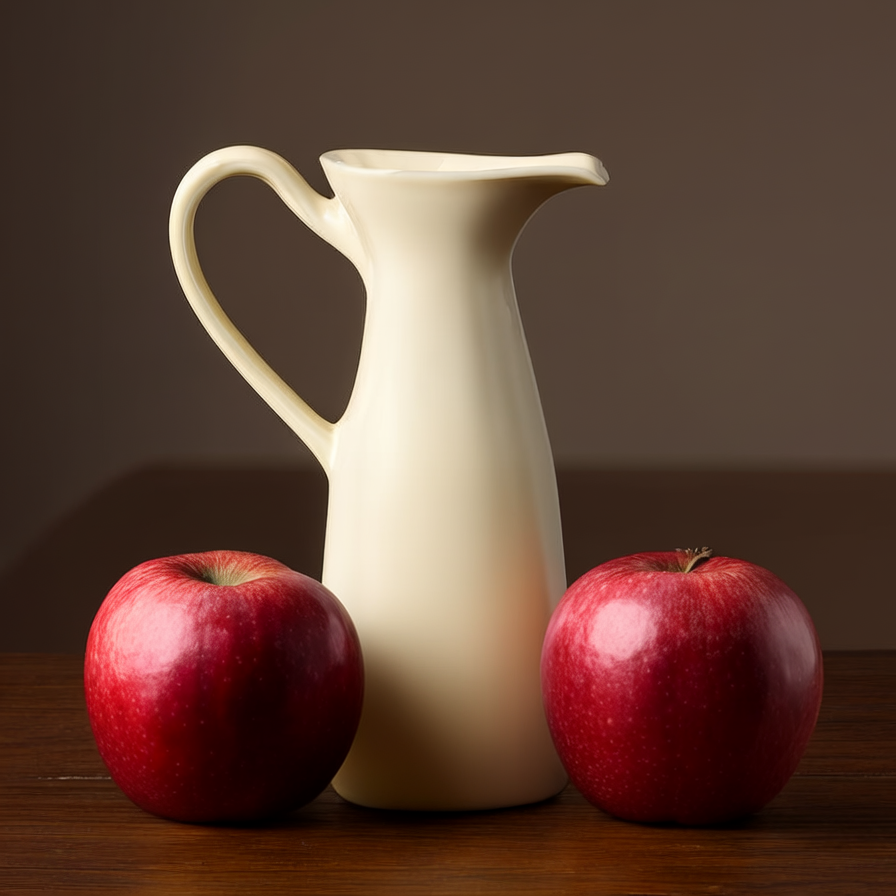}}}\par
\vspace{2mm}
\begingroup

\phantomsection\label{vis:joint-storyboard}\thispagestyle{plain}
\noindent\begin{minipage}{\linewidth}
\begin{minipage}[c]{.32\linewidth}\raggedright
{\fontsize{7.3}{8.4}\selectfont Ocean waves wash against dark coastal rocks beneath a pale overcast sky. Fixed camera; preserve the rocky shoreline and horizon.\par Waves and white sea foam continue moving around the same rocks.\par}
\end{minipage}\hfill
\begin{minipage}[c]{.668\linewidth}\centering
\includegraphics[width=.25\linewidth]{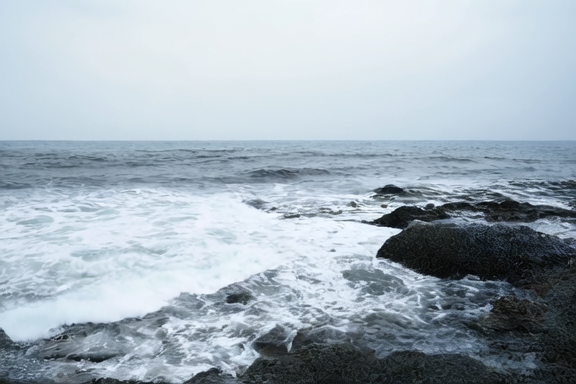}\includegraphics[width=.25\linewidth]{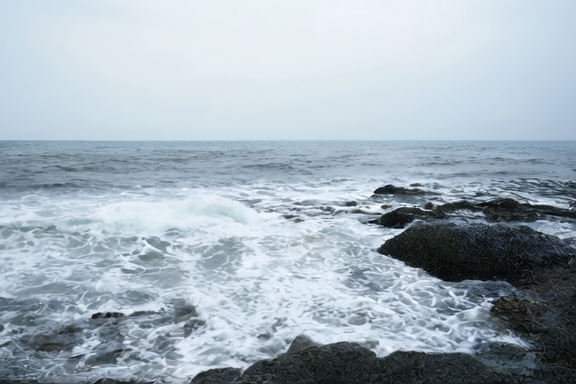}\includegraphics[width=.25\linewidth]{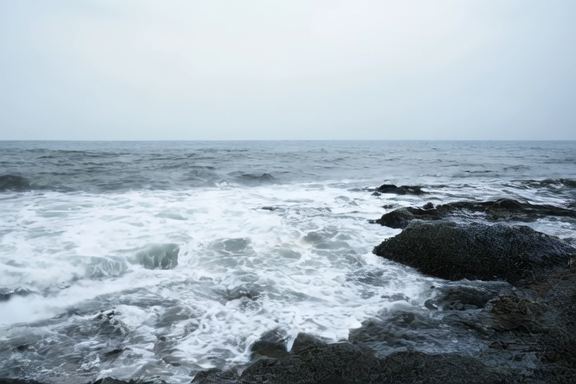}\includegraphics[width=.25\linewidth]{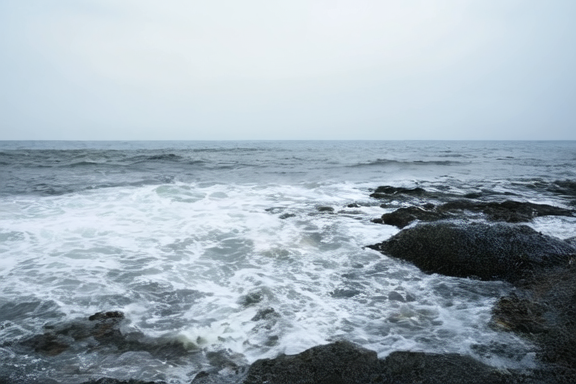}
\end{minipage}
\end{minipage}\par\vspace{.4mm}
{\color{black!20}\hrule height .25pt}\vspace{.4mm}
\noindent\begin{minipage}{\linewidth}
\begin{minipage}[c]{.32\linewidth}\raggedright
{\fontsize{7.3}{8.4}\selectfont A pink rose bud on a green stem against a softly blurred garden. Fixed camera and consistent background.\par The petals gradually unfurl into a full bloom.\par}
\end{minipage}\hfill
\begin{minipage}[c]{.668\linewidth}\centering
\includegraphics[width=.25\linewidth]{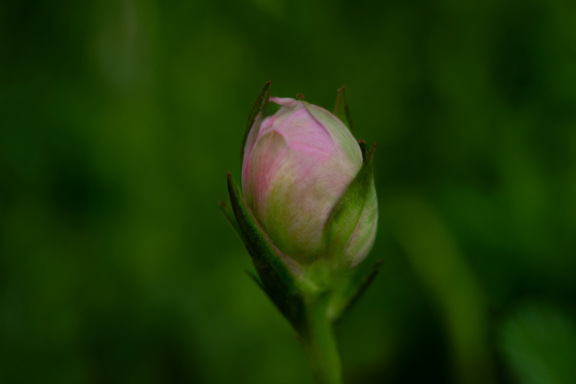}\includegraphics[width=.25\linewidth]{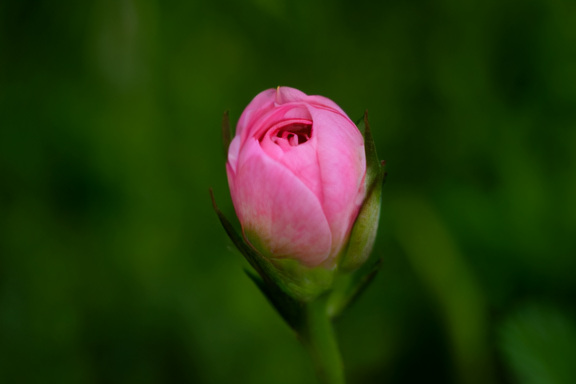}\includegraphics[width=.25\linewidth]{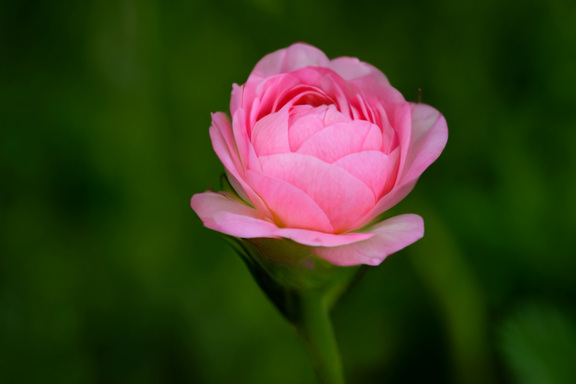}\includegraphics[width=.25\linewidth]{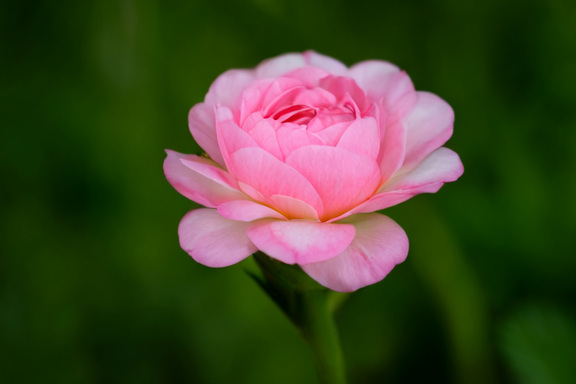}
\end{minipage}
\end{minipage}\par\vspace{.4mm}
{\color{black!20}\hrule height .25pt}\vspace{.4mm}
\noindent\begin{minipage}{\linewidth}
\begin{minipage}[c]{.32\linewidth}\raggedright
{\fontsize{7.3}{8.4}\selectfont A green parrot with a red forehead perches on a wooden branch against a warm gray background. Fixed camera, soft light, same bird and perch.\par Raise the folded wings, stretch them outward, then fold them back.\par}
\end{minipage}\hfill
\begin{minipage}[c]{.668\linewidth}\centering
\includegraphics[width=.25\linewidth]{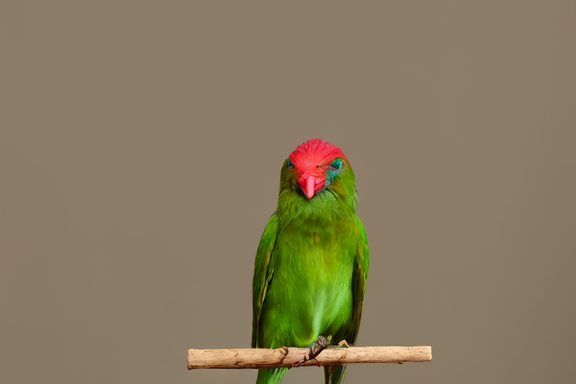}\includegraphics[width=.25\linewidth]{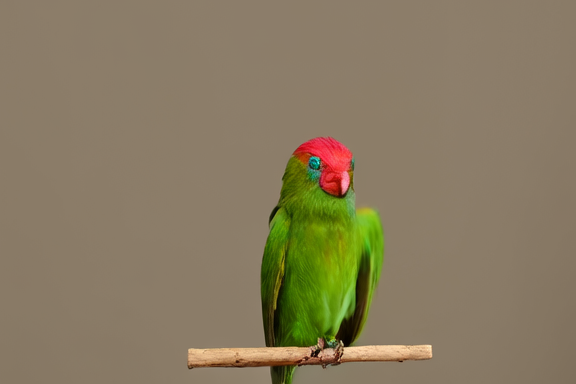}\includegraphics[width=.25\linewidth]{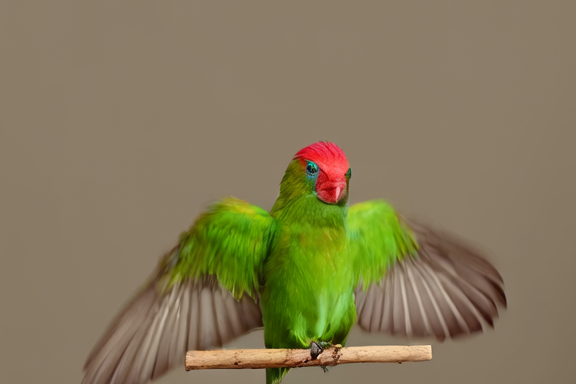}\includegraphics[width=.25\linewidth]{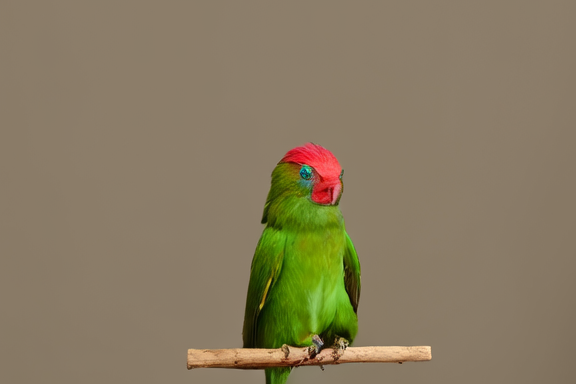}
\end{minipage}
\end{minipage}\par\vspace{.4mm}
{\color{black!20}\hrule height .25pt}\vspace{.4mm}
\noindent\begin{minipage}{\linewidth}
\begin{minipage}[c]{.32\linewidth}\raggedright
{\fontsize{7.3}{8.4}\selectfont A fixed-camera close photograph of a white ceramic cup filled with hot tea on a dark wooden table. Soft side light, plain dark background; preserve the cup and table.\par Wisps of steam curl upward from the hot tea.\par Continue the steam rising above the same cup.\par}
\end{minipage}\hfill
\begin{minipage}[c]{.668\linewidth}\centering
\includegraphics[width=.25\linewidth]{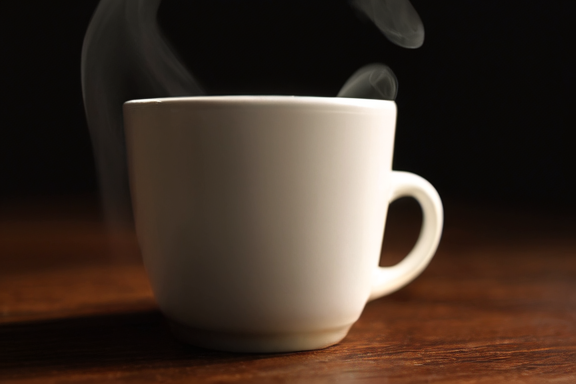}\includegraphics[width=.25\linewidth]{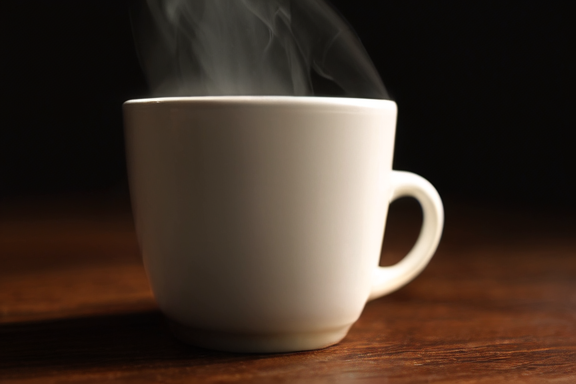}\includegraphics[width=.25\linewidth]{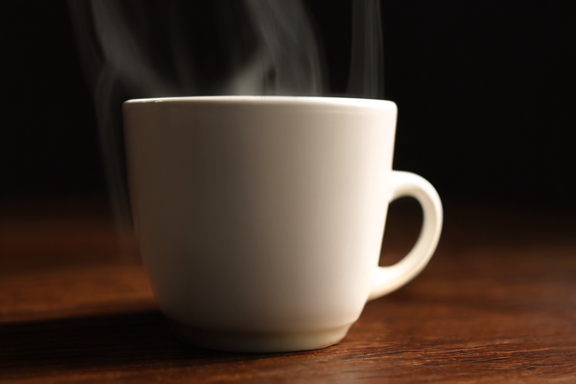}\includegraphics[width=.25\linewidth]{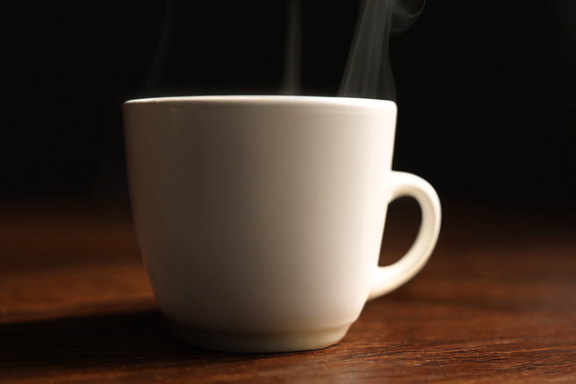}
\end{minipage}
\end{minipage}\par\vspace{.4mm}
{\color{black!20}\hrule height .25pt}\vspace{.4mm}
\noindent\begin{minipage}{\linewidth}
\begin{minipage}[c]{.32\linewidth}\raggedright
{\fontsize{7.3}{8.4}\selectfont A campfire inside a stone ring on bare ground. Fixed camera; keep the stones, logs and surroundings consistent.\par Small flames grow brighter, settle, and leave ash and embers.\par}
\end{minipage}\hfill
\begin{minipage}[c]{.668\linewidth}\centering
\includegraphics[width=.25\linewidth]{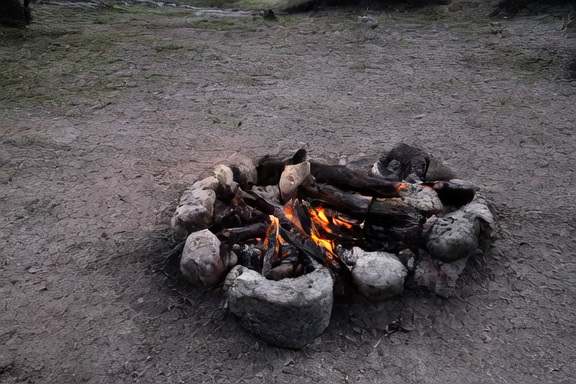}\includegraphics[width=.25\linewidth]{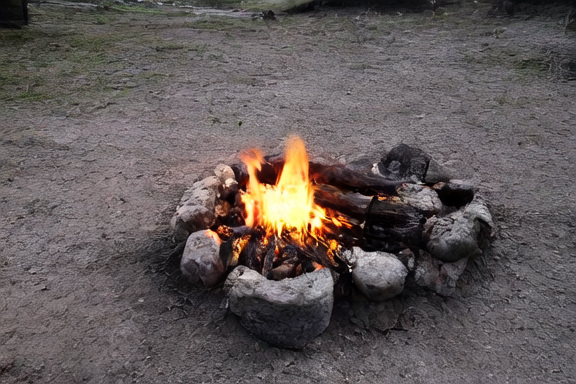}\includegraphics[width=.25\linewidth]{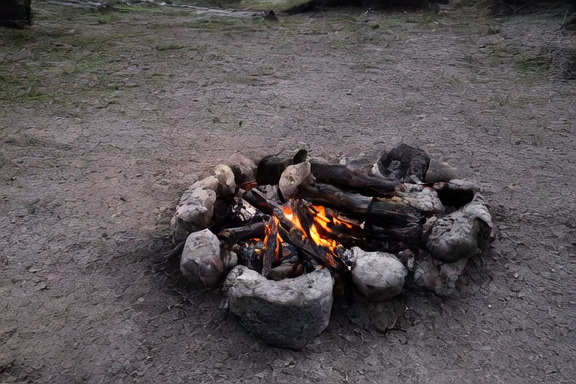}\includegraphics[width=.25\linewidth]{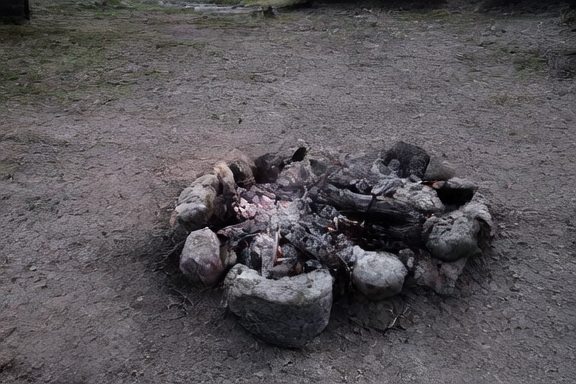}
\end{minipage}
\end{minipage}\par\vspace{.4mm}
{\color{black!20}\hrule height .25pt}\vspace{.4mm}
\endgroup

\end{minipage}\hfill\mbox{}\par
\captionof{figure}{\textbf{Image editing and interleaved generation.} Editing pairs: source (left), result (right).}\label{fig:showcase-portrait}\label{fig:showcase-editing}\label{fig:showcase-storyboards}
\clearpage\endgroup\restoregeometry

\clearpage
\section*{Contents}
\makeatletter
\begingroup
\@starttoc{toc}
\endgroup
\makeatother

\clearpage
\section{Introduction}
\label{sec:introduction}

Multimodal understanding and visual generation place complementary
demands on a model. Understanding requires interpreting visual inputs,
whereas generation requires synthesizing visual content from text or
visual conditions. Combining autoregressive language modeling with
continuous visual prediction has become a central approach to
unification, as demonstrated by Transfusion and
Show-o2~\citep{zhou2024transfusion,xie2025showo2}. BAGEL brings
understanding, generation, and editing into a shared multimodal sequence
and backbone through modality-specific transformer
parameters~\citep{bagel}, while SenseNova-U1 further explores native
unified representations and interleaved vision-language
generation~\citep{diao2026sensenovau1}. In interleaved sequences,
completed visual outputs can in turn serve as context for subsequent
text and image predictions, linking understanding and generation
across successive outputs.

UniAR~\citep{peng2026uniar} enables such context reuse through a shared discrete visual
tokenizer, allowing generated tokens to become context without
re-encoding the image. In continuous visual
generation, however, context reuse must account for the iterative
denoising process. Consider a model that generates an image, explains it,
and then edits it in response to a new instruction. During denoising,
the image is represented by a noisy target; once generated, its clean
representation becomes context for subsequent predictions. Retaining
intermediate denoising states would expose later predictions to transient
noise, whereas discarding visual history would remove the reference
needed for consistent editing. Unified modeling therefore requires
separating noisy prediction targets from reusable clean visual context.

High-resolution images and multi-image sequences increase the
visual-token workload of unified multimodal models. Hybrid attention
combines full attention for explicit token
interactions~\citep{vaswani2017attention} with recurrent linear attention
that propagates context through a persistent state at a cost linear
in sequence length~\citep{katharopoulos2020linear,yang2025gateddelta}.
Kimi Linear and Olmo Hybrid explore layer-wise combinations of these
operators in language
modeling~\citep{kimi2025linear,merrill2026olmohybrid}, and Qwen3.5
provides a backbone combining full attention with GatedDeltaNet
layers~\citep{qwen2026qwen35modelcard}. We investigate how full and
linear attention interact across layers to support causal text
prediction and iterative visual denoising within a unified model,
and characterize its multimodal capabilities and computational
trade-offs.

We present \textbf{\methodname{}}, a unified model for multimodal understanding and visual generation. The \modelname{}
configuration has approximately 9B parameters in each of its
understanding and generation branches. Building on the
Mixture-of-Transformer-Experts formulation~\citep{liang2024mixture,bagel}
and the Qwen3.5 hybrid-attention backbone~\citep{qwen2026qwen35modelcard},
\methodname{} combines expert routing and hybrid attention in a \textbf{Hybrid MoT Architecture}. Text and semantic visual features use the understanding expert, whereas VAE latents use the generation expert. Within each multimodal sequence, the experts exchange information through full attention and shared recurrent states. Autoregressive text prediction and visual flow matching~\citep{lipman2023flowmatching} jointly support multimodal understanding, text-to-image generation, reference-conditioned image editing, and interleaved prediction within the same model.

The information-flow design separates \emph{private prediction targets} from \emph{public context}. A noisy visual block reads its public prefix but is hidden from later blocks. Full-attention layers implement this rule with block visibility masks. Linear-attention layers process the block recurrently, but discard its private state and convolution-history updates after local computation on a noisy block.
Clean visual representations instead update the public context. During training, this context is obtained from ground-truth
images using teacher forcing~\citep{williams1989teacher}; at inference,
it is built by encoding completed model outputs.

The training data span visual understanding, image generation and editing, interleaved sequences, and text-only language modeling. Understanding data cover captioning, visual question answering, OCR, GUI tasks, and grounding. Task-specific adapters map each data type to a common sequence plan. Online packing combines samples within token budgets while preserving
sample boundaries, within-sample block structure, and supervision locations. Full-attention masks prevent cross-sample visibility, while linear-attention layers reset their recurrent state and convolution history at sample boundaries.
The selected sources contain 159.7M generation samples and 213.4M understanding samples, including 10.0M OCR and 4.5M grounding samples, as well as 624.8K interleaved sequences. The data extend beyond image–caption pairs to support general semantic learning, specialized visual tasks, and the modeling of state changes over time.

Training proceeds through four main stages. The Joint Multimodal Foundation stage trains the model on text prediction, visual understanding, and low-resolution image generation. The Progressive Resolution Scaling and High-Resolution Refinement stages extend image generation to 2K while retaining understanding supervision. Supervised Capability Refinement (SFT) then improves understanding accuracy, instruction adherence, image-generation quality, and editing consistency. 
Following these four stages, we further adapt the model to video generation and editing using a mixture of image and video data.

Our main contributions are:
\begin{itemize}[leftmargin=*,itemsep=3pt,topsep=3pt]
    \item \textbf{Hybrid MoT Architecture.} We combine token-type expert
    routing with full attention and GatedDeltaNet linear attention in a
    shared backbone. 
    This hybrid design combines explicit multimodal interactions with
    efficient context propagation across long visual sequences.

    \item \textbf{Context handling for interleaved prediction.} We
    introduce a blockwise information-flow protocol that keeps noisy
    visual targets private while allowing clean visual content to
    condition later text and visual outputs under both attention operators.

    \item \textbf{Unified data interface and progressive training.}
    We map understanding, generation, editing, interleaved, and
    text-only data to task-specific sequence plans, then pack them
    through a shared interface. A four-stage training curriculum yields
    a single model for visual understanding, 2K image generation,
    editing, and interleaved prediction.
\end{itemize}


\section{Related Work}
\label{sec:related_work}

\subsection{Unified Multimodal Understanding and Generation}
\label{sec:related_unified}

\revised{{}Unified multimodal models use a shared model
to understand and generate visual content. Transfusion combines
autoregressive text prediction with image diffusion~\citep{zhou2024transfusion}.
Mixture-of-Transformers uses separate parameters for each modality while
sharing attention across them~\citep{liang2024mixture}. BAGEL builds on
this approach and trains a unified model on interleaved multimodal
data~\citep{bagel}. These works form the basis of our shared architecture
and joint text and visual training objectives.}

\revised{{}Other approaches explore alternative visual representations and broader task coverage. Tuna-2 uses direct pixel-patch embeddings in
place of separate visual encoders~\citep{liu2026tuna2}. SenseNova-U1
supports understanding, image generation, and interleaved vision-language
generation with dense and MoE backbones~\citep{diao2026sensenovau1}.
SenseNova-Vision handles different vision tasks through text, image, or
mixed outputs without separate prediction heads~\citep{diao2026sensenovavision}.
\methodname{} uses semantic visual features and VAE latents, and studies
how hybrid attention lets these representations share context.}

\revised{{}Sharing visual context also allows one
output to guide the next. UniAR uses a shared discrete visual tokenizer, allowing generated visual tokens to be reused directly as context without re-encoding the image~\citep{peng2026uniar}. Context Unrolling studies reasoning through
text, images, video, and geometry before making
predictions~\citep{yang2026contextunrolling}. \methodname{} addresses context reuse in continuous visual generation by separating noisy targets from clean context through attention masks and recurrent-state retention rules.}

\revised{{}A model that supports both tasks does not
necessarily improve one task by learning the other. VGAU-Diag tests when
generated visual aids help understanding and finds cases where models
fail to use even correct visual aids~\citep{zhu2026vgau}. Wu et al. examine
how understanding and generation help each other at the representation,
task, and system levels, and identify task specialization and shared
knowledge as important factors~\citep{wu2026synergy}. These studies highlight the distinction between joint task support and cross-task synergy.}

\subsection{Efficient and Hybrid Attention Mechanisms}
\label{sec:related_hybrid}

\revised{{}Recurrent attention stores past information
in a state that is updated as tokens are processed. Gated DeltaNet uses
adaptive forgetting and delta-rule updates~\citep{yang2025gateddelta}.
Kimi Linear adds finer gating through Kimi Delta Attention and combines
it with Multi-Head Latent Attention across layers~\citep{kimi2025linear}.
Gated DeltaNet-2 uses separate erase and write gates for each memory
channel~\citep{gateddelta2026two}. These methods control how memory is
updated. These methods govern token-level memory updates; \methodname{} additionally enforces block-level retention rules to prevent noisy-target updates from entering public context.}

\revised{{}Hybrid backbones combine recurrent layers
with full attention. \methodname{} starts from Qwen3.5, which combines
GatedDeltaNet and full-attention layers~\citep{qwen2026qwen35modelcard}.
Olmo Hybrid studies this combination through theory and controlled
language-modeling experiments~\citep{merrill2026olmohybrid}. In vision,
SoLA-Vision studies how to arrange linear and softmax attention across
layers~\citep{li2026solavision}. SANA-Video 2.0 combines gated linear
attention, periodic softmax layers, and attention residuals for video
diffusion~\citep{chen2026sanavideo2}. Our work explores hybrid attention
in a model trained jointly for multimodal understanding and generation.}

\revised{{}Attention types can also be combined within
a layer. HydraHead assigns full and linear attention to different heads,
keeps full attention for heads important to retrieval, and normalizes
their outputs before combining them~\citep{hydrahead2026}.
\methodname{} combines the two attention types across layers. Reusing
context also requires careful state handling: HARTS studies shared
prefixes in branching agent rollouts, including state transfers and
activation recomputation~\citep{harts2026}. Although HARTS focuses on
agent reinforcement learning, it illustrates the need to coordinate
recurrent states when computation branches or context is reused.}


\section{Model Design}
\label{sec:architecture}

\revised{\methodname{} uses a \textbf{Hybrid MoT Architecture} for
multimodal understanding and generation. Building on the multimodal
Mixture-of-Transformer-Experts (MoT) formulation of
BAGEL~\citep{bagel} and the hybrid attention backbone of
Qwen3.5~\citep{qwen2026qwen35modelcard}, the architecture combines
specialized understanding and generation experts with a layer-wise
mixture of full attention and recurrent linear attention in one shared
backbone. \modelname{} uses approximately 9B parameters per branch,
with the 9B designation referring to the per-branch backbone size rather
than the total model size.}

\begin{figure}[t]
    \centering
    \includegraphics[width=\linewidth]{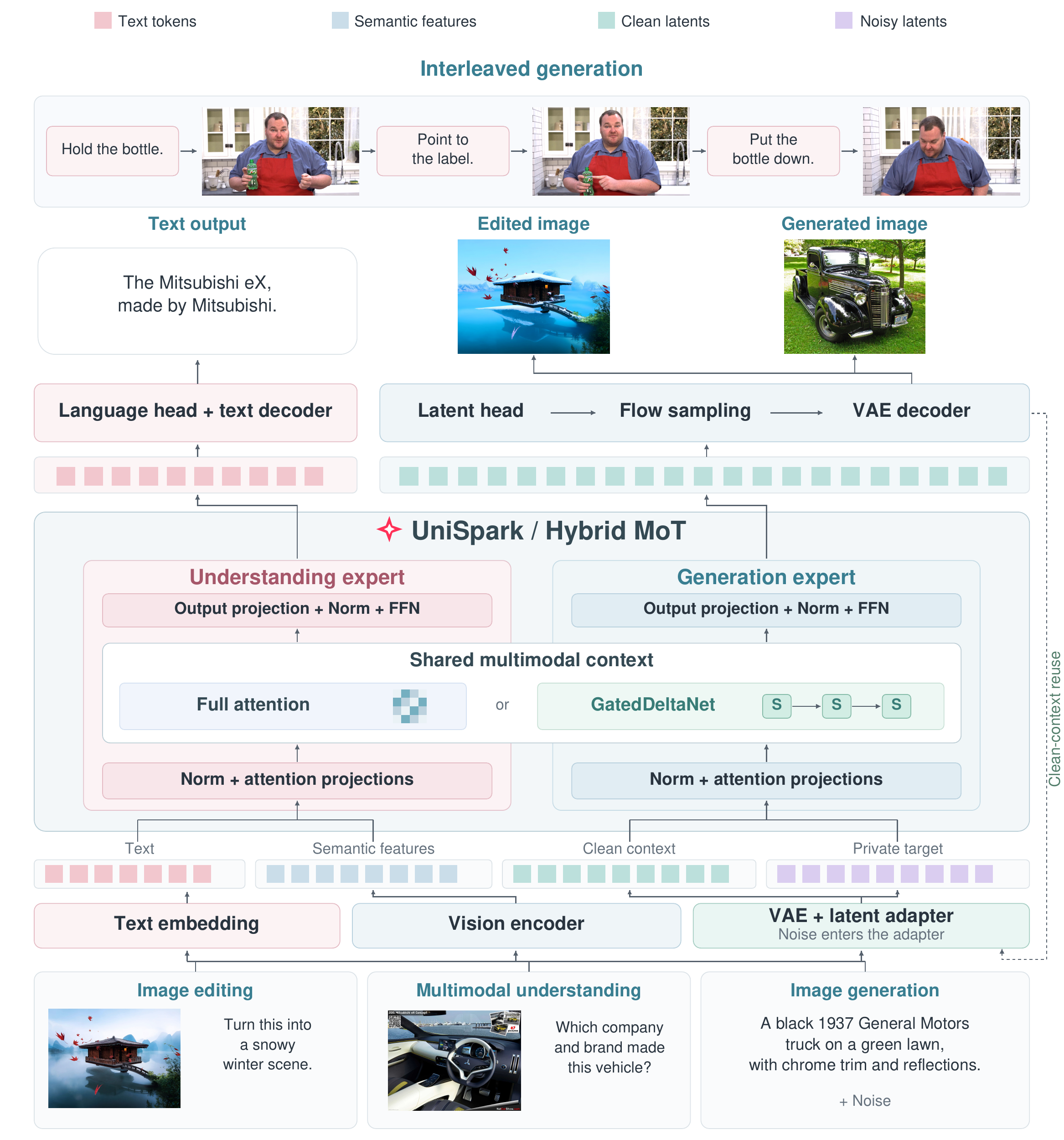}
    \caption{\revised{\textbf{Hybrid MoT Architecture of \methodname{}.}
    Token-type routing selects understanding or generation parameters,
    while full-attention and recurrent linear-attention layers exchange
    shared multimodal context for text prediction, image generation,
    editing, and interleaved outputs.}}
    \label{fig:huge_architecture}
\end{figure}

Causal language prediction and iterative visual generation impose
different requirements on information flow: text predicts the next token
from its prefix, visual generation refines a block of noisy latent tokens,
and subsequent outputs must condition on completed visual content without
inheriting intermediate noise states. The experts exchange context through
full and recurrent linear attention, with explicit rules governing context
reuse across visual outputs.
\figref{fig:huge_architecture} summarizes this organization.

\subsection{Unified Multimodal Representations}
\label{sec:huge_representations}

A multimodal example is represented as an ordered sequence of text and
visual blocks, together with conversation roles, media boundaries, and
supervision indicators. Text is mapped through the language embedding
layer. Images and video frames used for understanding are encoded by the
Qwen3.5 vision encoder, whose output features are mapped into the language
model's hidden space. Images and videos used as generation targets or clean
context are both encoded by the Wan2.2 VAE~\citep{wan2025wan22}; the resulting latents are patchified and
mapped to the backbone hidden space through a latent adapter. Semantic
features provide an understanding-oriented representation, while VAE
latents provide the continuous representation used for visual prediction.
Predicting in a learned visual latent space follows the latent generative
modeling approach of Rombach et al.~\citep{rombach2022latentdiffusion}.

Let $\mathbf{z}_0$ denote a sequence of clean visual latent patches and
$\mathbf{z}_t$ its noisy counterpart at noise level $t$. The
generation-side input takes the form
\begin{equation}
    \mathbf{h}^{\mathrm{vis}}_t
    = A_{\mathrm{in}}(\mathbf{z}_t)
      + E_{\mathrm{time}}(t)
      + E_{\mathrm{pos}}(\mathbf{p}),
    \label{eq:huge_visual_embedding}
\end{equation}
where $A_{\mathrm{in}}$ projects latent patches to the backbone width,
$E_{\mathrm{time}}$ embeds the noise level $t$, and
$E_{\mathrm{pos}}(\mathbf{p})$ encodes the spatial or spatiotemporal
coordinates $\mathbf{p}$. In addition to this additive position embedding,
full-attention layers use multimodal rotary position embeddings to
organize text and visual tokens within the sequence. Clean context latents
are assigned $t=0$, while noisy target latents use $t>0$ and receive
flow-matching supervision.

\subsection[Hybrid MoT Architecture]{\revised{Hybrid MoT Architecture}}
\label{sec:huge_hybrid_attention}

\revised{The architecture combines token-type expert routing with a
layer-wise mixture of full attention and GatedDeltaNet linear attention.}

\paragraph{Expert specialization.}
Text and semantic visual tokens use the understanding expert, while VAE
latent tokens use the generation expert. Each decoder layer contains
expert-specific normalization, attention projections, and feed-forward
parameters. The experts exchange information through the context permitted
by each attention operator.

\paragraph{Hybrid attention.}
Full-attention layers compute explicit query--key interactions over the
permitted context. GatedDeltaNet layers~\citep{yang2025gateddelta}
summarize preceding context in a recurrent matrix state, with computation
that scales linearly with sequence length.
A simplified single-head update is
\begin{equation}
    \begin{aligned}
        \overline{\mathbf{S}}_i &= \alpha_i\mathbf{S}_{i-1},\\
        \mathbf{S}_i &= \overline{\mathbf{S}}_i
          + \beta_i\mathbf{k}_i
            \bigl(\mathbf{v}_i
            - \overline{\mathbf{S}}_i^{\top}\mathbf{k}_i\bigr)^{\top},\\
        \mathbf{o}_i &= \mathbf{S}_i^{\top}\mathbf{q}_i,
    \end{aligned}
    \label{eq:huge_gated_delta}
\end{equation}
where $i$ indexes tokens, $\alpha_i$ controls state decay, and $\beta_i$
controls the delta update. Understanding and generation tokens use
their respective parameters to compute these quantities but share the
recurrent state and short convolution history along the permitted
multimodal history. Switching experts therefore does not reset the
accumulated context.

\begin{figure}[t]
    \centering
    \input{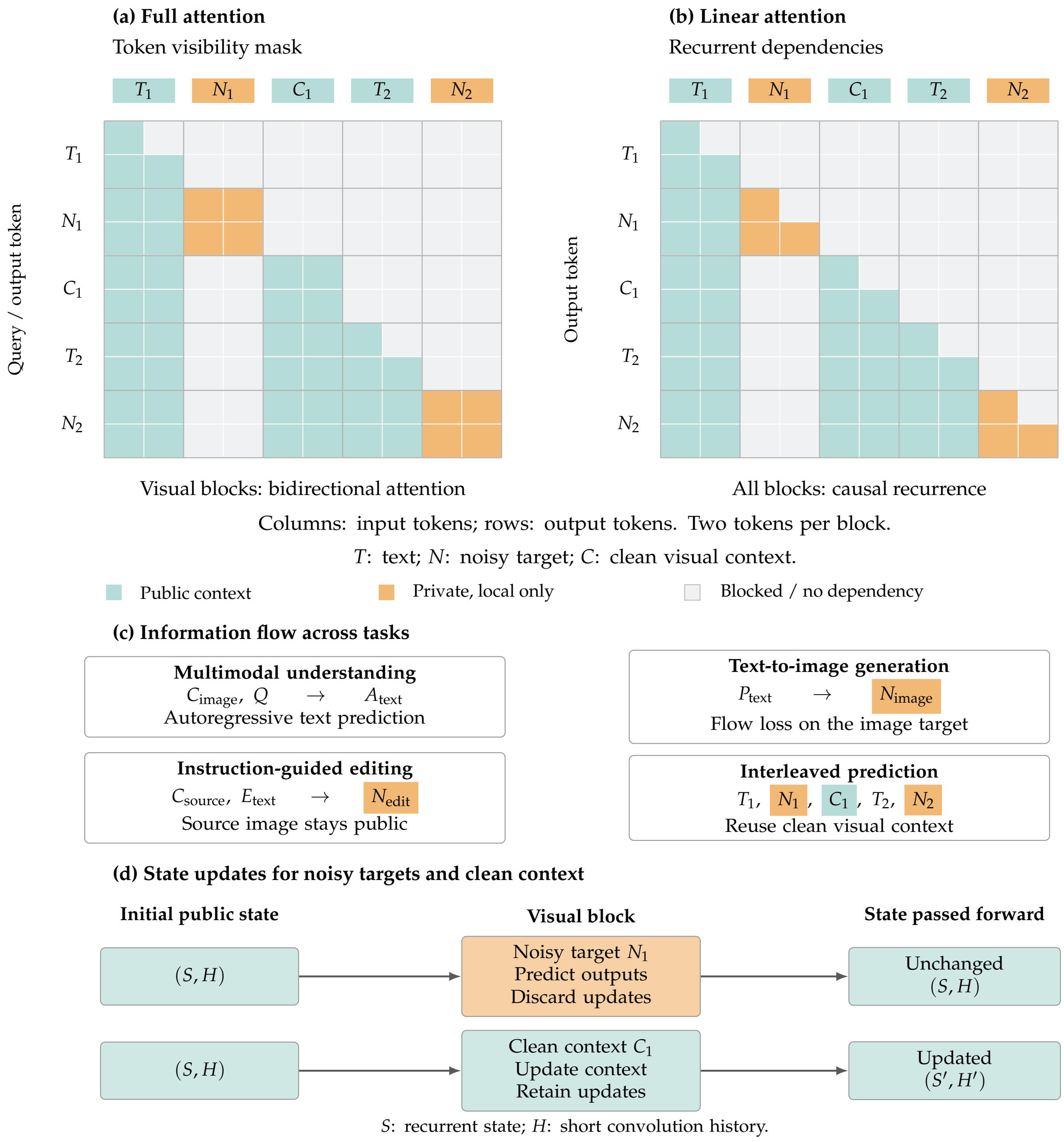}
    \caption{\textbf{Information flow under hybrid attention.} \revised{Visibility masks and recurrent-state rules preserve the same public/private boundary, allowing clean context reuse without propagating noisy targets.}}
    \label{fig:huge_attention_information_flow}
\end{figure}

\paragraph{Public context and private targets.}
\figref{fig:huge_attention_information_flow} illustrates which information
later predictions can access and how recurrent states are retained or
discarded. For the sequence $[T_1,N_1,C_1,T_2,N_2]$, $T$ denotes text,
$N$ a noisy visual target, and $C$ clean visual context. $T_2$ and $N_2$
can use $C_1$ but not $N_1$, while $N_1$ cannot access its later clean
counterpart $C_1$.

In full-attention layers, each block can attend to preceding text and
clean visual blocks, which form the public context. Noisy targets remain
private: they can read this context, but no other block can attend to
them. Within each block, text uses causal attention, while clean visual
blocks and noisy targets use bidirectional attention.

Linear-attention layers implement this separation by controlling which
state updates are retained. Text and clean visual blocks update the state
passed to subsequent blocks. A noisy target starts from the preceding
public state and performs local updates to produce its outputs, but those
updates are discarded afterward. Recurrence remains causal within every
block. Let $\mathbf{S}^{\mathrm{pub}}_b$ denote the public state before
block $B_b$, and let
$\mathcal{R}(B_b,\mathbf{S}^{\mathrm{pub}}_b)$ denote the state obtained
after processing that block. The public state is updated as
\begin{equation}
    \mathbf{S}^{\mathrm{pub}}_{b+1} =
    \begin{cases}
        \mathcal{R}(B_b,\mathbf{S}^{\mathrm{pub}}_b),
            & B_b\text{ is public},\\
        \mathbf{S}^{\mathrm{pub}}_b,
            & B_b\text{ is private}.
    \end{cases}
    \label{eq:huge_state_commit}
\end{equation}

\subsection{Joint Understanding and Generation Objectives}
\label{sec:huge_objectives}

The backbone is trained jointly with autoregressive text prediction and
visual flow matching. For supervised assistant-text positions
$\mathcal{T}$, the language objective is
\begin{equation}
    \mathcal{L}_{\mathrm{text}}
    = -\mathbb{E}\!\left[
       \frac{1}{|\mathcal{T}|}\sum_{i\in\mathcal{T}}
       \log p_{\theta}(w_i\mid\mathcal{C}_{<i})\right],
    \label{eq:huge_text_loss}
\end{equation}
where $\mathcal{C}_{<i}$ is the multimodal context visible to position $i$
under the information-flow rules described above.

Visual generation uses flow matching~\citep{lipman2023flowmatching}.
Given clean latents $\mathbf{z}_0$, noise
$\boldsymbol{\epsilon}\sim\mathcal{N}(\mathbf{0},\mathbf{I})$, and a
sampled noise level $t\in(0,1)$, we construct
\begin{equation}
    \mathbf{z}_t=(1-t)\mathbf{z}_0+t\boldsymbol{\epsilon},
    \qquad
    \mathbf{u}_t=\boldsymbol{\epsilon}-\mathbf{z}_0.
    \label{eq:huge_flow_path}
\end{equation}
The output adapter predicts the velocity from target-token hidden states,
with the objective
\begin{equation}
    \mathcal{L}_{\mathrm{vis}}
    = \mathbb{E}_{\mathbf{z}_0,\boldsymbol{\epsilon},t}
      \left[\left\|
        \widehat{\mathbf{u}}_{\theta}(\mathbf{z}_t,t;\mathcal{C})
        -\mathbf{u}_t
      \right\|_2^2\right],
    \label{eq:huge_flow_loss}
\end{equation}
where $\mathcal{C}$ contains the public context preceding the target.
The joint objective is
\begin{equation}
    \mathcal{L}=\lambda_{\mathrm{text}}\mathcal{L}_{\mathrm{text}}
+\lambda_{\mathrm{vis}}\mathcal{L}_{\mathrm{vis}}
    \label{eq:joint_loss}
\end{equation}
At inference, sampling integrates the predicted velocity field from
$t=1$ (noise) to $t=0$ (data).


\subsection{Interleaved Generation with Persistent Context}
\label{sec:huge_interleaved}

Interleaved generation models text and image outputs as an ordered
sequence of blocks $B_1,\ldots,B_K$:
\begin{equation}
    p_{\theta}(B_1,\ldots,B_K)
    = \prod_{k=1}^{K}p_{\theta}(B_k\mid B_{<k}),
    \label{eq:huge_interleaved_factorization}
\end{equation}
Text blocks are predicted autoregressively, while visual blocks use
conditional flow matching.

During interleaved training, an image that conditions subsequent outputs
appears in two roles: a private noisy target for visual supervision and
clean public context for later predictions. A two-image training sequence
takes the form
\begin{equation}
    \bigl[
       T_1,\ N_1(t_1),\ C_1,\ T_2,\ N_2(t_2)
    \bigr],
    \label{eq:huge_interleaved_example}
\end{equation}
where $T_k$ denotes text, $N_k(t_k)$ a noisy visual target, and $C_1$
the clean representation of the image targeted by $N_1$.
Information flow follows the block rules in
Sec.~\ref{sec:huge_hybrid_attention}.
The clean context $C_1$ is encoded from the observed image through teacher
forcing. Visual supervision applies only to the noisy targets; the clean
context carries no denoising loss.

At inference, clean context comes from completed model outputs.
The public prefix remains fixed throughout the sampling of each image,
and intermediate denoising updates are discarded. Once sampling finishes,
the completed image is encoded into clean visual representations and
appended to the public context. Full-attention layers cache the resulting
public keys and values, while linear-attention layers retain the recurrent
state and convolution history updated by these clean representations.
The inference interface accepts interleaved text--image conditions and
produces one image per call, optionally preceded by generated text.
For multi-image decoding, a controller repeatedly samples images,
inserts their clean representations into context, and determines
when to stop.
\section{Data Preparation}
\label{sec:data}

\methodname{} combines data for multimodal understanding, image generation, image
editing, text-only learning, and interleaved prediction during its main training phase. We organize these
sources by their supervision structure, while adjusting resolution and
sampling separately across training stages. The curriculum comprises
\emph{Joint Multimodal Foundation}, \emph{Progressive Resolution Scaling},
\emph{High-Resolution Refinement}, and \emph{Supervised Capability Refinement
(SFT)}. The first three stages establish joint modeling and extend visual
generation to resolutions up to 2K. SFT targets more accurate understanding,
better instruction following, and improved visual generation and editing.
Separately, we collect video generation and editing data to further fine-tune the model, as an exploratory post-SFT effort to develop video capabilities.

\paragraph{Data scale and coverage.}
The selected data sources for main training contain 159.66M generation samples and 213.39M understanding samples. The latter comprise 99.98M captioning samples, 97.90M visual knowledge QA samples, 9.99M OCR samples, 4.51M grounding samples, and 1.00M GUI samples. These sources also include 97.90M text-only samples, 0.50M editing examples and 624.8K interleaved sequences. This coverage supports supervision for both general semantics and specialized visual skills, while interleaved examples retain transitions between images. The total understanding count sums the five listed task groups; it is not an independently deduplicated image count. Cross-group overlap and actual use during training must be counted separately. The video construction source list in \secref{sec:huge_interleaved_data} is an upstream source pool and is not added to these listed training-source counts.

\begin{figure}[t]
    \centering
    \includegraphics[width=\linewidth]{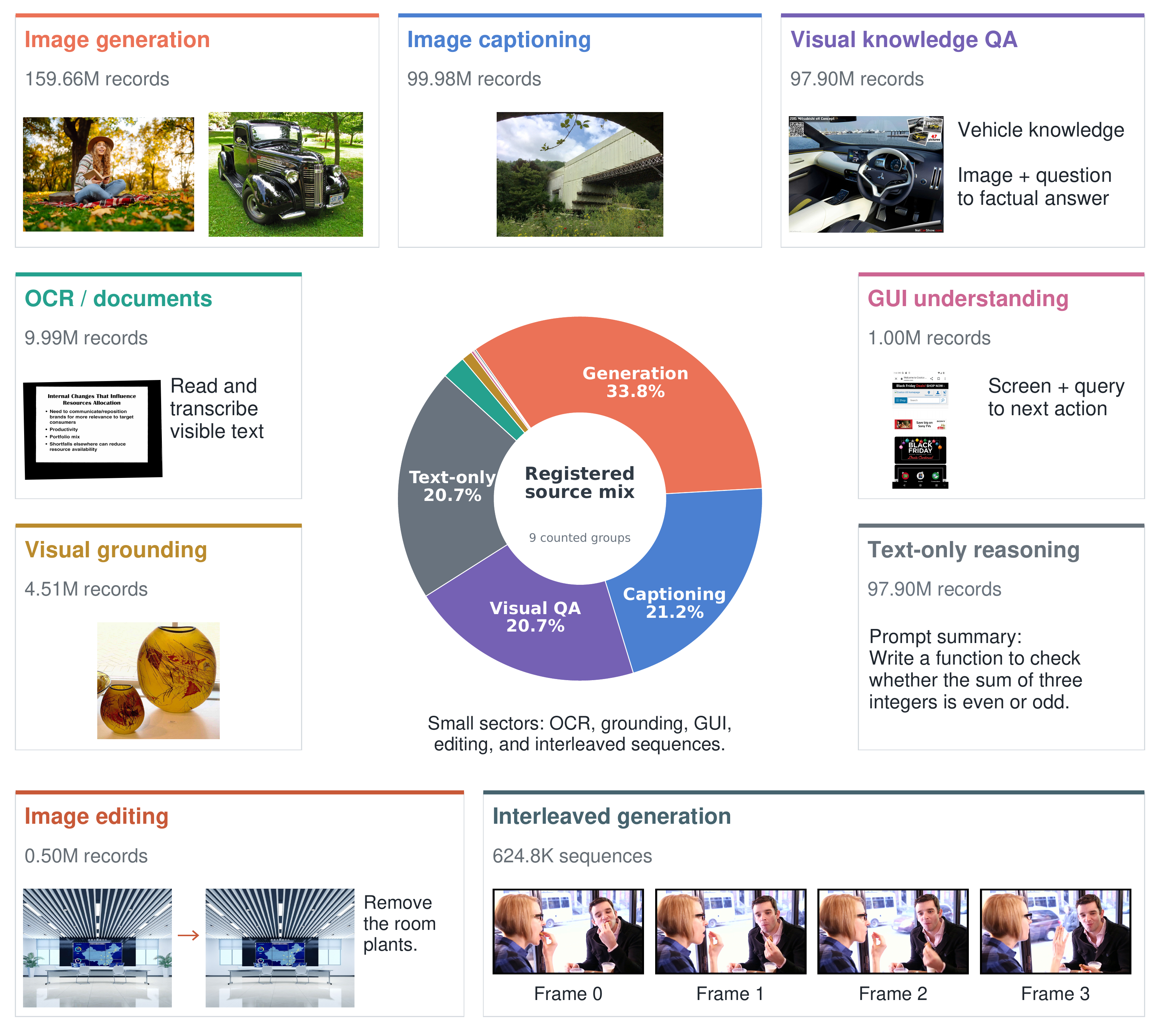}
    \caption{\textbf{Training-data composition and task examples.} \revised{Source counts and examples summarize the task coverage of joint training, from multimodal understanding to image generation, editing, and interleaved prediction.}}

    \label{fig:huge_data_registry}
\end{figure}

\figref{fig:huge_data_registry} pairs source dataset sizes with training
examples from all nine task groups. Image generation, captioning, and
visual knowledge QA account for the largest counted source pools. Editing
examples preserve source--target relations, while interleaved examples
retain an ordered sequence of frames.

\begin{figure}[t]
    \centering
    \includegraphics[width=\linewidth]{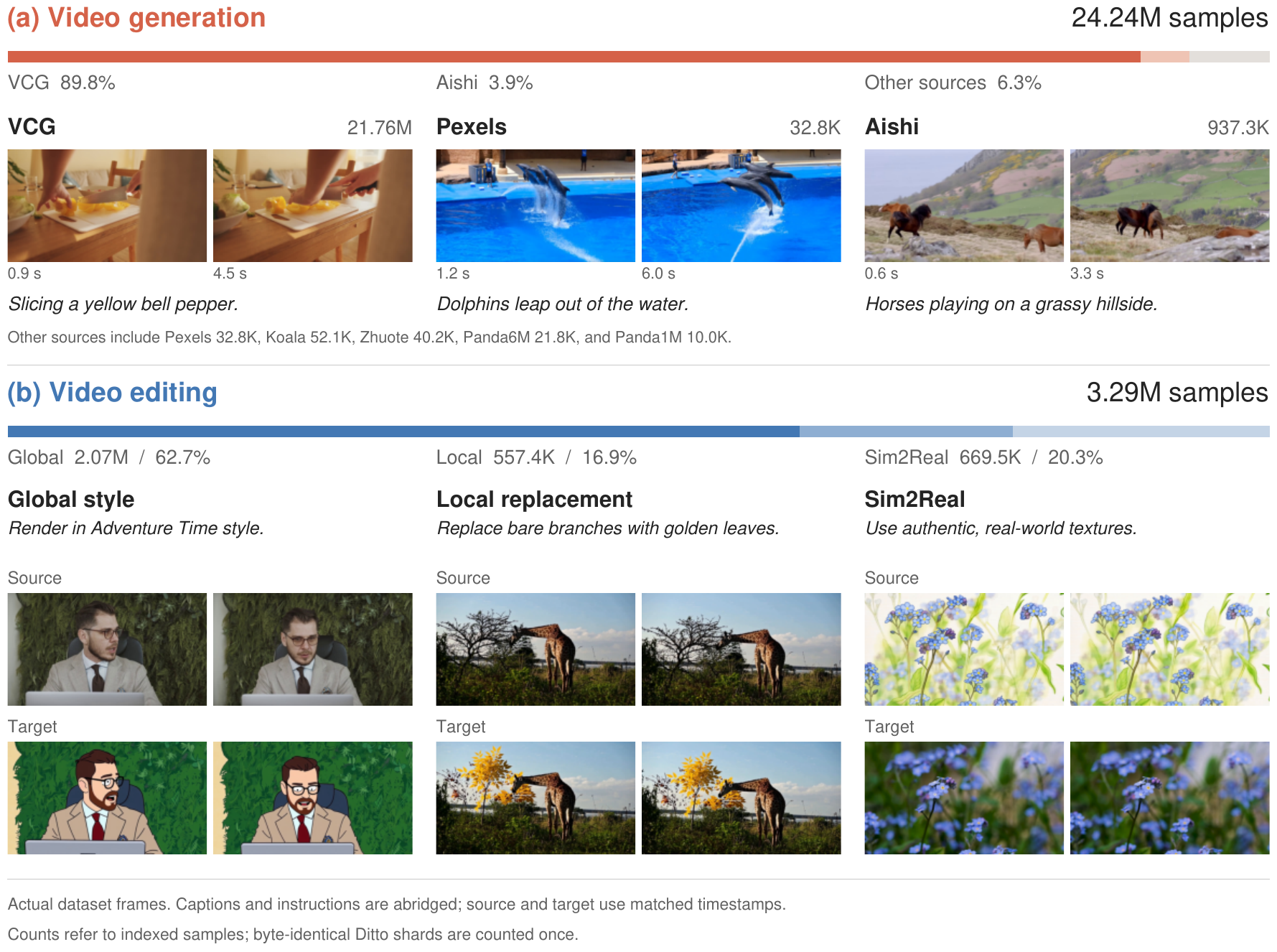}
    \caption{\textbf{Video-data composition and examples.} Generation sources (a) and aligned editing pairs (b) provide 24.24M and 3.29M samples, respectively, for post-SFT video adaptation. Counts refer to indexed samples; byte-identical Ditto shards are counted once.}
    \label{fig:video_data_statistics}
\end{figure}

\revised{{}Beyond these main-training sources, the data for training video capabilities comprise 24.24M generation samples and 3.29M editing samples. \figref{fig:video_data_statistics} summarizes their composition and shows representative examples.
VCG accounts for 89.8\% of the video generation samples, complemented by Aishi and other sources.
The video editing data cover global edits (2.07M), local edits (557.4K), and simulation-to-real (Sim2Real) transformations (669.5K), with temporally aligned source--target clips.
These data support exploratory post-SFT fine-tuning; their collection and filtering are described in \secref{sec:huge_data_video}.}

\subsection{Multimodal Understanding}
\label{sec:huge_data_understanding}

The understanding mixture comprises image captioning, knowledge-oriented
visual question answering, OCR, GUI understanding, and visual grounding.
These correspond to separate sampling groups, allowing broad semantic
supervision and specialized visual tasks to be balanced independently.
Captioning supplies descriptions of image content, while visual questions
connect image evidence to a requested answer. The specialized groups cover
complementary aspects of visual understanding:
\begin{itemize}[leftmargin=*,noitemsep,topsep=3pt]
    \item \textbf{OCR and document content.} Image-conditioned text
    supervision emphasizes reading visible text and relating it to the
    surrounding visual context.
    \item \textbf{GUI understanding.} Interface images and associated
    instructions connect language to visual elements in structured screens.
    \item \textbf{Visual grounding.} Localization-oriented examples connect
    referring expressions to spatially specified visual content.
    \item \textbf{Visual knowledge and question answering.} Image-grounded
    questions and responses broaden semantic coverage beyond captioning.
\end{itemize}

Images are processed by the understanding frontend with
aspect-ratio-aware resizing, patch/merge alignment, and a pixel budget.
The lower-resolution mixtures use a target long-edge setting of 384;
expanded mixtures use a target setting of 980, aligned to 960 by the
spatial factor of 32. These transforms are independent of the generation
resolution. Conversation roles and supervision masks identify the assistant
text to predict, while the image remains an input condition.

\subsection{Image Generation}
\label{sec:huge_data_generation}

The generation stream provides images and associated text conditions through
a dedicated image-data reader. Each example is converted into conditioning
text and a target VAE latent sequence. The target receives flow-matching
supervision, while text provides its semantic condition. Generation is
sampled separately from understanding and editing, enabling the mixture to
control their relative training exposure.

Generation data follow the resolution curriculum rather than a single
fixed-size preprocessing rule. The foundation setting uses 256-scale
images; resolution scaling introduces a 512-scale setting; high-resolution
refinement extends the range through 1024-scale settings and up to 2K.
Images retain their aspect-ratio structure subject to resize, stride, and
pixel-budget constraints, so image dimensions vary within each scale.
Semantic image inputs retain their own understanding-side transform.

\subsection{Image Editing}
\label{sec:huge_data_editing}

The editing stream contains an instruction, a source image, and a target
image. The source is retained as public visual context through semantic
features and, where configured, clean VAE latents. The target is represented
as a private noisy latent block carrying the visual loss. The sequence plan
keeps the source and target roles explicit, preserving the direction of
the transformation. Generation-side resize settings apply to editing
images, while semantic reference features use the understanding transform.

\subsection{Interleaved Data}
\label{sec:huge_interleaved_data}

\paragraph{Sources and sequence construction.}
The interleaved stream builds related text--image sequences from seven
listed visual-sequence collections. Its adapters support alternating
text--image examples and chained image edits. The interleave-enabled joint
mixtures use the chained-edit format, retaining transition text and target
captions. Each chain begins with an opening prompt and source image, then
specifies successive transitions and visual targets:
\begin{equation}
  (P,I_0)\ \longrightarrow\
  [\Delta_1,D_1,I_1,\Delta_2,D_2,I_2,\ldots],
  \label{eq:huge_data_edit_chain}
\end{equation}
where $\Delta_k$ is a transition instruction, $D_k$ is the target caption,
and $I_k$ is the corresponding target image. Reasoning fields used during data construction are omitted from the training response.

\begin{wraptable}{r}{0.43\textwidth}
\vspace{-\intextsep}
\centering
\small
\setlength{\tabcolsep}{4pt}
\captionsetup{font=small}
\begin{tabular}{lr}
\toprule
Source identifier & Source samples \\
\midrule
Panda1M & 10.02 K \\
Panda6M & 21.82 K\\
Koala clips & 52.08 K \\
Editing & 1.38 M \\
VCG & 21.76 M \\
Pexel & 32.79 K \\
Aishi & 937.31 K \\
Zhuote & 40.19 K\\
\midrule
Total & 24.24 M \\
\bottomrule
\end{tabular}
\caption{\textbf{Video annotation sources.} Source samples before filtering and annotation acceptance.}
\label{tab:huge_interleave_inventory}
\end{wraptable}

\paragraph{Video sources and annotation settings.}
Our video annotation effort draws on existing text-to-video, egocentric,
and interaction-rich video collections. The batch annotation source list contains
24.24M source samples (\tabref{tab:huge_interleave_inventory}).

\paragraph{Event-driven keyframes and transition annotation.}
The construction protocol asks a video-language model to select three to
five timestamps marking meaningful visual changes, such as contact,
grasping, release, object movement, or a camera-view change. Timestamps
must be strictly increasing; the first is normally zero unless the opening
frames are invalid. The batch configuration samples video observations at
1 FPS. This sampling rate controls the annotation input, whereas the
selected keyframes follow event boundaries rather than a fixed output
interval. The annotation model supplies descriptions and transition instructions
for the selected frames.

The first instruction describes the initial static scene. Subsequent
instructions combine a target-state description $D_k$ with a transition
$\Delta_k$ from the preceding keyframe. The protocol emphasizes subject
pose, object position and orientation, contact relations, and changes in
scene or camera viewpoint, including contact changes, object motion, and
final placement. Annotation must
remain grounded in visible evidence rather than inferred intentions or
unobserved outcomes. Timestamp references are resolved to extracted frames,
producing alternating user-text and assistant-image messages. This yields
\eqref{eq:huge_data_edit_chain} after adaptation to the chained-edit format.
The visual targets are observed video frames; the annotation model supplies
their descriptions and transition instructions.

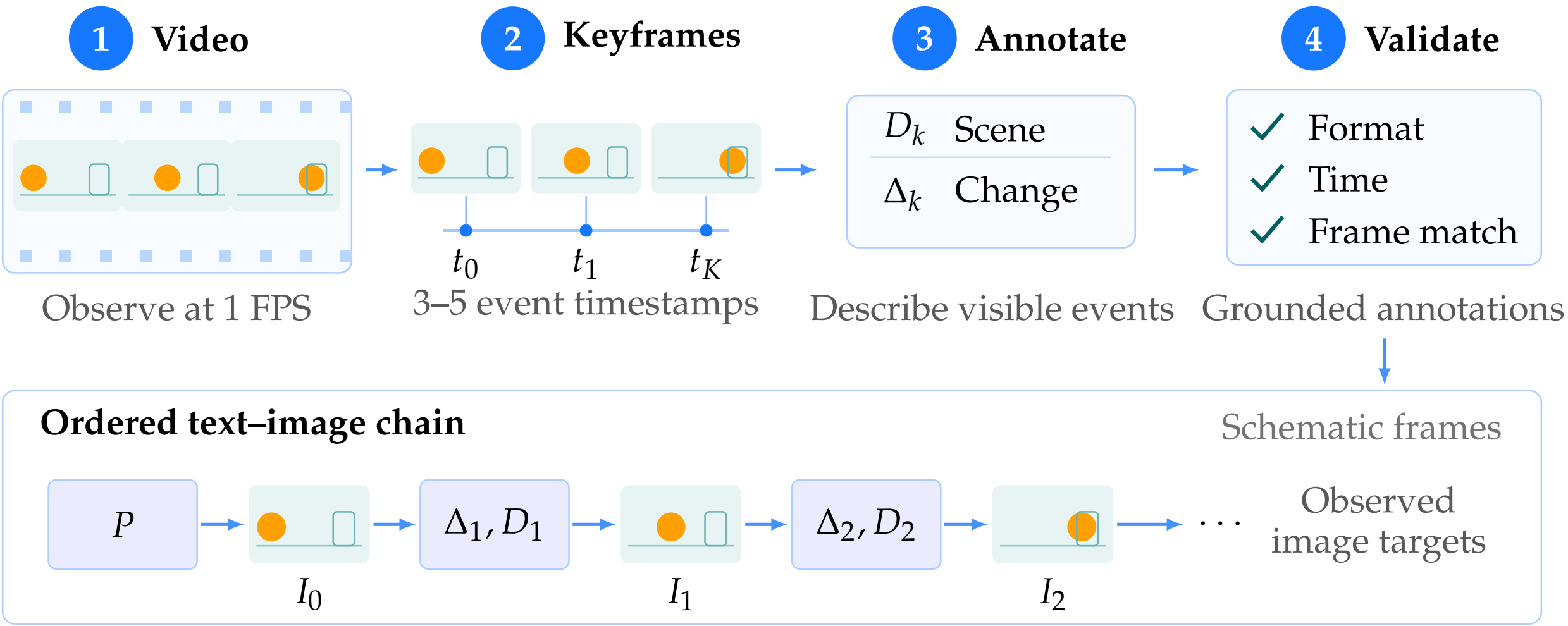
\begin{figure}[t]
    \centering
    \resizebox{\linewidth}{!}{
\begin{tikzpicture}[x=1cm,y=1cm,font=\small,
 arr/.style={-{Latex[length=1.5mm]},thick,draw=title_blue!80,shorten <=.7pt,shorten >=.7pt},
 card/.style={draw=title_blue!35,fill=title_blue!3,rounded corners=3pt},
 label/.style={font=\small,align=center},
 pics/scene/.style args={#1}{code={
   \fill[teal!9,rounded corners=2pt] (-.48,-.31) rectangle (.48,.31);
   \draw[teal!45] (-.42,-.17)--(.42,-.17);
   \fill[orange!75!yellow] (#1,-.02) circle (.115);
   \draw[teal!60,rounded corners=1pt] (.19,-.17) rectangle (.36,.10);
 }}]
\foreach \x/\n/\t in {1.5/1/Video,5.1/2/Keyframes,8.7/3/Annotate,12.1/4/Validate}{
 \node[circle,fill=title_blue,text=white,inner sep=3pt,font=\small\bfseries] at (\x-.64,0) {\n};
 \node[anchor=west,font=\small\bfseries] at (\x-.33,0) {\t};
}
\draw[card] (0,-2.05) rectangle (3.05,-.45);
\foreach \x/\pos in {.57/-.30,1.52/-.08,2.47/.23}{
 \pic at (\x,-1.20) {scene={\pos}};
}
\foreach \x in {.15,.50,.85,1.20,1.55,1.90,2.25,2.60,2.95}{
 \fill[title_blue!30] (\x,-.65) rectangle ++(.10,.10);
 \fill[title_blue!30] (\x,-1.95) rectangle ++(.10,.10);
}
\node[label,text=black!65] at (1.52,-2.35) {Observe at 1 FPS};
\draw[title_blue!40,thick] (3.85,-1.68)--(6.35,-1.68);
\foreach \x/\pos/\t in {4.05/-.30/t_0,5.10/-.08/t_1,6.15/.23/t_K}{
 \pic at (\x,-1.05) {scene={\pos}};
 \fill[title_blue] (\x,-1.68) circle (.055);
 \draw[title_blue!50] (\x,-1.38)--(\x,-1.63);
 \node[font=\footnotesize] at (\x,-1.97) {$\t$};
}
\node[label,text=black!65] at (5.1,-2.35) {3--5 event timestamps};
\draw[card] (7.38,-1.83) rectangle (9.90,-.50);
\node[anchor=west,font=\small\bfseries] at (7.57,-.79) {$D_k$};
\node[anchor=west,font=\small] at (8.20,-.79) {Scene};
\draw[title_blue!25] (7.58,-1.04)--(9.69,-1.04);
\node[anchor=west,font=\small\bfseries] at (7.57,-1.36) {$\Delta_k$};
\node[anchor=west,font=\small] at (8.20,-1.36) {Change};
\node[label,text=black!65] at (8.65,-2.35) {Describe visible events};
\draw[card] (10.70,-1.98) rectangle (13.45,-.45);
\foreach \y/\t in {-.78/Format,-1.23/Time,-1.68/Frame match}{
 \draw[teal!75!black,line width=1pt] (10.92,\y)--++(.09,-.09)--++(.18,.21);
 \node[anchor=west,font=\small] at (11.29,\y) {\t};
}
\node[label,text=black!65] at (12.07,-2.35) {Grounded annotations};
\draw[arr] (3.15,-1.15)--(3.49,-1.15);
\draw[arr] (6.73,-1.15)--(7.15,-1.15);
\draw[arr] (10.04,-1.15)--(10.49,-1.15);
\draw[card,fill=white] (0,-5.12) rectangle (13.45,-3.08);
\node[anchor=west,font=\small\bfseries] at (.20,-3.39) {Ordered text--image chain};
\node[anchor=east,font=\footnotesize,text=black!55] at (13.23,-3.39) {Schematic frames};
\draw[arr] (12.08,-2.60)--(12.08,-3.06);
\foreach \x/\t/\id in {1.05/P/p,4.30/{\Delta_1,D_1}/d1,7.55/{\Delta_2,D_2}/d2}{
 \node[draw=title_blue!35,fill=blue!8,rounded corners=2pt,minimum width=1.3cm,minimum height=.78cm] (\id) at (\x,-4.25) {$\t$};
}
\foreach \x/\pos/\t/\id in {2.68/-.30/I_0/i0,5.93/-.08/I_1/i1,9.18/.23/I_2/i2}{
 \pic[scale=1.10] at (\x,-4.25) {scene={\pos}};
 \node[minimum width=1.06cm,minimum height=.69cm,inner sep=0pt] (\id) at (\x,-4.25) {};
 \node[font=\footnotesize] at (\x,-4.86) {$\t$};
}
\foreach \a/\b in {p/i0,i0/d1,d1/i1,i1/d2,d2/i2}{\draw[arr] (\a.east)--(\b.west);}
\draw[arr] (i2.east)--(10.35,-4.25);
\node at (10.67,-4.25) {$\cdots$};
\node[align=center,font=\small,text=black!65] at (12.03,-4.25) {Observed\\image targets};
\end{tikzpicture}}
    \caption{\textbf{Video-based interleaved data construction.} \revised{Event-selected video frames are paired with scene descriptions and transition instructions to form grounded text--image training sequences.}}
    \label{fig:huge_interleave_construction}
\end{figure}

\paragraph{Quality screening and annotation validation.}
The planned screening uses resolution, duration, clarity, exposure,
aesthetics, motion, shot boundaries, and overlays to identify unsuitable
clips and split clips longer than 30 seconds.

Before use, annotations must be checked for correct format and timing:
valid role alternation, one target image per assistant turn, valid
media references, strictly increasing in-range timestamps, and consistency
between transition text and the adjacent frames. The image-target
protocol keeps visual-state descriptions and transition instructions tied
to the extracted frames.

\paragraph{Annotation pipeline.}
Video decoding, quality screening, and annotation are separate steps,
allowing source-specific filtering while keeping a common output format.

\paragraph{Structural filtering.}
Samples must be parsed successfully and checked for suitability for
interleaved training. The adapter
also verifies the message structure and requires each successive source
image reference to match the preceding target reference. Non-final target
images are followed by clean representations of the observed image, making
completed visual content available to later text and image prediction.

\subsection{Text-Only Data}
\label{sec:huge_data_text}

The text-only group contains 97.90M samples across 100 source entries covering mathematics,
programming, knowledge, reasoning, general instructions, conversation,
tool use, and long-context tasks. It supplies direct language supervision
alongside visual learning. Reasoning examples belong to this text mixture;
they are distinct from the reasoning used to construct interleaved editing
samples, which is removed before training. During capability refinement, the intended
role of text data is to maintain response quality and instruction-following
coverage while understanding and generation are refined jointly.

\subsection{Video Data}
\label{sec:huge_data_video}

The video data comprise separate groups for video generation and
source-conditioned video editing, used in the exploratory post-SFT
fine-tuning described above. Video generation data are collected from
VCG, Koala, Aishi, Pexels, Panda, and internal sources, while video editing data are drawn from Ditto \citep{bai2026scaling}. We exclude videos longer than eight seconds
and those with excessively low or high original frame rates. We further
filter the data using VLM-based assessments of visible text, aesthetics,
and visual quality. Retained videos are resampled to 16 FPS; for editing,
source and target clips share the same temporal sampling plan.

\subsection{Multimodal Sequence Packing}
\label{sec:huge_data_packing}

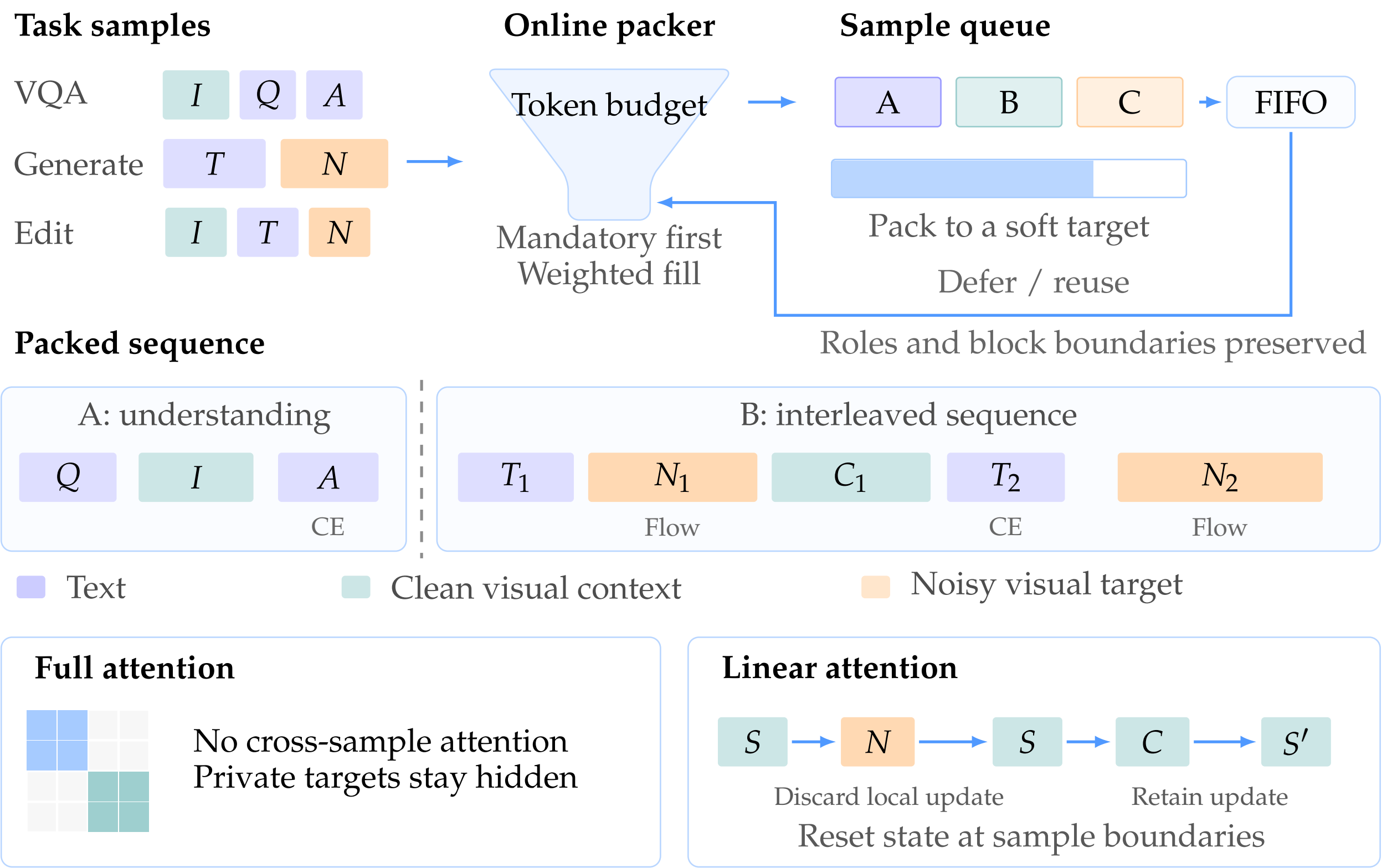
\begin{figure}[t]
    \centering
    \resizebox{\linewidth}{!}{
\begin{tikzpicture}[x=1cm,y=1cm,font=\small,
 arr/.style={-{Latex[length=1.6mm]},thick,draw=title_blue!75,shorten <=1pt,shorten >=1pt},
 panel/.style={draw=title_blue!30,fill=title_blue!2,rounded corners=3pt},
 tok/.style={rounded corners=1pt,minimum height=.48cm,inner sep=2pt,font=\footnotesize},
 txt/.style={tok,fill=blue!13},ctx/.style={tok,fill=teal!20},gen/.style={tok,fill=orange!30}]
\node[anchor=west,font=\small\bfseries] at (0,0) {Task samples};
\foreach \y/\name in {-.65/VQA,-1.32/Generate,-1.99/Edit}{
 \node[anchor=west,font=\footnotesize,text=black!70] at (0,\y) {\name};
}
\node[ctx,minimum width=.65cm] at (1.90,-.65) {$I$};
\node[txt,minimum width=.55cm] at (2.60,-.65) {$Q$};
\node[txt,minimum width=.55cm] at (3.25,-.65) {$A$};
\node[txt,minimum width=1.0cm] at (2.08,-1.32) {$T$};
\node[gen,minimum width=1.05cm] at (3.25,-1.32) {$N$};
\node[ctx,minimum width=.60cm] at (1.90,-1.99) {$I$};
\node[txt,minimum width=.60cm] at (2.60,-1.99) {$T$};
\node[gen,minimum width=.60cm] at (3.30,-1.99) {$N$};
\draw[arr] (3.92,-1.30)--(4.55,-1.30);
\draw[panel,fill=title_blue!7] (4.72,-.40)--(7.14,-.40)--(6.33,-1.47)--(6.33,-1.87)--(5.53,-1.87)--(5.53,-1.47)--cycle;
\node[font=\small\bfseries] at (5.93,0) {Online packer};
\node[font=\footnotesize] at (5.93,-.78) {Token budget};
\node[font=\footnotesize,align=center,text=black!65] at (5.93,-2.25) {Mandatory first\\Weighted fill};
\draw[arr] (7.25,-.72)--(7.80,-.72);
\node[anchor=west,font=\small\bfseries] at (8.05,0) {Sample queue};
\foreach \x/\c/\t in {8.65/blue/A,9.83/teal/B,11.01/orange/C}{
 \node[tok,draw=\c!40,fill=\c!12,minimum width=1.03cm] at (\x,-.72) {\t};
}
\draw[title_blue!40,rounded corners=1pt] (8.10,-1.28) rectangle (11.56,-1.65);
\fill[title_blue!30] (8.11,-1.29) rectangle (10.65,-1.64);
\node[font=\footnotesize,text=black!65] at (9.83,-1.96) {Pack to a soft target};
\node[panel,minimum width=1.25cm,minimum height=.50cm,font=\footnotesize] (fifo) at (12.58,-.72) {FIFO};
\draw[arr] (11.65,-.72)--(fifo.west);
\draw[arr] (fifo.south)--(12.58,-2.80)--(7.55,-2.80)--(7.55,-1.70)--(6.35,-1.70);
\node[font=\footnotesize,text=black!60] at (10.07,-2.48) {Defer / reuse};
\node[anchor=west,font=\small\bfseries] at (0,-3.10) {Packed sequence};
\node[anchor=east,font=\footnotesize,text=black!60] at (13.45,-3.10) {Roles and block boundaries preserved};
\draw[panel] (0,-3.50) rectangle (3.95,-5.10);
\draw[panel] (4.25,-3.50) rectangle (13.45,-5.10);
\node[font=\footnotesize,text=black!70] at (1.98,-3.80) {A: understanding};
\node[font=\footnotesize,text=black!70] at (8.85,-3.80) {B: interleaved sequence};
\node[txt,minimum width=.95cm] at (.65,-4.38) {$Q$};
\node[ctx,minimum width=1.12cm] at (1.90,-4.38) {$I$};
\node[txt,minimum width=.98cm] at (3.19,-4.38) {$A$};
\node[txt,minimum width=1.13cm] at (5.02,-4.38) {$T_1$};
\node[gen,minimum width=1.65cm] at (6.55,-4.38) {$N_1$};
\node[ctx,minimum width=1.55cm] at (8.29,-4.38) {$C_1$};
\node[txt,minimum width=1.15cm] at (9.80,-4.38) {$T_2$};
\node[gen,minimum width=2.00cm] at (11.89,-4.38) {$N_2$};
\draw[black!45,dashed,thick] (4.10,-3.43)--(4.10,-5.17);
\node[font=\scriptsize,text=black!60] at (3.19,-4.86) {CE};
\foreach \x/\t in {6.55/Flow,9.8/CE,11.89/Flow}{\node[font=\scriptsize,text=black!60] at (\x,-4.86) {\t};}
\foreach \x/\c/\t in {.15/blue/Text,3.32/teal/Clean visual context,8.39/orange/Noisy visual target}{
 \fill[\c!20,rounded corners=1pt] (\x,-5.56) rectangle ++(.28,.22);
 \node[anchor=west,font=\footnotesize,text=black!70] at (\x+.36,-5.45) {\t};
}
\draw[panel,fill=white] (0,-8.18) rectangle (6.43,-5.94);
\draw[panel,fill=white] (6.70,-8.18) rectangle (13.45,-5.94);
\node[anchor=west,font=\small\bfseries] at (.20,-6.23) {Full attention};
\node[anchor=west,font=\small\bfseries] at (6.90,-6.23) {Linear attention};
\foreach \i in {0,...,3}{\foreach \j in {0,...,3}{
 \draw[fill=black!3,draw=white] (.25+.30*\j,-6.65-.30*\i) rectangle ++(.29,-.29);
}}
\foreach \i in {0,1}{\foreach \j in {0,1}{
 \fill[title_blue!38] (.25+.30*\j,-6.65-.30*\i) rectangle ++(.29,-.29);
 \fill[teal!38] (.85+.30*\j,-7.25-.30*\i) rectangle ++(.29,-.29);
}}
\node[align=left,anchor=west,font=\footnotesize] at (1.75,-7.16) {No cross-sample attention\\Private targets stay hidden};
\node[ctx,minimum width=.68cm] (s0) at (7.33,-6.96) {$S$};
\node[gen,minimum width=.72cm] (n) at (8.55,-6.96) {$N$};
\node[ctx,minimum width=.68cm] (s1) at (10.01,-6.96) {$S$};
\node[ctx,minimum width=.72cm] (c) at (11.23,-6.96) {$C$};
\node[ctx,minimum width=.68cm] (s2) at (12.63,-6.96) {$S'$};
\draw[arr] (s0)--(n);\draw[arr] (n)--(s1);\draw[arr] (s1)--(c);\draw[arr] (c)--(s2);
\node[font=\scriptsize,text=black!65] at (8.66,-7.52) {Discard local update};
\node[font=\scriptsize,text=black!65] at (11.79,-7.52) {Retain update};
\node[font=\footnotesize,text=black!65] at (10.05,-7.91) {Reset state at sample boundaries};
\end{tikzpicture}}
    \caption{\textbf{Multimodal sequence packing.} \revised{Online packing combines task-specific samples within a token budget while preserving sample isolation and the public/private boundaries required by hybrid attention.}}
    \label{fig:huge_sequence_packing}
\end{figure}

\figref{fig:huge_sequence_packing} illustrates how task-specific samples
become a packed sequence while retaining the boundaries needed by hybrid
attention and the two prediction objectives.

\paragraph{Unified sequence assembly.}
The task adapters convert samples from different tasks into a common sequence
plan containing text roles, media identities, attention-block types, and
supervision locations. Packing concatenates multiple examples while
retaining their individual sequence boundaries and within-example block
structure. Text labels and visual targets are gathered through separate
loss indices, so conditioning content can share the input sequence without
receiving a prediction loss. For interleaved samples, the complete sequence
of transitions and images remains within the same example; packing does
not turn successive editing steps into unrelated image--text pairs.

\paragraph{Online token-budget packing.}
The loader starts each pack by inserting eligible examples from mandatory
task groups. It then selects additional examples according to normalized
group weights, using modality-aware token estimates to account for text
and visual content. A soft token target determines when a pack is normally
produced, while a per-example threshold rejects oversized samples and a
separate packed-token budget guides additional insertions. Examples that
do not fit can be stored in a fixed-size buffer for later use. The loader reuses buffered
examples in first-in, first-out order when the current pack is below a
configured occupancy threshold. This online policy accommodates the
changing token costs of image resolution, aspect ratio, and interleaved
sequence length without requiring a fixed number of images per batch.
The final accepted example or mandatory insertion can exceed the nominal
token budget.
Reference budgets are listed in \tabref{tab:huge_training_parameters}.

\paragraph{Sample isolation and hybrid-attention metadata.}
Packed sample lengths prevent attention or recurrent state from crossing
between independent examples. Within an example, block lengths and
attention modes preserve the conditioning relationships described in
\secref{sec:huge_hybrid_attention}. In particular, a noisy visual target
remains private, whereas an observed clean image may contribute public
context for subsequent prediction. The same sequence layout supplies
visibility metadata to full-attention layers and state-retention boundaries
to linear-attention layers. These layers retain their respective attention
behavior. Variable-length execution can avoid padding every example to a
common text--visual sequence length; visual tensor padding and
backend-specific alignment can still be required.

\paragraph{Data mixture statistics.}
Group weights govern sampling decisions, rather than final sample, token,
or compute percentages. Mandatory insertion and buffering further affect
the actual mixture. We therefore distinguish configured weights from
consumed examples and tokens when describing data allocation across
stages. SFT uses the same packing interfaces, with task selection and
weighting focused on capability refinement.

\begin{figure}[t]
    \centering
    \includegraphics[width=0.94\linewidth,trim=0 36bp 0 0,clip]{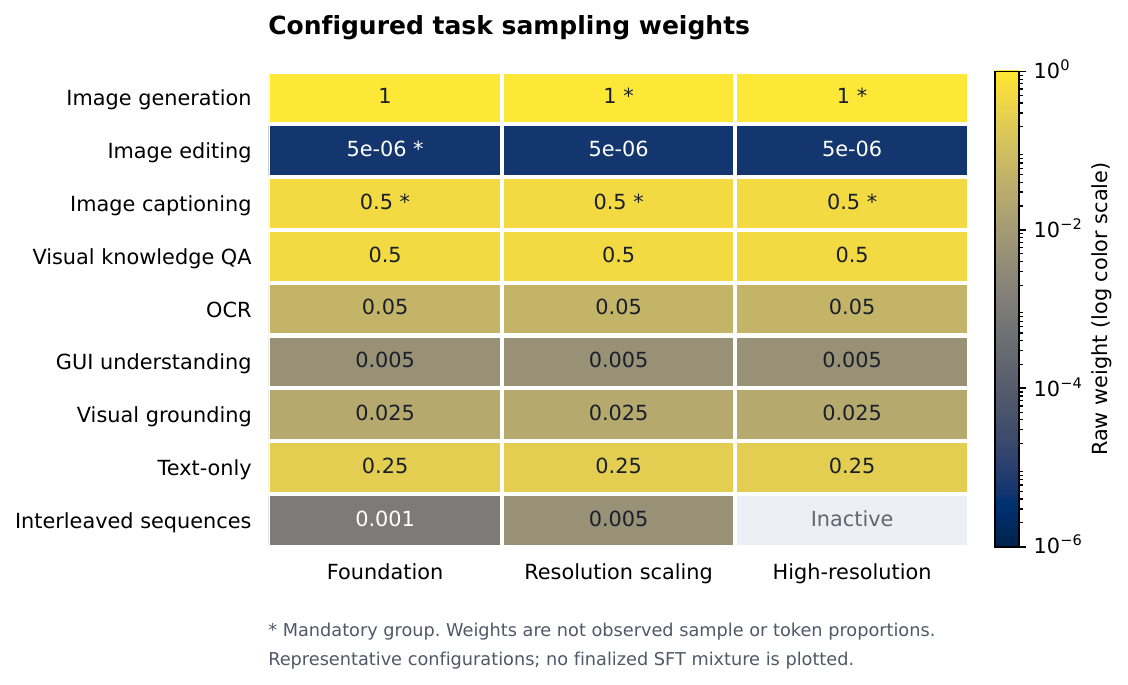}
    \caption{\textbf{Data sampling across training stages.} \revised{Representative task weights are shown for three pre-SFT stages; asterisks mark mandatory groups. These weights do not represent observed sample or token proportions.}}

    \label{fig:huge_data_weights}
\end{figure}

\figref{fig:huge_data_weights} compares sampling weights across three
training stages. Required task groups are included in each pack regardless
of their sampling weight. These weights do not directly determine the
sample or token proportions used during training. Colors encode weights
on a logarithmic scale; asterisks identify required groups and gray cells
identify inactive groups. The high-resolution example uses a 1024-scale
configuration.

\begin{table}[t]
    \centering
    \small
    \setlength{\tabcolsep}{5pt}
    \renewcommand{\arraystretch}{1.15}
    \begin{tabularx}{\linewidth}{@{}p{0.07\linewidth} p{0.26\linewidth} p{0.16\linewidth} X@{}}
        \toprule
        \textbf{Stage} & \textbf{Name} & \textbf{Generation scale} & \textbf{Primary objective} \\
        \midrule
        I & Joint Multimodal Foundation & 256-scale & Establish joint text, visual understanding, and latent-generation modeling. \\
        II & Progressive Resolution Scaling & 512-scale & Expand spatial detail while continuing multimodal supervision. \\
        III & High-Resolution Refinement & 1024-scale to 2K & Extend visual generation to longer target sequences and finer detail. \\
        IV & Supervised Capability Refinement (SFT) & Task-dependent & Refine understanding accuracy, instruction following, generation quality, and editing consistency. \\
        \bottomrule
    \end{tabularx}
    \caption{\textbf{Four-stage training curriculum.} Resolution labels describe
    training scales rather than fixed square image sizes.}
    \label{tab:huge_training_stages}
\end{table}

\section{Training Strategy}
\label{sec:model-training}

We organize the \methodname{} curriculum by its modeling objectives:
\emph{Joint Multimodal Foundation}, \emph{Progressive Resolution Scaling},
\emph{High-Resolution Refinement}, and \emph{Supervised Capability Refinement
(SFT)}. This organization separates learning a shared multimodal backbone,
increasing spatial detail, extending high-resolution generation, and
refining task performance. \tabref{tab:huge_training_stages} summarizes the
four stages. The backbone retains the joint text and visual objectives in
\secref{sec:huge_objectives} throughout this design.

\subsection{Joint Multimodal Foundation}
\label{sec:huge_foundation}

The foundation stage trains the hybrid MoT backbone with a low visual token
budget. Text and semantic visual tokens use the understanding expert, while
VAE latent tokens use the generation expert. Language modeling and visual
flow matching provide direct supervision for the two output types. 

The 256-scale setting keeps visual targets relatively short while
establishing interactions between the two experts. The language backbone, vision encoder, and expert
branches and adapters are trained jointly, while the VAE remains frozen.
The stage's purpose is to establish the shared prediction interface before
increasing the spatial sequence length.

\subsection{Progressive Resolution Scaling}
\label{sec:huge_resolution_scaling}

The resolution-scaling stage increases generation to a 512-scale setting
while retaining the same token roles and loss definitions. More latent
patches represent each image, allowing training to cover finer spatial
detail. Understanding inputs have an independent resolution budget, so
higher-resolution generation does not require every semantic condition to
expand at the same rate.

Interleaved sequences can continue to contribute text and visual targets in
this stage. Clean representations of earlier images remain public context,
while noisy prediction blocks remain private. This exposes hybrid attention
to both longer visual blocks and dependencies across multiple outputs.

\begin{table}[t]
    \centering
    \small
    \setlength{\tabcolsep}{4pt}
    \renewcommand{\arraystretch}{1.15}
    \begin{tabularx}{\linewidth}{@{}l *{4}{>{\centering\arraybackslash}X}@{}}
        \toprule
        \textbf{Parameter} & \textbf{Foundation} & \textbf{Resolution scaling} & \textbf{High-resolution}$^{a}$ & \textbf{SFT}$^{b}$ \\
        \midrule
        Learning rate & $10^{-4}$ & $10^{-4}$ & $10^{-4}$ & $10^{-4}$ \\
        LR schedule & Constant & Constant & Constant & Constant \\
        Warmup updates & 2,500 & 2,500 & 2,500 & 500 \\
        EMA decay & 0.9999 & 0.9999 & 0.9999 & 0.995 \\
        Timestep shift & 1.0 & 1.0 & 1.0 & 4.0 \\
        Text / visual loss weights & 1 / 1 & 1 / 1 & 0.25 / 1 & 0.25 / 1 \\
        Packed-token target & 16,384 & 10,240 & 16,384 & 10,240 \\
        Per-sample token limit & 26,384 & 16,384 & 26,384 & 16,384 \\
        Packed-token limit & 46,864 & 36,864 & 46,864 & 36,864 \\
        \bottomrule
    \end{tabularx}
    \caption{\textbf{Reference optimization settings by training stage.}
    The packed-token target applies to one packed sequence, not a global batch.
    $^{a}$High-resolution values refer to the 1024-scale setting.
    $^{b}$SFT settings are preliminary.
    All learning-rate schedules are constant after warmup.}

    \label{tab:huge_training_parameters}
\end{table}

\subsection{High-Resolution Refinement}
\label{sec:huge_high_resolution}

High-resolution refinement extends generation through 1024-scale settings
and up to 2K. The objective remains conditional flow matching, with
understanding and language supervision retained in the joint training
design. The increased latent-token count makes memory use and packed
sequence composition more sensitive to aspect ratio and the number of
visual targets per example.

The 1024-scale reference setting assigns coefficients of
$0.25$ and $1.0$ to the text and visual losses, respectively. These are loss
multipliers, not a text-to-image data ratio. It also omits the interleaved
data group, whereas a preceding resolution-scaling configuration includes
that group. High-resolution refinement thus need not preserve every task
at the same sampling rate.

\subsection{Supervised Capability Refinement}
\label{sec:huge_sft}

\tabref{tab:huge_training_parameters} gives preliminary SFT settings. SFT is the final capability-oriented stage \revised{of the main multimodal training curriculum}. Its goal is to improve
understanding and generation together through task-focused supervision,
using the data categories in \secref{sec:data}. The refinement design has
four priorities:
\begin{itemize}[leftmargin=*,noitemsep,topsep=3pt]
    \item \textbf{Understanding accuracy:} answer visual questions,
    recognize text, and ground responses in the supplied image.
    \item \textbf{Generation fidelity:} follow the requested subjects,
    attributes, spatial relations, and text in the image.
    \item \textbf{Editing consistency:} perform the instructed change while
    retaining unrelated reference content.
    \item \textbf{Interleaved coherence:} use preceding visual outputs and
    text consistently when predicting subsequent content.
\end{itemize}

The same text cross-entropy and visual flow-matching objectives support
these tasks. SFT changes the emphasis of data selection and sampling; it
does not require a separate understanding model or visual generator.

\subsection{Optimization}
\label{sec:huge_optimization}

\paragraph{Optimizer and precision.}
Training uses AdamW~\citep{loshchilov2019adamw} with coefficients
$(\beta_1,\beta_2)=(0.9,0.95)$, numerical stability constant $10^{-15}$,
zero weight decay, and gradient-norm clipping at $1.0$. Forward computation
uses bfloat16 mixed precision. The joint-training reference settings use learning
rate $10^{-4}$ with a constant schedule after 2,500 warmup updates and EMA
decay $0.9999$. \tabref{tab:huge_training_parameters} summarizes
the reference settings for all four stages.

\paragraph{Noise-time sampling.}
The visual objective samples a time through a sigmoid transform and then
applies the configured shift $s$:
\begin{equation}
    t'=\frac{s\,t}{1+(s-1)t}.
    \label{eq:huge_training_time_shift}
\end{equation}
The joint configurations set $s=1$, while the SFT reference uses $s=4$.
The latter gives larger noise times for intermediate $t$ under the
convention of \eqref{eq:huge_flow_path}. This parameter changes the training
time distribution, not the number of denoising steps used at inference.




\subsection{Adaptation to Video Generation and Editing}
\label{sec:video_train}

\begin{wraptable}{r}{0.48\textwidth}
    \vspace{-\intextsep}
    \centering
    \small
    \captionsetup{font=small}
    \setlength{\tabcolsep}{4pt}
    \renewcommand{\arraystretch}{1.08}
    \begin{tabular}{@{}lc@{}}
        \toprule
        \textbf{Parameter} & \textbf{Value} \\
        \midrule
        Learning rate & $5\times10^{-5}$ \\
        LR schedule & Constant \\
        Warmup updates & 1,000 \\
        EMA decay & 0.999 \\
        Timestep shift & 1.0 \\
        Text conditioning dropout & 0.1 \\
        VAE / ViT conditioning dropout & 0.3 / 0.3 \\
        Packed-token target & 16,384 \\
        Per-sample token limit & 40,960 \\
        Packed-token limit & 40,960 \\
        \bottomrule
    \end{tabular}
    \caption{\textbf{Optimization settings for video adaptation stage.}
    The understanding branch is frozen.}
    \label{tab:video_training_parameters}
\end{wraptable}

Following SFT, we further fine-tune the model on video generation and
editing data as an exploratory extension of its visual capabilities.
The fine-tuning mixture also retains image generation and editing data.
Video generation, video editing, image generation, and image editing
are sampled at a ratio of $5:1:1:0.1$.
For video generation and editing, the longer side of each frame is
set to 1280 and 768 pixels, respectively, while image generation and editing retain 2K resolution.
Video clips are sampled at 16 FPS and span 3--5.2 seconds.

The understanding branch is frozen during this adaptation. \tabref{tab:video_training_parameters} summarizes the
fine-tuning settings.

\subsection{Training Infrastructure and Efficiency}
\label{sec:huge_infrastructure}

\paragraph{Hybrid sharding and activation memory.}
Training uses FSDP~\citep{zhao2023fsdp} hybrid sharding to distribute parameters, gradients,
and optimizer states across GPUs. Decoder-layer and vision-encoder
activation checkpointing reduce memory use through recomputation~\citep{chen2016checkpointing}.
These methods support training with longer visual sequences.

\paragraph{Attention execution for packed multimodal sequences.}
Variable-length attention backends directly use the packing metadata
in \secref{sec:huge_data_packing}. The full-attention implementation
supports block-mask execution, caches derived block-sparse metadata, and
provides a native grouped-query path that avoids explicitly repeating
key/value heads. Backend selection depends on the head configuration,
requested execution path, and installed kernels. The linear-attention MoT
implementation provides optimized execution paths alongside a scalar
reference implementation. Expert-specific projections and convolution
weights operate with shared public recurrent state and convolution
history. Private target blocks read this context but do not retain their
local updates in public context. Optimizing these operations must preserve both sample
isolation and the public/private state rules; full-attention visibility and
linear-attention recurrence are not interchangeable.

\paragraph{Data delivery and visual encoding.}
The input pipeline prefetches samples and assembles packed batches.
A frozen VAE encodes visual targets and conditions before the backbone
forward pass, reducing the memory required for visual encoding.

\paragraph{Workload-aware monitoring.}
The implementation supports text and visual loss monitoring by task group,
records token throughput, and estimates utilization with an architecture-aware
FLOP model that distinguishes full-attention and linear-attention layers
and accounts for visual computation. Executed-compute estimates also
consider activation recomputation. Token throughput, peak memory, and
packed sequence lengths provide complementary workload measurements.

\section{Experiments}
\label{sec:experiments}

We evaluate \methodname{} on multimodal understanding, text-to-image
generation, and instruction-based image editing, and measure inference
latency and memory across image resolutions. The comparisons include both
task-specific models and unified multimodal models. \revised{{}Fig.~\ref{fig:benchmark-preview}
summarizes selected benchmarks, while the following tables provide
per-benchmark and per-category results.}

\subsection{Evaluation Setup and Baselines}
\label{sec:huge_eval_setup}

\modelname{} uses approximately 9B parameters in each of its understanding
and generation branches. Model-size conventions and inference settings
are specified in the corresponding comparisons.

\revised{{}For multimodal understanding, we use
MME~\citep{fu2023mme} for perception and cognition,
MMBench~\citep{Mmbench} and MM-Vet~\citep{yu2024mmvet} for visual question
answering and integrated multimodal capabilities,
MMMU~\citep{yue2024mmmu} for subject-level reasoning,
MathVista~\citep{mathvista} for visual mathematics, and
MMVP~\citep{tong2024eyes} for fine-grained visual discrimination.}

\revised{{}For text-to-image generation, we evaluate
compositional alignment with GenEval~\citep{ghosh2023geneval} and detailed
prompt following with DPG-Bench~\citep{hu2024ella}. For instruction-based
image editing, ImgEdit~\citep{ye2025imgedit} measures whether the model
makes the requested changes while preserving unrelated content.
The following subsections specify evaluation protocols and analyze
the corresponding results.}

\revised{We compare understanding-only models, including} InternVL2/2.5~\citep{opengvlab2024internvl2,chen2024internvl25} and
Qwen2/2.5-VL~\citep{wang2024qwen2vl,bai2025qwen25vl} families,
Kimi-VL~\citep{kimi2025vl}, and DeepSeek-VL2~\citep{wu2024deepseekvl2},
with unified models such as Janus-Pro~\citep{chen2025janus},
MetaQuery-XL~\citep{pan2025metaquery},
BAGEL~\citep{bagel}, and SenseNova-U1~\citep{diao2026sensenovau1}.
Qwen3.5-4B and Qwen3.5-9B~\citep{qwen2026qwen35modelcard} provide
additional understanding references. This grouping compares \methodname{}
with both specialized vision-language models and models that also generate images.

The GenEval and DPG-Bench comparisons include specialized generators such
as FLUX.1 [Dev]~\citep{blackforestlabs2024flux1}, Seedream 3.0~\citep{gao2025seedream},
Qwen-Image~\citep{wu2025qwen}, Emu3-Gen~\citep{wang2024emu3}, and
Lumina-mGPT 2.0~\citep{xin2025lumina}. Unified baselines include
BAGEL, OmniGen2~\citep{wu2025omnigen2}, InternVL-U~\citep{tian2026internvl},
Lumina-DiMOO~\citep{dimoo}, and LLaDA2.0-Uni~\citep{ai2026llada2}.
The tables distinguish autoregressive, continuous-diffusion, and
discrete-diffusion architectures.

ImgEdit compares four specialized editors (FLUX.1 Kontext~\citep{blackforestlabs2025kontext},
Step1X-Edit~\citep{liu2025step1xedit}, Qwen-Image-Edit~\citep{qwen2025imageedit},
and Z-Image-Edit~\citep{cai2025z}) with five unified models: BAGEL,
OmniGen2, InternVL-U, Lumina-DiMOO, and LLaDA2.0-Uni.
\revised{For inference efficiency, BAGEL and SenseNova-U1 serve as references
under matched hardware and workload settings.}

\revised{Understanding baseline scores come from BAGEL's comparison and additional
Qwen3.5 and SenseNova-U1 evaluations.} Generation and editing baselines
follow LLaDA2.0-Uni~\citep{ai2026llada2}. Published scores retain their
source evaluation settings; \methodname{} scores and the inference-efficiency
measurements are reported separately. Dashes indicate unreported results.

\subsection{Multimodal Understanding}
\label{sec:huge_eval_understanding}

\revised{We report MME perception (MME-P) and combined perception-plus-cognition
(MME-S), MMBench v1.0-dev-en circular accuracy, MMMU validation accuracy
without CoT, and MMVP paired accuracy. Protocol differences among the
reported models are documented in \tabref{tab:huge_understanding_published}.}

\begin{table}[t]
\centering\footnotesize
\setlength{\tabcolsep}{3pt}
\renewcommand{\arraystretch}{1.12}
\definecolor{bagelrow}{HTML}{D2E1FF}
\resizebox{\linewidth}{!}{%
\begin{tabular}{@{}clcrrrrrrr@{}}
\toprule
\textbf{Type} & \textbf{Model} & \makecell{\textbf{\# LLM}\\\textbf{Params}} & \textbf{MME-P}$\uparrow$ & \textbf{MME-S}$\uparrow$ & \textbf{MMBench}$\uparrow$ & \textbf{MMMU}$\uparrow$ & \textbf{MM-Vet}$\uparrow$ & \textbf{MathVista}$\uparrow$ & \textbf{MMVP}$\uparrow$ \\
\midrule
\multirow{15}{*}{\rotatebox{90}{\textit{Und. Only}}} & InternVL2~\citeyearpar{opengvlab2024internvl2} & 1.8B & 1440.0 & 1877.0 & 73.2 & 34.3 & 44.6 & 46.4 & 35.3 \\
 & InternVL2.5~\citeyearpar{chen2024internvl25} & 1.8B & -- & 2138.0 & 74.7 & 43.6 & 60.8 & 51.3 & -- \\
 & Qwen2-VL~\citeyearpar{wang2024qwen2vl} & 1.5B & -- & 1872.0 & 74.9 & 41.1 & 49.5 & 43.0 & -- \\
 & BLIP-3~\citeyearpar{xue2024blip3} & 4B & -- & -- & 76.8 & 41.1 & -- & 39.6 & -- \\
 & LLaVA-OV~\citeyearpar{li2024llavaonevision} & 7B & 1580.0 & -- & 80.8 & 48.8 & 57.5 & 63.2 & -- \\
 & InternVL2~\citeyearpar{opengvlab2024internvl2} & 7B & 1648.0 & 2210.0 & 81.7 & 49.3 & 54.2 & 58.3 & 51.3 \\
 & InternVL2.5~\citeyearpar{chen2024internvl25} & 7B & -- & 2344.0 & 84.6 & 56.0 & 62.8 & 64.4 & -- \\
 & Qwen2-VL~\citeyearpar{wang2024qwen2vl} & 7B & -- & 2327.0 & 83.0 & 54.1 & 62.0 & 58.2 & -- \\
 & Emu3-Chat$^{**}$~\citeyearpar{wang2024emu3} & 8B & 1244.0 & -- & 58.5 & 31.6 & 37.2 & -- & 36.6 \\
 & DeepSeek-VL2~\citeyearpar{wu2024deepseekvl2} & 4.1B/28B & -- & -- & -- & 51.1 & 60.0 & 62.8 & -- \\
 & Qwen2.5-VL~\citeyearpar{bai2025qwen25vl} & 3B & -- & 2157.0 & 79.1 & 53.1 & 61.8 & 62.3 & -- \\
 & Qwen2.5-VL~\citeyearpar{bai2025qwen25vl} & 7B & -- & 2347.0 & 83.5 & \textbf{58.6} & 67.1 & 68.2 & -- \\
 & Kimi-VL~\citeyearpar{kimi2025vl} & 2.8B/16B & -- & -- & -- & \underline{57.0} & 66.7 & 68.7 & -- \\
 & Qwen3.5-4B~\citeyearpar{qwen2026qwen35modelcard} & 4B & 1686.5 & 2348.3 & 81.4 & 29.2 & \underline{76.5} & \textbf{80.7} & 64.7 \\
 & Qwen3.5-9B~\citeyearpar{qwen2026qwen35modelcard} & 9B & \underline{1748.7} & \underline{2439.1} & \underline{85.1} & 43.1 & \textbf{79.6} & \underline{79.6} & \underline{70.7} \\
\midrule
 & Show-o$_{512}$~\citeyearpar{xie2024showo} & 1.3B & 1097.0 & -- & -- & 26.7 & -- & -- & -- \\
 & Janus~\citeyearpar{wu2024janus} & 1.5B & 1338.0 & -- & 69.4 & 30.5 & 34.3 & -- & -- \\
 & ILLUME~\citeyearpar{wang2024illume} & 7B & 1445.0 & -- & 75.1 & 38.2 & 37.0 & -- & -- \\
 & VILA-U$_{256}^{**}$~\citeyearpar{wu2024vilau} & 7B & 1336.0 & -- & 66.6 & 32.2 & 27.7 & -- & 22.0 \\
 & Chameleon$^{**}$~\citeyearpar{team2024chameleon} & 7B & -- & -- & 35.7 & 28.4 & 8.3 & -- & 0.0 \\
 & LlamaFusion$^{**}$~\citeyearpar{shi2024llamafusion} & 8B & 1604.0 & -- & 72.1 & 41.7 & -- & -- & -- \\
 & MetaMorph~\citeyearpar{tong2024metamorph} & 8B & -- & -- & 75.2 & 41.8 & -- & -- & 48.3 \\
 & SEED-X~\citeyearpar{ge2024seedx} & 13B & 1457.0 & -- & 70.1 & 35.6 & 43.0 & -- & -- \\
 & TokenFlow-XL~\citeyearpar{qu2024tokenflow} & 13B & 1546.0 & -- & 68.9 & 38.7 & 40.7 & -- & -- \\
 & MUSE-VL~\citeyearpar{xie2024musevl} & 32B & -- & -- & 81.8 & 50.1 & -- & 55.9 & -- \\
 & Janus-Pro~\citeyearpar{chen2025janus} & 1.5B & 1444.0 & -- & 75.5 & 36.3 & 39.8 & -- & -- \\
 & BAGEL~\citeyearpar{bagel} & 1.5B MoT & 1610.0 & 2183.0 & 79.2 & 43.2 & 48.2 & 63.4 & 54.7 \\
 & Janus-Pro~\citeyearpar{chen2025janus} & 7B & 1567.0 & -- & 79.2 & 41.0 & 50.0 & -- & -- \\
 & MetaQuery-XL$^{\dagger}$~\citeyearpar{pan2025metaquery} & 7B & 1685.0 & -- & 83.5 & \textbf{58.6} & 66.6 & -- & -- \\
 & BAGEL~\citeyearpar{bagel} & 7B MoT & 1687.0 & 2388.0 & 85.0 & 55.3 & 67.2 & 73.1 & 69.3 \\
 & SenseNova-U1~\citeyearpar{diao2026sensenovau1} & 8B MoT & 1683.5 & 2343.5 & 82.8 & 53.1 & 63.0 & 71.5 & 58.7 \\
\rowcolor{bagelrow}
\multirow{-17}{*}{\rotatebox{90}{\textit{Unified}}} & \textbf{\methodname{}} & 9B MoT & \textbf{1766.0} & \textbf{2470.3} & \textbf{86.7} & 56.7 & 73.4 & 77.2 & \textbf{72.0} \\
\bottomrule
\end{tabular}%
}
\caption{\textbf{Bold} and \underline{underlining} indicate the best and
second-best scores. MoT parameters are per branch; MoE parameters are
active/total. MMBench answer extraction differs: \methodname{} uses rules
only, while Qwen3.5 and SenseNova-U1 use a Qwen fallback.
$\dagger$: frozen Qwen2.5-VL backbone.
$**$: scores taken from MetaMorph or MetaQuery in the source comparison.
Dashes indicate unreported scores.}
\label{tab:huge_understanding_published}
\end{table}

\revised{In \tabref{tab:huge_understanding_published}, \methodname{} records the
highest listed MME-P (1766.0), MME-S (2470.3), MMBench (86.7), and MMVP
(72.0) scores, with the MMBench protocol differences noted in the caption.
Compared with BAGEL (7B MoT), it scores higher across all six benchmark
families, including gains of 6.2 points on MM-Vet and 4.1 points on
MathVista. These results indicate broad understanding capability spanning
perception, visual question answering, and visual mathematics.}

\revised{The remaining gaps are concentrated in specific reasoning benchmarks.
On MMMU-Val, \methodname{} scores 56.7, below MetaQuery-XL's 58.6.
Qwen3.5-9B also scores higher on MM-Vet and MathVista by 6.2 and 2.4
points, respectively. Thus, the strong perception results do not imply
uniformly better reasoning. Differences in model size, training data,
and evaluation protocols also prevent attributing these comparisons
solely to hybrid attention.}

\subsection{Text-to-Image Generation}
\label{sec:text-to-image-generation}
\label{sec:general-evaluation}

\revised{We report overall and category-level GenEval and DPG-Bench scores.}

Baseline scores in \tabref{tab:huge_geneval_results} and
\tabref{tab:huge_dpg_results} \revised{{}follow LLaDA2.0-Uni~\citep{ai2026llada2}.}
\revised{We report \methodname{} results after unified supervised fine-tuning (SFT).}
\revised{{}We label Lumina-mGPT 2.0 as autoregressive (AR), following its architecture~\citep{xin2025lumina}; this corrects the diffusion label in the DPG-Bench source table.}

\begin{table}[t]
\centering\footnotesize
\definecolor{bagelrow}{HTML}{D2E1FF}
\setlength{\tabcolsep}{3pt}
\renewcommand{\arraystretch}{1.12}
\begin{tabularx}{\linewidth}{@{}cXcrrrrrrr@{}}
\toprule
\textbf{Type} & \textbf{Model} & \textbf{Architecture} & \makecell{\textbf{Single}\\\textbf{object}} & \makecell{\textbf{Two}\\\textbf{objects}} & \textbf{Counting} & \textbf{Colors} & \textbf{Position} & \makecell{\textbf{Attribute}\\\textbf{binding}} & \textbf{Overall}$\uparrow$ \\
\midrule
\multirow{7}{*}{\rotatebox{90}{Gen. Only}} & FLUX.1 [Dev]~\citeyearpar{blackforestlabs2024flux1} & CD & 0.98 & 0.81 & 0.74 & 0.79 & 0.22 & 0.45 & 0.66 \\
 & Emu3-Gen~\citeyearpar{wang2024emu3} & AR & 0.98 & 0.71 & 0.34 & 0.81 & 0.17 & 0.21 & 0.54 \\
 & Lumina-mGPT 2.0~\citeyearpar{xin2025lumina} & AR & 0.99 & 0.87 & 0.44 & 0.85 & 0.44 & 0.54 & 0.69 \\
 & Seedream 3.0~\citeyearpar{gao2025seedream} & CD & 0.99 & 0.96 & 0.91 & 0.93 & 0.47 & 0.80 & 0.84 \\
 & Qwen-Image~\citeyearpar{wu2025qwen} & CD & 0.99 & 0.92 & 0.89 & 0.88 & 0.76 & 0.77 & 0.87 \\
 & LongCat-Image~\citeyearpar{team2025longcat} & CD & 0.99 & 0.98 & 0.86 & 0.86 & 0.75 & 0.73 & 0.87 \\
 & Z-Image-Turbo~\citeyearpar{cai2025z} & CD & 1.00 & 0.95 & 0.77 & 0.89 & 0.65 & 0.68 & 0.82 \\
\midrule
 & Janus-Pro~\citeyearpar{chen2025janus} & AR & 0.99 & 0.89 & 0.59 & 0.90 & 0.79 & 0.66 & 0.80 \\
 & BAGEL~\citeyearpar{bagel} & AR + CD & 0.99 & 0.94 & 0.81 & 0.88 & 0.64 & 0.63 & 0.82 \\
 & OmniGen2~\citeyearpar{wu2025omnigen2} & AR + CD & 1.00 & 0.95 & 0.64 & 0.88 & 0.55 & 0.76 & 0.80 \\
 & HunyuanImage-3.0~\citeyearpar{cao2025hunyuanimage} & AR + CD & 1.00 & 0.92 & 0.48 & 0.82 & 0.42 & 0.63 & 0.72 \\
 & MMaDA~\citeyearpar{Mmada} & DD & 0.99 & 0.76 & 0.61 & 0.84 & 0.20 & 0.37 & 0.63 \\
 & Lumina-DiMOO~\citeyearpar{dimoo} & DD & 1.00 & 0.94 & 0.85 & 0.89 & 0.85 & 0.76 & 0.88 \\
 & NextFlow~\citeyearpar{zhang2026nextflow} & AR + CD & 0.98 & 0.92 & 0.73 & 0.90 & 0.77 & 0.69 & 0.83 \\
 & InternVL-U~\citeyearpar{tian2026internvl} & AR + CD & 0.99 & 0.94 & 0.74 & 0.91 & 0.77 & 0.74 & 0.85 \\
 & LLaDA-o~\citeyearpar{you2026lladao} & DD + CD & 0.99 & 0.98 & 0.73 & 0.96 & 0.69 & 0.83 & 0.86 \\
 & LLaDA2.0-Uni~\citeyearpar{ai2026llada2} & DD + CD & 1.00 & 0.98 & 0.73 & 0.92 & 0.90 & 0.84 & 0.89 \\
\rowcolor{bagelrow}
\multirow{-11}{*}{\rotatebox{90}{\textit{Unified}}} & \textbf{\methodname{}} & AR + CD & 0.99 & 0.93 & 0.68 & 0.94 & 0.67 & 0.73 & 0.83 \\
\bottomrule
\end{tabularx}
\caption{\textbf{Text-to-image generation on GenEval.} Scores range from 0 to 1; higher is better. The blue row reports \methodname{} after unified SFT. AR, CD, and DD denote autoregressive, continuous diffusion, and discrete diffusion models, respectively; ``+'' indicates combined architectures.}
\label{tab:huge_geneval_results}
\end{table}

\begin{table}[t]
\centering\footnotesize
\definecolor{bagelrow}{HTML}{D2E1FF}
\setlength{\tabcolsep}{3pt}
\renewcommand{\arraystretch}{1.12}
\begin{tabularx}{\linewidth}{@{}cXcrrrrrr@{}}
\toprule
\textbf{Type} & \textbf{Model} & \textbf{Architecture} & \textbf{Global} & \textbf{Entity} & \textbf{Attribute} & \textbf{Relation} & \textbf{Other} & \textbf{Overall}$\uparrow$ \\
\midrule
\multirow{7}{*}{\rotatebox{90}{Gen. Only}} & FLUX.1 [Dev]~\citeyearpar{blackforestlabs2024flux1} & CD & 74.35 & 90.00 & 88.96 & 90.87 & 88.33 & 83.84 \\
 & Emu3-Gen~\citeyearpar{wang2024emu3} & AR & 85.21 & 86.68 & 86.84 & 90.22 & 83.15 & 80.60 \\
 & Lumina-mGPT 2.0~\citeyearpar{xin2025lumina} & AR & 86.63 & 91.97 & 90.20 & 94.85 & 84.80 & 87.20 \\
 & Seedream 3.0~\citeyearpar{gao2025seedream} & CD & 94.31 & 92.65 & 91.36 & 92.78 & 88.24 & 88.27 \\
 & Qwen-Image~\citeyearpar{wu2025qwen} & CD & 91.32 & 91.56 & 92.02 & 94.31 & 92.73 & 88.32 \\
 & LongCat-Image~\citeyearpar{team2025longcat} & CD & 89.10 & 92.54 & 92.00 & 93.28 & 87.50 & 86.80 \\
 & Z-Image-Turbo~\citeyearpar{cai2025z} & CD & 91.29 & 89.59 & 90.14 & 92.16 & 88.68 & 84.86 \\
\midrule
 & Janus-Pro~\citeyearpar{chen2025janus} & AR & 86.90 & 88.90 & 89.40 & 89.32 & 89.48 & 84.19 \\
 & BAGEL~\citeyearpar{bagel} & AR + CD & 88.94 & 90.37 & 91.29 & 90.82 & 88.67 & 85.07 \\
 & OmniGen2~\citeyearpar{wu2025omnigen2} & AR + CD & 88.81 & 88.83 & 90.18 & 89.37 & 90.27 & 83.57 \\
 & HunyuanImage-3.0~\citeyearpar{cao2025hunyuanimage} & AR + CD & 92.12 & 92.53 & 89.13 & 92.13 & 91.92 & 86.10 \\
 & MMaDA~\citeyearpar{Mmada} & DD & 77.81 & 78.48 & 81.74 & 84.79 & 63.20 & 69.97 \\
 & Lumina-DiMOO~\citeyearpar{dimoo} & DD & 81.46 & 92.08 & 88.98 & 94.31 & 82.00 & 86.04 \\
 & NextFlow~\citeyearpar{zhang2026nextflow} & AR + CD & 92.40 & 90.05 & 90.51 & 92.72 & 91.14 & 86.00 \\
 & InternVL-U~\citeyearpar{tian2026internvl} & AR + CD & 90.39 & 90.78 & 90.68 & 90.29 & 88.77 & 85.18 \\
 & LLaDA-o~\citeyearpar{you2026lladao} & DD + CD & 92.91 & 93.30 & 90.40 & 91.75 & 92.79 & 87.04 \\
 & LLaDA2.0-Uni~\citeyearpar{ai2026llada2} & DD + CD & 91.14 & 93.55 & 91.98 & 92.17 & 93.18 & 87.76 \\
\rowcolor{bagelrow}
\multirow{-11}{*}{\rotatebox{90}{\textit{Unified}}} & \textbf{\methodname{}} & AR + CD & 79.33 & 82.62 & 77.07 & 76.71 & 73.50 & 80.54 \\
\bottomrule
\end{tabularx}
\caption{\textbf{Text-to-image generation on DPG-Bench.} Higher is better. The blue row reports \methodname{} after unified SFT. AR, CD, and DD denote autoregressive, continuous diffusion, and discrete diffusion models, respectively; ``+'' indicates combined architectures.}
\label{tab:huge_dpg_results}
\end{table}

\revised{After unified SFT, \methodname{} achieves an overall GenEval score
of 0.83 (\tabref{tab:huge_geneval_results}), close to
BAGEL's direct-generation score of 0.82. Its scores are highest on
single objects (0.99), colors (0.94), and two objects (0.93), while
counting (0.68), position (0.67), and attribute binding (0.73) remain
more difficult. This pattern shows stronger object and color fidelity
than precise composition. LLaDA2.0-Uni and Lumina-DiMOO reach higher
overall scores of 0.89 and 0.88, leaving room to improve compositional
control. The BAGEL comparison uses direct generation, keeping results
with external prompt rewriting separate.}

\revised{\tabref{tab:huge_dpg_results} reports a DPG-Bench overall score of
80.54 for \methodname{}, below BAGEL (85.07) and LLaDA2.0-Uni (87.76).
Within its category results, Entity is strongest at 82.62, whereas
Attribute (77.07), Relation (76.71), and Other (73.50) are lower.
Together with GenEval, this pattern points to attribute and relational
control as priorities for improving prompt adherence.}

\subsection{Instruction-Based Image Editing}
\label{sec:image-editing}

\revised{{}We evaluate \methodname{} on all 737 ImgEdit
samples using Qwen3-VL-235B-A22B-Thinking as the judge.
\tabref{tab:huge_imgedit_results} reports nine editing categories and the
overall score. Baseline scores follow
LLaDA2.0-Uni~\citep{ai2026llada2}. Because the judges differ, our scores
are reference results and are not directly comparable to those baselines.}

\begin{figure}[t]
\centering
\begin{minipage}{.88\linewidth}
\centering
\begingroup

\noindent\makebox[.10\linewidth]{}\hfill
\makebox[.219\linewidth]{\scriptsize F1}\hfill
\makebox[.219\linewidth]{\scriptsize F2}\hfill
\makebox[.219\linewidth]{\scriptsize F3}\hfill
\makebox[.219\linewidth]{\scriptsize F4}\par\vspace{2pt}
\noindent\makebox[\linewidth][l]{\small\textbf{(a) Silk ribbon in a breeze}}\par
{\scriptsize\textit{Prompt summary:} Red silk ribbon on a wooden dowel, pale blue background. Lift, form a wave, then relax.\par}\vspace{3pt}
\noindent\begin{minipage}[c]{.10\linewidth}\raggedright\scriptsize \textbf{T2I}\par All frames generated\end{minipage}\hfill
\begin{minipage}[c]{.219\linewidth}\includegraphics[width=\linewidth]{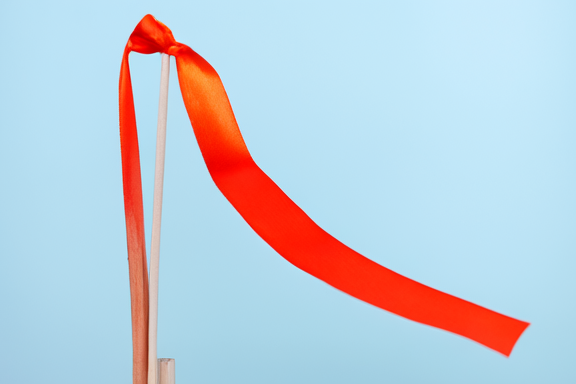}\end{minipage}\hfill
\begin{minipage}[c]{.219\linewidth}\includegraphics[width=\linewidth]{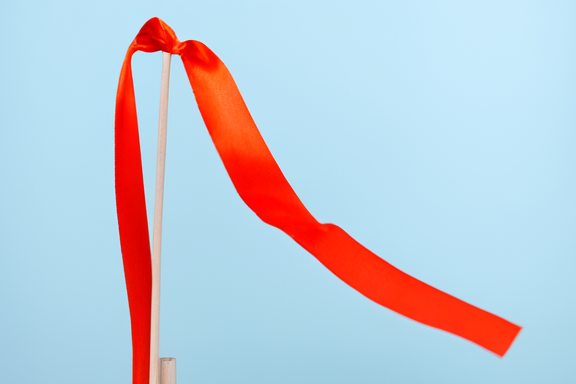}\end{minipage}\hfill
\begin{minipage}[c]{.219\linewidth}\includegraphics[width=\linewidth]{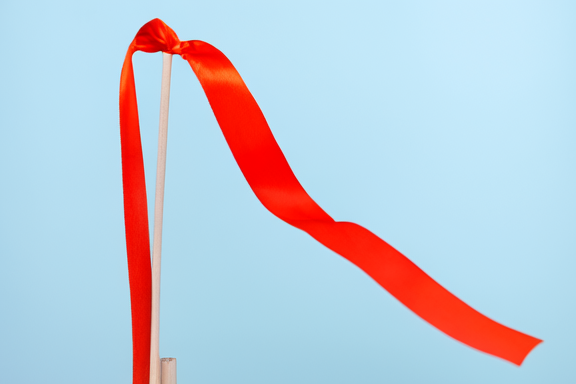}\end{minipage}\hfill
\begin{minipage}[c]{.219\linewidth}\includegraphics[width=\linewidth]{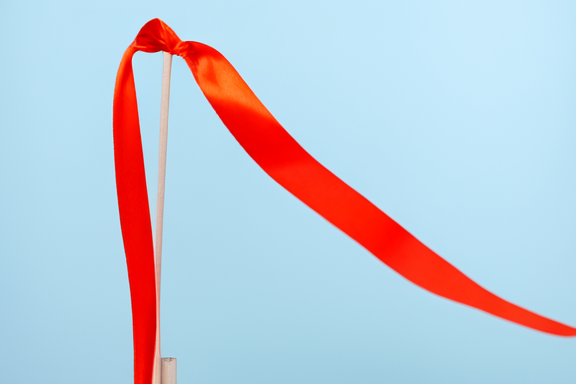}\end{minipage}\par\vspace{2pt}
\noindent\begin{minipage}[c]{.10\linewidth}\raggedright\scriptsize \textbf{Editing}\par F1: input\par F2--F4: generated\end{minipage}\hfill
\begin{minipage}[c]{.219\linewidth}\includegraphics[width=\linewidth]{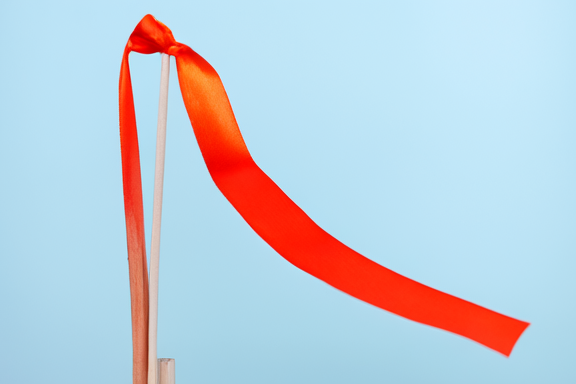}\end{minipage}\hfill
\begin{minipage}[c]{.219\linewidth}\includegraphics[width=\linewidth]{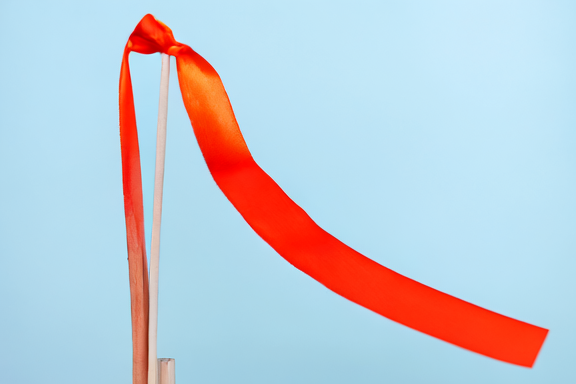}\end{minipage}\hfill
\begin{minipage}[c]{.219\linewidth}\includegraphics[width=\linewidth]{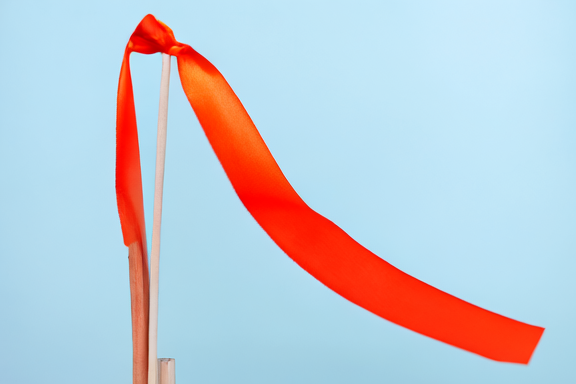}\end{minipage}\hfill
\begin{minipage}[c]{.219\linewidth}\includegraphics[width=\linewidth]{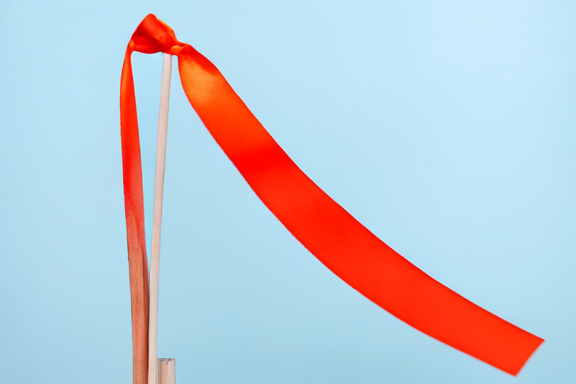}\end{minipage}\par\vspace{2pt}\vspace{4pt}
\noindent\makebox[\linewidth][l]{\small\textbf{(b) A pulsing jellyfish}}\par
{\scriptsize\textit{Prompt summary:} One pale pink jellyfish in a dark blue aquarium. Contract, expand, then contract again.\par}\vspace{3pt}
\noindent\begin{minipage}[c]{.10\linewidth}\raggedright\scriptsize \textbf{T2I}\par All frames generated\end{minipage}\hfill
\begin{minipage}[c]{.219\linewidth}\includegraphics[width=\linewidth]{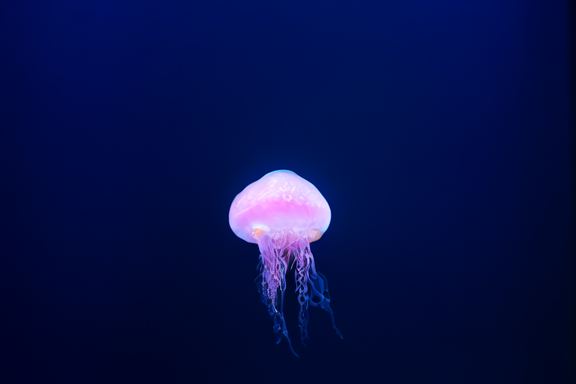}\end{minipage}\hfill
\begin{minipage}[c]{.219\linewidth}\includegraphics[width=\linewidth]{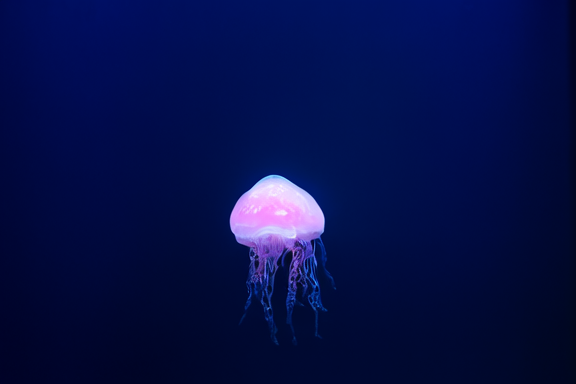}\end{minipage}\hfill
\begin{minipage}[c]{.219\linewidth}\includegraphics[width=\linewidth]{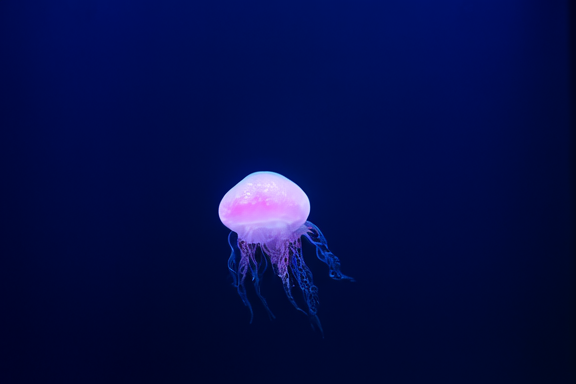}\end{minipage}\hfill
\begin{minipage}[c]{.219\linewidth}\includegraphics[width=\linewidth]{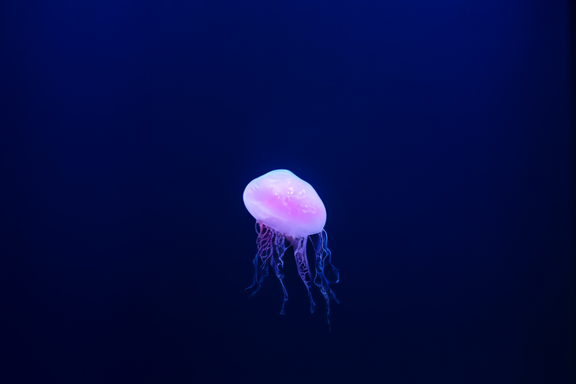}\end{minipage}\par\vspace{2pt}
\noindent\begin{minipage}[c]{.10\linewidth}\raggedright\scriptsize \textbf{Editing}\par F1: input\par F2--F4: generated\end{minipage}\hfill
\begin{minipage}[c]{.219\linewidth}\includegraphics[width=\linewidth]{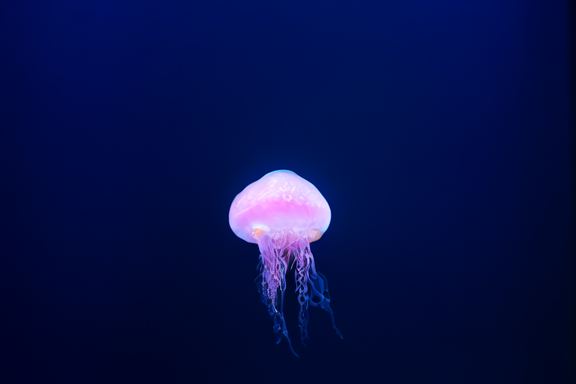}\end{minipage}\hfill
\begin{minipage}[c]{.219\linewidth}\includegraphics[width=\linewidth]{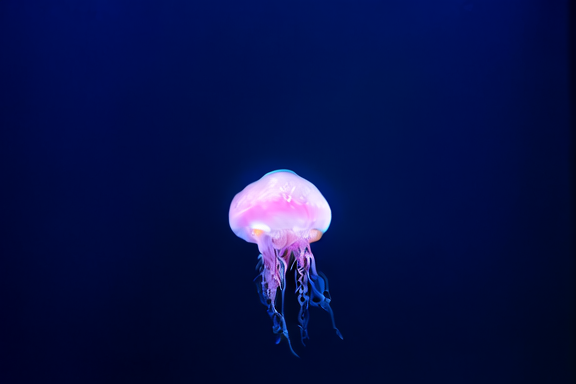}\end{minipage}\hfill
\begin{minipage}[c]{.219\linewidth}\includegraphics[width=\linewidth]{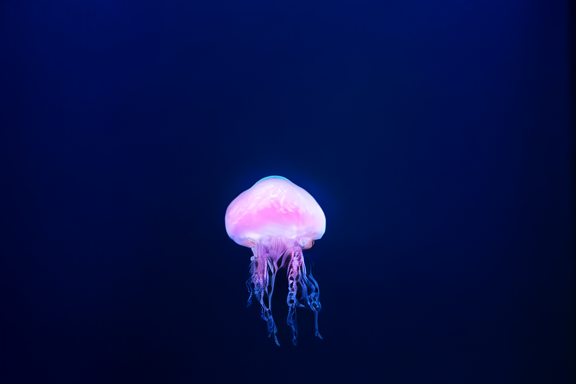}\end{minipage}\hfill
\begin{minipage}[c]{.219\linewidth}\includegraphics[width=\linewidth]{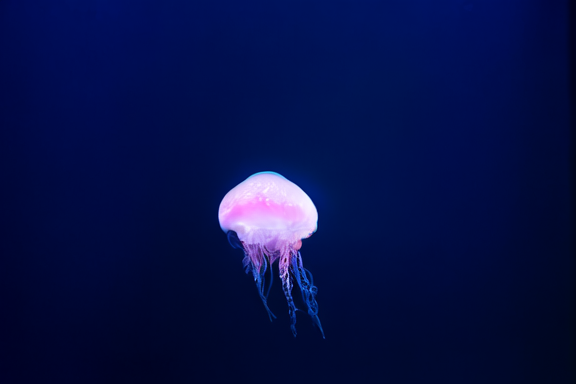}\end{minipage}\par\vspace{2pt}\vspace{4pt}
\noindent\makebox[\linewidth][l]{\small\textbf{(c) Candle flame in a breeze}}\par
{\scriptsize\textit{Prompt summary:} Ivory candle in a brass holder on a dark table. Flame upright, left, right, then upright.\par}\vspace{3pt}
\noindent\begin{minipage}[c]{.10\linewidth}\raggedright\scriptsize \textbf{T2I}\par All frames generated\end{minipage}\hfill
\begin{minipage}[c]{.219\linewidth}\includegraphics[width=\linewidth]{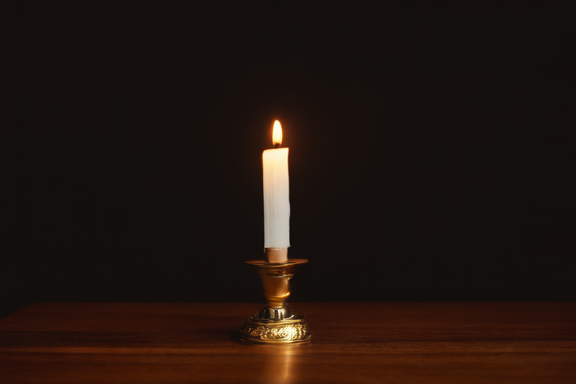}\end{minipage}\hfill
\begin{minipage}[c]{.219\linewidth}\includegraphics[width=\linewidth]{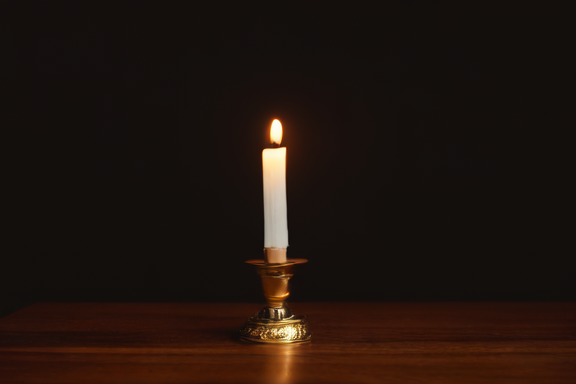}\end{minipage}\hfill
\begin{minipage}[c]{.219\linewidth}\includegraphics[width=\linewidth]{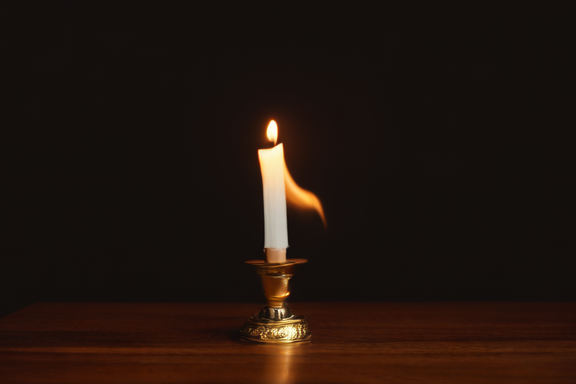}\end{minipage}\hfill
\begin{minipage}[c]{.219\linewidth}\includegraphics[width=\linewidth]{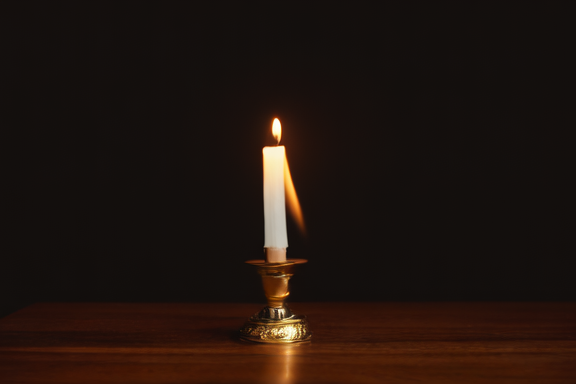}\end{minipage}\par\vspace{2pt}
\noindent\begin{minipage}[c]{.10\linewidth}\raggedright\scriptsize \textbf{Editing}\par F1: input\par F2--F4: generated\end{minipage}\hfill
\begin{minipage}[c]{.219\linewidth}\includegraphics[width=\linewidth]{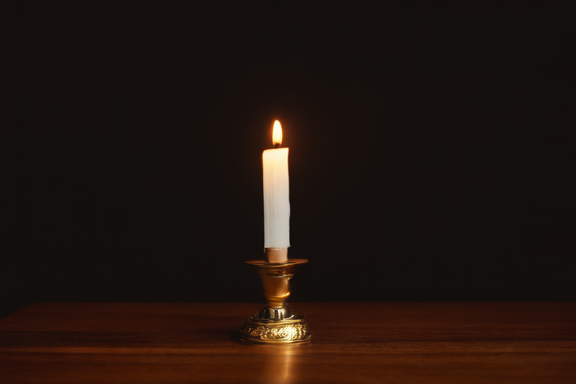}\end{minipage}\hfill
\begin{minipage}[c]{.219\linewidth}\includegraphics[width=\linewidth]{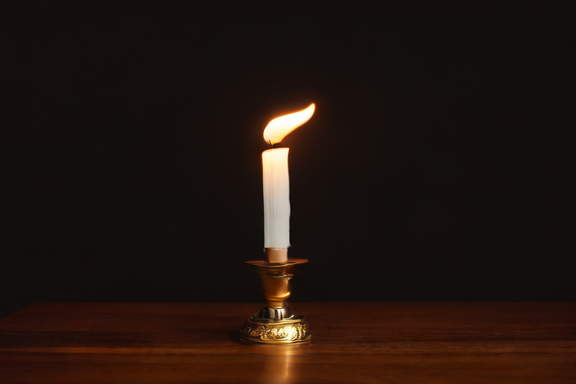}\end{minipage}\hfill
\begin{minipage}[c]{.219\linewidth}\includegraphics[width=\linewidth]{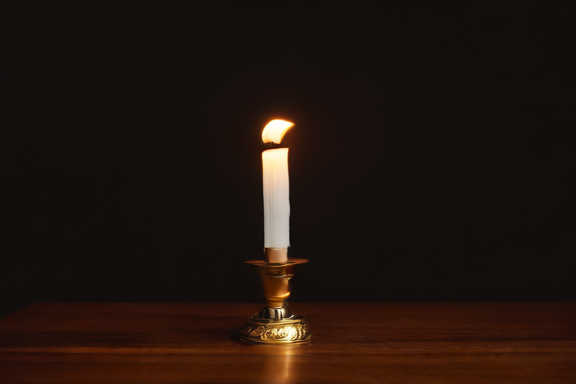}\end{minipage}\hfill
\begin{minipage}[c]{.219\linewidth}\includegraphics[width=\linewidth]{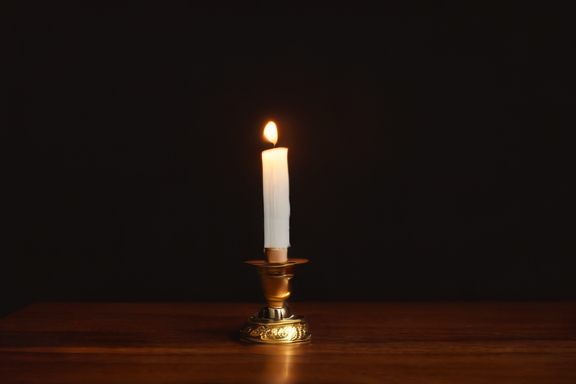}\end{minipage}\par\vspace{2pt}\vspace{4pt}
\endgroup
\end{minipage}
\caption{\textbf{Text-only generation (top) and first-frame-conditioned editing (bottom).} \revised{Generation predicts F1--F4 jointly, whereas editing conditions on F1 to predict F2--F4, illustrating visual continuity under two conditioning settings.}}
\label{fig:interleave-paired-novel}
\end{figure}

\begin{table}[t]
\centering\footnotesize
\definecolor{bagelrow}{HTML}{D2E1FF}
\setlength{\tabcolsep}{3pt}
\renewcommand{\arraystretch}{1.12}
\resizebox{\linewidth}{!}{%
\begin{tabular}{@{}clrrrrrrrrrr@{}}
\toprule
\textbf{Type} & \textbf{Model} & \textbf{Add} & \textbf{Adjust} & \textbf{Extract} & \textbf{Replace} & \textbf{Remove} & \textbf{Back.} & \textbf{Style} & \textbf{Hybrid} & \textbf{Action} & \textbf{Overall}$\uparrow$ \\
\midrule
\multirow{4}{*}{\rotatebox{90}{Gen. Only}} & FLUX.1 Kontext~\citeyearpar{blackforestlabs2025kontext} & 4.25 & 4.15 & 2.35 & 4.56 & 3.57 & 4.26 & 4.57 & 3.68 & 4.63 & 4.00 \\
 & Step1X-Edit~\citeyearpar{liu2025step1xedit} & 3.88 & 3.14 & 1.76 & 3.40 & 2.41 & 3.16 & 4.63 & 2.64 & 2.52 & 3.06 \\
 & Qwen-Image-Edit~\citeyearpar{qwen2025imageedit} & 4.32 & 4.36 & 4.04 & 4.64 & 4.52 & 4.37 & 4.84 & 3.39 & 4.71 & 4.35 \\
 & Z-Image-Edit~\citeyearpar{cai2025z} & 4.40 & 4.14 & 4.30 & 4.57 & 4.13 & 4.14 & 4.85 & 3.63 & 4.50 & 4.30 \\
\midrule
 & BAGEL~\citeyearpar{bagel} & 3.56 & 3.31 & 1.70 & 3.30 & 2.62 & 3.24 & 4.49 & 2.38 & 4.17 & 3.20 \\
 & OmniGen2~\citeyearpar{wu2025omnigen2} & 3.57 & 3.06 & 1.77 & 3.74 & 3.20 & 3.57 & 4.81 & 2.52 & 4.68 & 3.44 \\
 & Lumina-DiMOO~\citeyearpar{dimoo} & 3.41 & 2.38 & 1.90 & 3.26 & 2.21 & 2.11 & 4.19 & 2.26 & 3.17 & 2.77 \\
 & InternVL-U~\citeyearpar{tian2026internvl} & 4.13 & 3.40 & 2.27 & 4.13 & 3.39 & 3.84 & 4.77 & 3.03 & 4.05 & 3.67 \\
 & LLaDA2.0-Uni~\citeyearpar{ai2026llada2} & 3.76 & 4.16 & 2.40 & 4.04 & 3.82 & 4.07 & 4.60 & 3.97 & 4.42 & 3.92 \\
\rowcolor{bagelrow}
\multirow{-6}{*}{\rotatebox{90}{\textit{Unified}}} & \textbf{\methodname{}}$^{\dagger}$ & 3.72 & 2.54 & 1.68 & 3.51 & 3.52 & 3.47 & 4.05 & 1.62 & 3.44 & 3.13 \\
\bottomrule
\end{tabular}%
}
\caption{\textbf{Instruction-based image editing on ImgEdit.}
Back.: background editing; $^{\dagger}$: Qwen3-VL reference scores
(different judge).}
\label{tab:huge_imgedit_results}
\end{table}

\revised{In \tabref{tab:huge_imgedit_results}, \methodname{} obtains an overall
reference score of 3.13. Style editing is its strongest category (4.05),
followed by object addition (3.72). Extraction (1.68) and hybrid editing
(1.62) receive the lowest scores, identifying these operations as the
main weaknesses under the current evaluator. This variation shows why
the overall score alone is insufficient to characterize editing ability.
Because our Qwen3-VL judge differs from the judge used for the published
baselines, these scores support analysis across our editing categories
but do not establish a direct ranking against other models.}

\subsection{Interleaved Generation and Sequence Editing}
\label{sec:interleave-generation-editing}

Text-only generation jointly produces F1--F4. Sequence editing takes the
same F1 as input and jointly generates F2--F4.

\figref{fig:interleave-paired-novel} shows local changes in ribbon curvature,
jellyfish tentacles, and flame shape with recognizable subjects and backgrounds.
Flame direction and bell contraction remain imperfect. Selected from 12 paired runs, these cases
illustrate visual continuity, not reliable execution of every requested stage.

\subsection{Extension to Video Generation and Editing}
\label{sec:video_results}

Following the exploratory post-SFT adaptation described in
\secref{sec:video_train}, we examine the model's video generation and
editing capabilities through qualitative examples.

\begin{figure}[t]
    \centering
    \includegraphics[width=\textwidth]{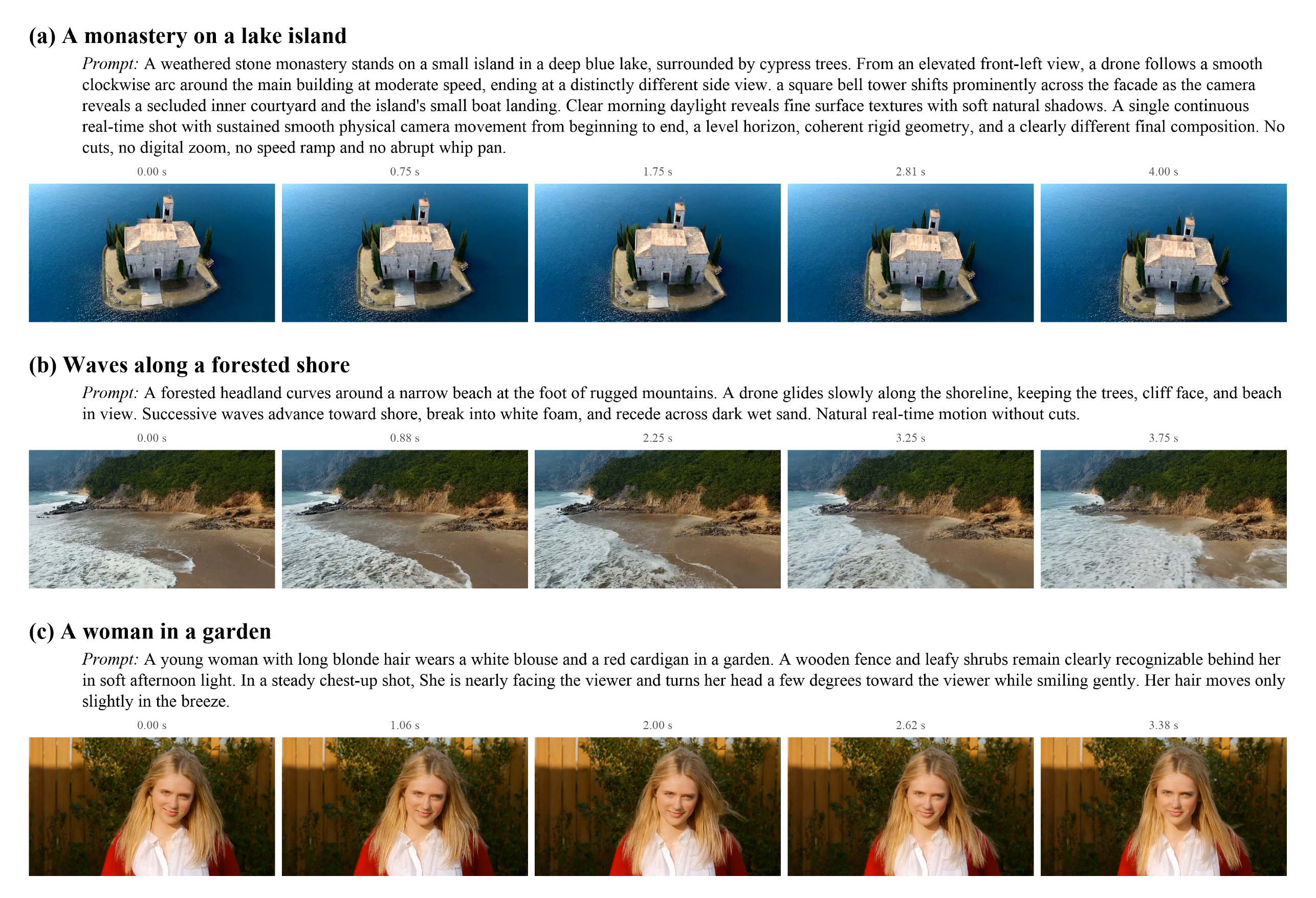}
    \caption{\textbf{Qualitative examples of video generation.}
    \revised{Chronological frames illustrate text-conditioned scene evolution, including camera movement, waves, and subtle human motion.}}
    \label{fig:video_generation}
\end{figure}

\figref{fig:video_generation} presents text-conditioned video generation
examples spanning camera movement around a building, waves along a
shoreline, and subtle human motion. Each example includes its text prompt
and five frames sampled from the generated clip, illustrating scene
appearance and its evolution over time.

\begin{figure}[t]
    \centering
    \includegraphics[width=\textwidth]{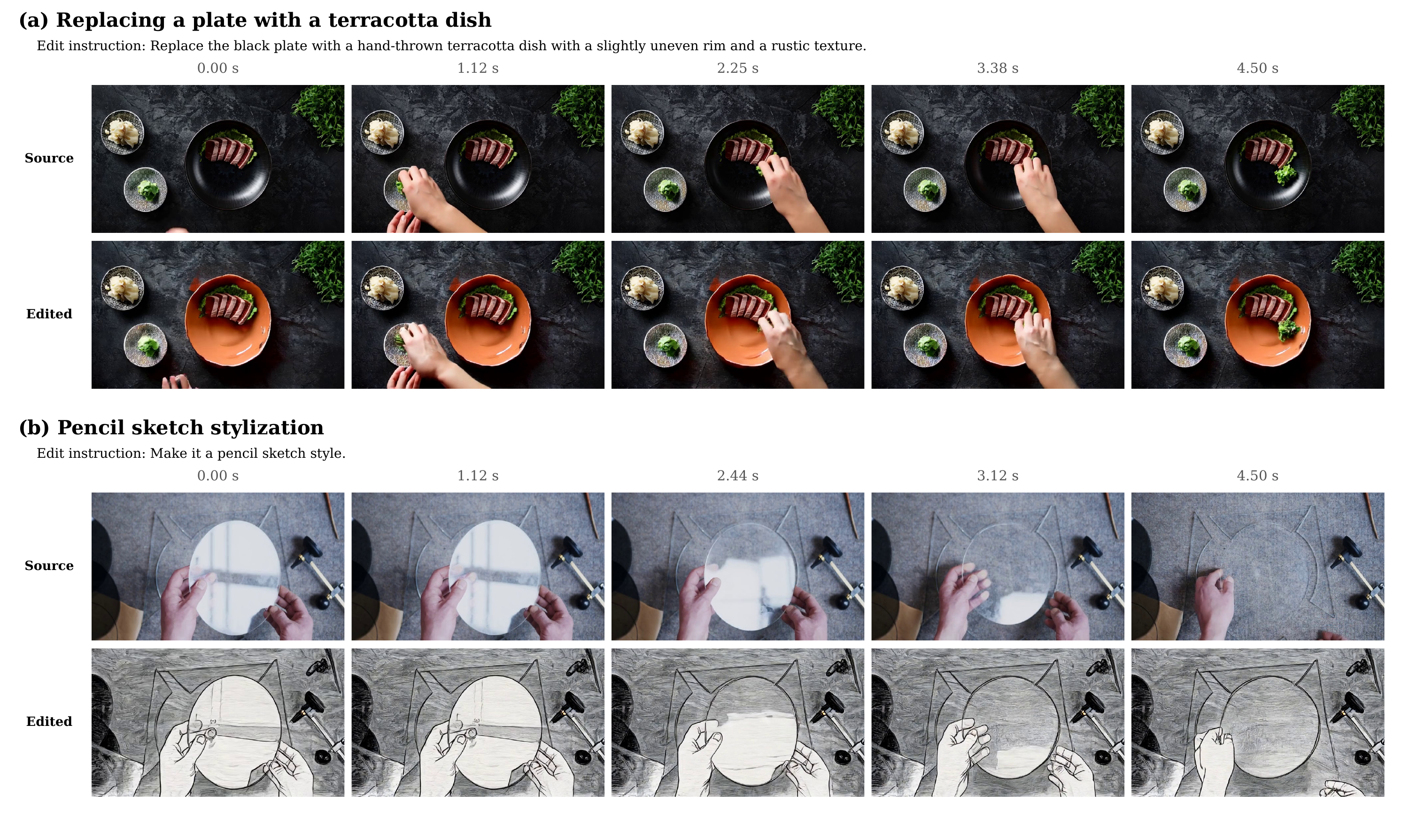}
    \caption{\textbf{Qualitative examples of video editing.}
    \revised{Temporally aligned source and edited frames illustrate object replacement and global stylization.}}
    \label{fig:video_editing}
\end{figure}

\figref{fig:video_editing} presents source-conditioned video editing
examples covering object replacement and global stylization.
The first replaces a black plate with a terracotta dish, while the
second transforms the video into a pencil-sketch style.
Source and edited frames are shown at corresponding timestamps to
illustrate the requested changes alongside the preservation of scene
content and action across frames.

\subsection{Inference Efficiency and Scaling}
\label{sec:huge_eval_ablation}

We compare \methodname{} with BAGEL~\citep{bagel} and
SenseNova-U1~\citep{diao2026sensenovau1} on image generation and visual
understanding. Generation is evaluated at matched output resolutions;
understanding is evaluated at matched visual-token budgets with a fixed
text output length. Each model runs on a dedicated H20-3e GPU in BF16
at batch size 1, with all model components GPU-resident. Each setting
uses five warm-ups and four input cases repeated three times. We report
median latency and throughput, descriptive P95 generation latency, and
the maximum allocated GPU memory across the 12 measured runs. Model
loading and disk I/O are excluded. Parameter counts in
Tables~\ref{tab:unispark_efficiency_results} and~\ref{tab:understanding_throughput}
follow the per-branch MoT convention
used in \tabref{tab:huge_understanding_published}.

\revised{For image generation, all models use the same prompts, 49 sampling updates, and 79 denoising
forwards.}
Text CFG is set to 4 over the noise-level interval $(0.4,1.0]$,
with time shift 1 and CFG normalization disabled. The interval is mapped
to each model's native time coordinate. Timing covers text prefill,
denoising, VAE decoding where applicable, and CPU image conversion.
Image-token throughput is the final spatial token count divided by
end-to-end latency; repeated denoising forwards do not count as newly
emitted tokens.

\begin{table}[t]
\centering\footnotesize
\definecolor{bagelrow}{HTML}{D2E1FF}
\setlength{\tabcolsep}{4pt}
\renewcommand{\arraystretch}{1.15}
\begin{tabularx}{\linewidth}{@{}cXccrrr@{}}
\toprule
\textbf{Resolution} & \textbf{Model} & \makecell{\textbf{\# LLM}\\\textbf{Params}} & \makecell{\textbf{Image}\\\textbf{tokens/s}$\uparrow$} & \makecell{\textbf{Median}\\\textbf{latency (s)}$\downarrow$} & \makecell{\textbf{P95}\\\textbf{latency (s)}$\downarrow$} & \makecell{\textbf{Peak allocated}\\\textbf{memory (GB)}$\downarrow$} \\
\midrule
 & BAGEL~\citeyearpar{bagel} & 7B MoT & 87.09 & 11.76 & 11.77 & 30.47 \\
 & SenseNova-U1~\citeyearpar{diao2026sensenovau1} & 8B MoT & 39.80 & 6.43 & 6.52 & 35.49 \\
\rowcolor{bagelrow}
\multirow{-3}{*}{$512\times512$} & \textbf{\methodname{}} & 9B MoT & 35.19 & 7.27 & 7.36 & 37.39 \\
\midrule
 & BAGEL~\citeyearpar{bagel} & 7B MoT & 89.44 & 45.80 & 45.81 & 33.29 \\
 & SenseNova-U1~\citeyearpar{diao2026sensenovau1} & 8B MoT & 73.90 & 13.86 & 13.91 & 35.84 \\
\rowcolor{bagelrow}
\multirow{-3}{*}{$1024\times1024$} & \textbf{\methodname{}} & 9B MoT & 72.33 & 14.16 & 14.21 & 42.47 \\
\midrule
 & BAGEL~\citeyearpar{bagel} & 7B MoT & -- & -- & -- & -- \\
 & SenseNova-U1~\citeyearpar{diao2026sensenovau1} & 8B MoT & 70.20 & 58.34 & 58.36 & 37.39 \\
\rowcolor{bagelrow}
\multirow{-3}{*}{$2048\times2048$} & \textbf{\methodname{}} & 9B MoT & 79.56 & 51.48 & 51.51 & 62.74 \\
\bottomrule
\end{tabularx}
\caption{\textbf{Image generation efficiency across output resolutions.}}
\label{tab:unispark_efficiency_results}
\end{table}

\revised{\tabref{tab:unispark_efficiency_results} shows that \methodname{}
reduces median generation latency relative to BAGEL from 11.76 to 7.27
seconds at $512^2$ and from 45.80 to 14.16 seconds at $1024^2$, giving
1.62$\times$ and 3.23$\times$ speedups. SenseNova-U1 is faster at these
two resolutions, but \methodname{} is faster at $2048^2$ (51.48 versus
58.34 seconds). This advantage comes with higher peak allocated memory:
62.74 GB versus 37.39 GB at $2048^2$. The results therefore show a
resolution-dependent latency benefit with a memory cost.}

\revised{Image-token throughput also depends on the representation size.
At $1024^2$, BAGEL uses 4,096 image tokens, compared with 1,024 for
\methodname{} and SenseNova-U1, so its higher tokens/s does not imply
faster image generation. At $2048^2$, the latter two models both use
4,096 tokens, and \methodname{} reaches 79.56 versus 70.20 tokens/s.
SenseNova-U1's two smaller settings are outside its recommended trained
resolution buckets, so these measurements characterize fixed workloads
rather than matched image quality.}

\revised{For visual understanding, we use the same four image--question pairs
with 256 or 1,024 visual tokens per image, checked at the language-model input.} To match these
budgets, BAGEL receives $224\times224$ or $448\times448$ images, while \methodname{}
and SenseNova-U1 receive $512\times512$ or $1024\times1024$ images. Each run makes
exactly 128 greedy token predictions, ignoring early termination.
Native tokenizers and chat templates are retained; any reasoning text
produced by SenseNova-U1 counts toward this fixed budget.

Images are loaded and resized before timing. Total latency includes
native preprocessing, vision encoding, multimodal prefill, and text
decoding. Decoding throughput is $127/(t_{128}-t_1)$, where $t_i$ is a
CUDA event recorded after the output projection for prediction $i$.
First-logit latency measures request start to $t_1$, before token
selection; it is not user-visible time to first token.

\begin{table}[t]
\centering\footnotesize
\definecolor{bagelrow}{HTML}{D2E1FF}
\setlength{\tabcolsep}{4pt}
\renewcommand{\arraystretch}{1.15}
\begin{tabularx}{\linewidth}{@{}cXccrrrr@{}}
\toprule
\makecell{\textbf{Visual}\\\textbf{tokens}} & \textbf{Model} & \makecell{\textbf{\# LLM}\\\textbf{Params}} & \textbf{Input} & \makecell{\textbf{Text decode}\\\textbf{tokens/s}$\uparrow$} & \makecell{\textbf{First logits}\\\textbf{(s)}$\downarrow$} & \makecell{\textbf{Median}\\\textbf{latency (s)}$\downarrow$} & \makecell{\textbf{Peak allocated}\\\textbf{memory (GB)}$\downarrow$} \\
\midrule
 & BAGEL~\citeyearpar{bagel} & 7B MoT & $224\times224$ & 36.09 & 0.176 & 3.70 & 29.58 \\
 & SenseNova-U1~\citeyearpar{diao2026sensenovau1} & 8B MoT & $512\times512$ & 17.92 & 0.165 & 7.26 & 35.32 \\
\rowcolor{bagelrow}
\multirow{-3}{*}{256} & \textbf{\methodname{}} & 9B MoT & $512\times512$ & 21.91 & 0.143 & 5.94 & 35.73 \\
\midrule
 & BAGEL~\citeyearpar{bagel} & 7B MoT & $448\times448$ & 36.60 & 0.263 & 3.76 & 29.77 \\
 & SenseNova-U1~\citeyearpar{diao2026sensenovau1} & 8B MoT & $1024\times1024$ & 18.09 & 0.289 & 7.32 & 35.85 \\
\rowcolor{bagelrow}
\multirow{-3}{*}{1,024} & \textbf{\methodname{}} & 9B MoT & $1024\times1024$ & 22.19 & 0.292 & 6.02 & 36.03 \\
\bottomrule
\end{tabularx}
\caption{\textbf{Visual understanding efficiency at matched visual-token budgets.}}
\label{tab:understanding_throughput}
\end{table}

\revised{\tabref{tab:understanding_throughput} places \methodname{} between
BAGEL and SenseNova-U1 in decoding throughput and total latency at both
visual-token budgets. At 1,024 tokens, it decodes at 22.19 tokens/s and
finishes in 6.02 seconds, compared with 36.60 tokens/s and 3.76 seconds
for BAGEL, and 18.09 tokens/s and 7.32 seconds for SenseNova-U1.
Increasing the visual budget from 256 to 1,024 tokens changes its total
latency only from 5.94 to 6.02 seconds, although first-logit latency rises
from 0.143 to 0.292 seconds. For these fixed 128-token outputs, decoding
therefore accounts for most of the request time. Peak allocated memory
is slightly higher than SenseNova-U1 and remains above BAGEL at both
budgets. These measurements describe computational cost rather than
answer quality.}

\revised{The two tables distinguish generation speed from autoregressive
decoding speed: \methodname{} reduces generation latency relative to
BAGEL at the shared resolutions, whereas BAGEL remains faster and uses
less allocated memory for understanding. Matched workloads control
sequence lengths, but model sizes, native tokenization, vision encoders,
and implementations still differ. These system-level measurements do
not isolate the effect of hybrid attention.}

\FloatBarrier

\section{Conclusion}
\label{sec:conclusion}

\revised{In this work, we present \methodname{}, a unified multimodal model
built on a \textbf{Hybrid MoT Architecture}. Specialized understanding and
generation experts share multimodal context through full-attention and
recurrent linear-attention layers. This architecture combines explicit
visual interactions with compact recurrent context in the same backbone.}
Joint language modeling and visual flow matching connect image
understanding, generation, and instruction-based editing, while clean
visual context links successive outputs in interleaved sequences.
Diverse multimodal supervision and progressive resolution training extend
this framework to a broad range of tasks and image resolutions up to 2K.
Experiments demonstrate strong multimodal understanding, and visual examples
illustrate generation, editing, and interleaved prediction within the unified
framework.
\revised{Our results provide empirical evidence that hybrid attention can
support both multimodal understanding and visual generation within a
unified model. The context-handling design, training recipe, and evaluation
results offer a practical reference for further studies of hybrid
architectures for unified multimodal modeling.}


\begin{thebibliography}{77}
\providecommand{\natexlab}[1]{#1}
\providecommand{\url}[1]{\texttt{#1}}
\expandafter\ifx\csname urlstyle\endcsname\relax
  \providecommand{\doi}[1]{doi: #1}\else
  \providecommand{\doi}{doi: \begingroup \urlstyle{rm}\Url}\fi

\bibitem[Bai et~al.(2026)Bai, Wang, Ouyang, Yu, Wang, Wang, Cheng, Ma, Zeng,
  Liu, Xu, Shen, and Chen]{bai2026scaling}
Qingyan Bai, Qiuyu Wang, Hao Ouyang, Yue Yu, Hanlin Wang, Wen Wang, Ka~Leong
  Cheng, Shuailei Ma, Yanhong Zeng, Zichen Liu, Yinghao Xu, Yujun Shen, and
  Qifeng Chen.
\newblock Scaling instruction-based video editing with a high-quality synthetic
  dataset.
\newblock In \emph{Proceedings of the IEEE/CVF Conference on Computer Vision
  and Pattern Recognition}, 2026.

\bibitem[Bai et~al.(2025)Bai, Chen, Liu, Wang, Ge, Song,
  et~al.]{bai2025qwen25vl}
Shuai Bai, Keqin Chen, Xuejing Liu, Jialin Wang, Wenbin Ge, Sibo Song, et~al.
\newblock {Qwen2.5-VL} technical report.
\newblock \emph{arXiv preprint arXiv:2502.13923}, 2025.
\newblock \doi{10.48550/arXiv.2502.13923}.
\newblock URL \url{https://arxiv.org/abs/2502.13923}.

\bibitem[{Black Forest Labs}(2024)]{blackforestlabs2024flux1}
{Black Forest Labs}.
\newblock {FLUX.1 [dev] Model Card}.
\newblock \url{https://huggingface.co/black-forest-labs/FLUX.1-dev}, 2024.

\bibitem[{Black Forest Labs} et~al.(2025){Black Forest Labs}, Batifol,
  Blattmann, Boesel, Consul, Diagne, Dockhorn, English, English, Esser,
  et~al.]{blackforestlabs2025kontext}
{Black Forest Labs}, Stephen Batifol, Andreas Blattmann, Frederic Boesel,
  Saksham Consul, Cyril Diagne, Tim Dockhorn, Jack English, Zion English,
  Patrick Esser, et~al.
\newblock {FLUX.1 Kontext: Flow Matching for In-Context Image Generation and
  Editing in Latent Space}.
\newblock \emph{arXiv preprint arXiv:2506.15742}, 2025.
\newblock URL \url{https://arxiv.org/abs/2506.15742}.

\bibitem[{Chameleon Team}(2024)]{team2024chameleon}
{Chameleon Team}.
\newblock {Chameleon: Mixed-Modal Early-Fusion Foundation Models}.
\newblock \emph{arXiv preprint arXiv:2405.09818}, 2024.
\newblock URL \url{https://arxiv.org/abs/2405.09818}.

\bibitem[Chen et~al.(2026)Chen, Yu, Li, Xue, Liu, Xin, Zhao, Ye, Wu, Wang,
  Zhou, Luo, Han, and Xie]{chen2026sanavideo2}
Junsong Chen, Jincheng Yu, Yitong Li, Shuchen Xue, Haozhe Liu, Jingyu Xin,
  Yuyang Zhao, Tian Ye, Zhangjie Wu, Zian Wang, Daquan Zhou, Ping Luo, Song
  Han, and Enze Xie.
\newblock {SANA-Video 2.0}: Hybrid linear attention with attention residuals
  for efficient video generation.
\newblock \emph{arXiv preprint arXiv:2607.21553}, 2026.
\newblock \doi{10.48550/arXiv.2607.21553}.
\newblock URL \url{https://arxiv.org/abs/2607.21553}.

\bibitem[Chen et~al.(2016)Chen, Xu, Zhang, and Guestrin]{chen2016checkpointing}
Tianqi Chen, Bing Xu, Chiyuan Zhang, and Carlos Guestrin.
\newblock Training deep nets with sublinear memory cost.
\newblock \emph{arXiv preprint arXiv:1604.06174}, 2016.
\newblock URL \url{https://arxiv.org/abs/1604.06174}.

\bibitem[Chen et~al.(2025)Chen, Wu, Liu, Pan, Liu, Xie, Yu, and
  Ruan]{chen2025janus}
Xiaokang Chen, Zhiyu Wu, Xingchao Liu, Zizheng Pan, Wen Liu, Zhenda Xie,
  Xingkai Yu, and Chong Ruan.
\newblock {Janus-Pro: Unified Multimodal Understanding and Generation with Data
  and Model Scaling}.
\newblock \emph{arXiv preprint arXiv:2501.17811}, 2025.
\newblock URL \url{https://arxiv.org/abs/2501.17811}.

\bibitem[Chen et~al.(2024)Chen, Wang, Cao, Liu, Gao, Cui, Zhu, Ye, Tian, Liu,
  et~al.]{chen2024internvl25}
Zhe Chen, Weiyun Wang, Yue Cao, Yangzhou Liu, Zhangwei Gao, Erfei Cui, Jinguo
  Zhu, Shenglong Ye, Hao Tian, Zhaoyang Liu, et~al.
\newblock {Expanding Performance Boundaries of Open-Source Multimodal Models
  with Model, Data, and Test-Time Scaling}.
\newblock \emph{arXiv preprint arXiv:2412.05271}, 2024.
\newblock URL \url{https://arxiv.org/abs/2412.05271}.

\bibitem[Deng et~al.(2025)Deng, Zhu, Li, Gou, Li, Wang, Zhong, Yu, Nie, Song,
  et~al.]{bagel}
Chaorui Deng, Deyao Zhu, Kunchang Li, Chenhui Gou, Feng Li, Zeyu Wang, Shu
  Zhong, Weihao Yu, Xiaonan Nie, Ziang Song, et~al.
\newblock Emerging properties in unified multimodal pretraining.
\newblock \emph{arXiv preprint arXiv:2505.14683}, 2025.
\newblock \doi{10.48550/arXiv.2505.14683}.
\newblock URL \url{https://arxiv.org/abs/2505.14683}.

\bibitem[Diao et~al.(2026)Diao, Wu, Deng, Wang, Bai, Wu, Fan, Ye, Tong, Fan,
  Li, Wang, Cao, Lin, Yang, Cai, Niu, Zhu, Liu, Lv, Yu, Xie, Wang, Fan, Li, Lu,
  Ni, Xu, Liang, Shi, Dai, Wang, Qian, Gao, Liu, Sun, Shen, Wang, Ma, Yang,
  Xie, Li, Zhong, Kong, Shi, Gao, Yao, Wang, Bai, Lin, Yin, Sun, Gong, Wang,
  Lu, Yang, Liu, and Lin]{diao2026sensenovau1}
Haiwen Diao, Penghao Wu, Hanming Deng, Jiahao Wang, Shihao Bai, Silei Wu,
  Weichen Fan, Wenjie Ye, Wenwen Tong, Xiangyu Fan, Yan Li, Yubo Wang, Zhijie
  Cao, Zhiqian Lin, Zhitao Yang, Zhongang Cai, Yuwei Niu, Yue Zhu, Bo~Liu,
  Chengguang Lv, Haojia Yu, Haozhe Xie, Hongli Wang, Jianan Fan, Jiaqi Li,
  Jiefan Lu, Jingcheng Ni, Junxiang Xu, Kaihuan Liang, Lianqiang Shi, Linjun
  Dai, Linyan Wang, Oscar Qian, Peng Gao, Pengfei Liu, Qingping Sun, Rui Shen,
  Ruisi Wang, Shengnan Ma, Shuang Yang, Siyi Xie, Siying Li, Tianbo Zhong,
  Xiangli Kong, Xuanke Shi, Yang Gao, Yongqiang Yao, Yves Wang, Zhengqi Bai,
  Zhengyu Lin, Zixin Yin, Wenxiu Sun, Ruihao Gong, Quan Wang, Lewei Lu, Lei
  Yang, Ziwei Liu, and Dahua Lin.
\newblock {SenseNova-U1: Unifying Multimodal Understanding and Generation with
  NEO-unify Architecture}.
\newblock \emph{arXiv preprint arXiv:2605.12500}, 2026.
\newblock \doi{10.48550/arXiv.2605.12500}.
\newblock URL \url{https://arxiv.org/abs/2605.12500}.

\bibitem[Fu et~al.(2023)Fu, Chen, Shen, Qin, Zhang, Lin, Qiu, Lin, Yang, Zheng,
  Li, Sun, and Ji]{fu2023mme}
Chaoyou Fu, Peixian Chen, Yunhang Shen, Yulei Qin, Mengdan Zhang, Xu~Lin,
  Zhenyu Qiu, Wei Lin, Jinrui Yang, Xiawu Zheng, Ke~Li, Xing Sun, and Rongrong
  Ji.
\newblock {MME: A Comprehensive Evaluation Benchmark for Multimodal Large
  Language Models}.
\newblock \emph{arXiv preprint arXiv:2306.13394}, 2023.
\newblock \doi{10.48550/arXiv.2306.13394}.
\newblock URL \url{https://arxiv.org/abs/2306.13394v1}.

\bibitem[Gao et~al.(2025)Gao, Gong, Guo, Hou, Lai, Li, Li, Lian, Liao, Liu,
  et~al.]{gao2025seedream}
Yu~Gao, Lixue Gong, Qiushan Guo, Xiaoxia Hou, Zhichao Lai, Fanshi Li, Liang Li,
  Xiaochen Lian, Chao Liao, Liyang Liu, et~al.
\newblock {Seedream 3.0 Technical Report}.
\newblock \emph{arXiv preprint arXiv:2504.11346}, 2025.
\newblock URL \url{https://arxiv.org/abs/2504.11346}.

\bibitem[Ge et~al.(2024)Ge, Zhao, Zhu, Ge, Yi, Song, Li, Ding, and
  Shan]{ge2024seedx}
Yuying Ge, Sijie Zhao, Jinguo Zhu, Yixiao Ge, Kun Yi, Lin Song, Chen Li,
  Xiaohan Ding, and Ying Shan.
\newblock {SEED-X: Multimodal Models with Unified Multi-granularity
  Comprehension and Generation}.
\newblock \emph{arXiv preprint arXiv:2404.14396}, 2024.
\newblock URL \url{https://arxiv.org/abs/2404.14396}.

\bibitem[Ghosh et~al.(2023)Ghosh, Hajishirzi, and Schmidt]{ghosh2023geneval}
Dhruba Ghosh, Hanna Hajishirzi, and Ludwig Schmidt.
\newblock {GenEval: An Object-Focused Framework for Evaluating Text-to-Image
  Alignment}.
\newblock \emph{arXiv preprint arXiv:2310.11513}, 2023.
\newblock URL \url{https://arxiv.org/abs/2310.11513}.

\bibitem[Han et~al.(2026)Han, Li, Deng, Chen, Shi, Wang, Li, Wang, Xie, You,
  Quan, Cai, Diao, Liu, Yang, Lin, and Wang]{diao2026sensenovavision}
Xiaoyang Han, Jianhua Li, Kewang Deng, Zukai Chen, Xuanke Shi, Sihan Wang,
  Boxuan Li, Linyan Wang, Siyi Xie, Xin You, Jinsheng Quan, Zhongang Cai,
  Haiwen Diao, Ziwei Liu, Lei Yang, Dahua Lin, and Quan Wang.
\newblock {Vision as Unified Multimodal Generation}.
\newblock \emph{arXiv preprint arXiv:2607.06560}, 2026.
\newblock \doi{10.48550/arXiv.2607.06560}.
\newblock URL \url{https://arxiv.org/abs/2607.06560}.

\bibitem[Hatamizadeh et~al.(2026)Hatamizadeh, Choi, and
  Kautz]{gateddelta2026two}
Ali Hatamizadeh, Yejin Choi, and Jan Kautz.
\newblock {Gated DeltaNet-2: Decoupling Erase and Write in Linear Attention}.
\newblock \emph{arXiv preprint arXiv:2605.22791}, 2026.
\newblock \doi{10.48550/arXiv.2605.22791}.
\newblock URL \url{https://arxiv.org/abs/2605.22791}.

\bibitem[Hu et~al.(2024)Hu, Wang, Fang, Fu, Cheng, and Yu]{hu2024ella}
Xiwei Hu, Rui Wang, Yixiao Fang, Bin Fu, Pei Cheng, and Gang Yu.
\newblock {ELLA: Equip Diffusion Models with LLM for Enhanced Semantic
  Alignment}.
\newblock \emph{arXiv preprint arXiv:2403.05135}, 2024.
\newblock URL \url{https://arxiv.org/abs/2403.05135}.

\bibitem[{Inclusion AI} et~al.(2026){Inclusion AI}, Bie, Chen, Chen, Cheng,
  Cui, Gan, Huang, Lan, Li, et~al.]{ai2026llada2}
{Inclusion AI}, Tiwei Bie, Haoxing Chen, Tieyuan Chen, Zhenglin Cheng, Long
  Cui, Kai Gan, Zhicheng Huang, Zhenzhong Lan, Haoquan Li, et~al.
\newblock {LLaDA2.0-Uni: Unifying Multimodal Understanding and Generation with
  Diffusion Large Language Model}.
\newblock \emph{arXiv preprint arXiv:2604.20796}, 2026.
\newblock URL \url{https://arxiv.org/abs/2604.20796}.

\bibitem[Katharopoulos et~al.(2020)Katharopoulos, Vyas, Pappas, and
  Fleuret]{katharopoulos2020linear}
Angelos Katharopoulos, Apoorv Vyas, Nikolaos Pappas, and Fran{\c{c}}ois
  Fleuret.
\newblock Transformers are {RNN}s: Fast autoregressive transformers with linear
  attention.
\newblock In \emph{Proceedings of the 37th International Conference on Machine
  Learning}, volume 119, pp.\  5156--5165, 2020.
\newblock URL \url{https://proceedings.mlr.press/v119/katharopoulos20a.html}.

\bibitem[{Kimi Team} et~al.(2025{\natexlab{a}}){Kimi Team}, Du, Yin, Xing, Qu,
  Wang, Chen, Zhang, Du, Wei, et~al.]{kimi2025vl}
{Kimi Team}, Angang Du, Bohong Yin, Bowei Xing, Bowen Qu, Bowen Wang, Cheng
  Chen, Chenlin Zhang, Chenzhuang Du, Chu Wei, et~al.
\newblock {Kimi-VL Technical Report}.
\newblock \emph{arXiv preprint arXiv:2504.07491}, 2025{\natexlab{a}}.
\newblock URL \url{https://arxiv.org/abs/2504.07491}.

\bibitem[{Kimi Team} et~al.(2025{\natexlab{b}}){Kimi Team}, Zhang, Lin, Yao,
  Hu, Meng, Liu, Men, Yang, Li, Li, Lu, Liu, Chen, Xu, Yu, Wang, Fan, Zhong,
  Yuan, Zhang, Zhang, Liu, Wang, Fang, He, Liu, Li, Su, Qiu, Pang, Yan, Jiang,
  Huang, Yin, You, Wei, Wang, Hong, Chen, Chen, Wang, Zheng, Wang, Liu, Dong,
  Zhang, Pan, Wu, Wu, Guan, Tao, Fu, Xu, Wang, Lai, Wu, Zhou, Yang, and
  Du]{kimi2025linear}
{Kimi Team}, Yu~Zhang, Zongyu Lin, Xingcheng Yao, Jiaxi Hu, Fanqing Meng,
  Chengyin Liu, Xin Men, Songlin Yang, Zhiyuan Li, Wentao Li, Enzhe Lu, Weizhou
  Liu, Yanru Chen, Weixin Xu, Longhui Yu, Yejie Wang, Yu~Fan, Longguang Zhong,
  Enming Yuan, Dehao Zhang, Yizhi Zhang, T.~Y. Liu, Haiming Wang, Shengjun
  Fang, Weiran He, Shaowei Liu, Yiwei Li, Jianlin Su, Jiezhong Qiu, Bo~Pang,
  Junjie Yan, Zhejun Jiang, Weixiao Huang, Bohong Yin, Jiacheng You, Chu Wei,
  Zhengtao Wang, Chao Hong, Yutian Chen, Guanduo Chen, Yucheng Wang, Huabin
  Zheng, Feng Wang, Yibo Liu, Mengnan Dong, Zheng Zhang, Siyuan Pan, Wenhao Wu,
  Yuhao Wu, Longyu Guan, Jiawen Tao, Guohong Fu, Xinran Xu, Yuzhi Wang, Guokun
  Lai, Yuxin Wu, Xinyu Zhou, Zhilin Yang, and Yulun Du.
\newblock {Kimi Linear: An Expressive, Efficient Attention Architecture}.
\newblock \emph{arXiv preprint arXiv:2510.26692}, 2025{\natexlab{b}}.
\newblock \doi{10.48550/arXiv.2510.26692}.
\newblock URL \url{https://arxiv.org/abs/2510.26692}.

\bibitem[Li et~al.(2024)Li, Zhang, Guo, Zhang, Li, Zhang, Zhang, Zhang, Li,
  Liu, et~al.]{li2024llavaonevision}
Bo~Li, Yuanhan Zhang, Dong Guo, Renrui Zhang, Feng Li, Hao Zhang, Kaichen
  Zhang, Peiyuan Zhang, Yanwei Li, Ziwei Liu, et~al.
\newblock {LLaVA-OneVision: Easy Visual Task Transfer}.
\newblock \emph{arXiv preprint arXiv:2408.03326}, 2024.
\newblock URL \url{https://arxiv.org/abs/2408.03326}.

\bibitem[Li et~al.(2026)Li, Luo, Zhang, Gao, Li, and Hu]{li2026solavision}
Ruibang Li, Guan Luo, Yiwei Zhang, Jin Gao, Bing Li, and Weiming Hu.
\newblock {SoLA-Vision}: Fine-grained layer-wise linear softmax hybrid
  attention.
\newblock \emph{arXiv preprint arXiv:2601.11164}, 2026.
\newblock \doi{10.48550/arXiv.2601.11164}.
\newblock URL \url{https://arxiv.org/abs/2601.11164}.

\bibitem[Liang et~al.(2025)Liang, Yu, Luo, Iyer, Dong, Zhou, Ghosh, Lewis, Yih,
  Zettlemoyer, and Lin]{liang2024mixture}
Weixin Liang, Lili Yu, Liang Luo, Srinivasan Iyer, Ning Dong, Chunting Zhou,
  Gargi Ghosh, Mike Lewis, Wen-tau Yih, Luke Zettlemoyer, and Xi~Victoria Lin.
\newblock Mixture-of-transformers: A sparse and scalable architecture for
  multi-modal foundation models.
\newblock \emph{Transactions on Machine Learning Research}, 2025.
\newblock \doi{10.48550/arXiv.2411.04996}.
\newblock URL \url{https://arxiv.org/abs/2411.04996}.

\bibitem[Lipman et~al.(2023)Lipman, Chen, Ben-Hamu, Nickel, and
  Le]{lipman2023flowmatching}
Yaron Lipman, Ricky T.~Q. Chen, Heli Ben-Hamu, Maximilian Nickel, and Matt Le.
\newblock Flow matching for generative modeling.
\newblock In \emph{International Conference on Learning Representations}, 2023.
\newblock URL \url{https://arxiv.org/abs/2210.02747}.

\bibitem[Liu et~al.(2025)Liu, Han, Xing, Yin, Wang, Cheng, Liao, Wang, Fu, Han,
  et~al.]{liu2025step1xedit}
Shiyu Liu, Yucheng Han, Peng Xing, Fukun Yin, Rui Wang, Wei Cheng, Jiaqi Liao,
  Yingming Wang, Honghao Fu, Chunrui Han, et~al.
\newblock {Step1X-Edit: A Practical Framework for General Image Editing}.
\newblock \emph{arXiv preprint arXiv:2504.17761}, 2025.
\newblock URL \url{https://arxiv.org/abs/2504.17761}.

\bibitem[Liu et~al.(2024)Liu, Duan, Zhang, Li, Zhang, Zhao, Yuan, Wang, He,
  Liu, et~al.]{Mmbench}
Yuan Liu, Haodong Duan, Yuanhan Zhang, Bo~Li, Songyang Zhang, Wangbo Zhao, Yike
  Yuan, Jiaqi Wang, Conghui He, Ziwei Liu, et~al.
\newblock {MMBench: Is Your Multi-Modal Model an All-around Player?}
\newblock In \emph{Proceedings of the European Conference on Computer Vision
  (ECCV)}, 2024.
\newblock URL \url{https://arxiv.org/abs/2307.06281}.

\bibitem[Liu et~al.(2026)Liu, Ren, Huang, Chen, Li, Chen, Ji, He, Schult, Zeng,
  Xiang, Chen, Luo, Zettlemoyer, and Cong]{liu2026tuna2}
Zhiheng Liu, Weiming Ren, Xiaoke Huang, Shoufa Chen, Tianhong Li, Mengzhao
  Chen, Yatai Ji, Sen He, Jonas Schult, Belinda Zeng, Tao Xiang, Wenhu Chen,
  Ping Luo, Luke Zettlemoyer, and Yuren Cong.
\newblock {Tuna-2: Pixel Embeddings Beat Vision Encoders for Multimodal
  Understanding and Generation}.
\newblock \emph{arXiv preprint arXiv:2604.24763}, 2026.
\newblock \doi{10.48550/arXiv.2604.24763}.
\newblock URL \url{https://arxiv.org/abs/2604.24763}.

\bibitem[Loshchilov \& Hutter(2019)Loshchilov and Hutter]{loshchilov2019adamw}
Ilya Loshchilov and Frank Hutter.
\newblock Decoupled weight decay regularization.
\newblock In \emph{International Conference on Learning Representations}, 2019.
\newblock URL \url{https://arxiv.org/abs/1711.05101}.

\bibitem[Lu et~al.(2023)Lu, Bansal, Xia, Liu, Li, Hajishirzi, Cheng, Chang,
  Galley, and Gao]{mathvista}
Pan Lu, Hritik Bansal, Tony Xia, Jiacheng Liu, Chunyuan Li, Hannaneh
  Hajishirzi, Hao Cheng, Kai-Wei Chang, Michel Galley, and Jianfeng Gao.
\newblock {MathVista: Evaluating Mathematical Reasoning of Foundation Models in
  Visual Contexts}.
\newblock \emph{arXiv preprint arXiv:2310.02255}, 2023.
\newblock URL \url{https://arxiv.org/abs/2310.02255}.

\bibitem[{Meituan LongCat Team} et~al.(2025){Meituan LongCat Team}, Ma, Tan,
  Huang, Wu, He, Gao, Xiao, Wei, Ma, et~al.]{team2025longcat}
{Meituan LongCat Team}, Hanghang Ma, Haoxian Tan, Jiale Huang, Junqiang Wu,
  Jun-Yan He, Lishuai Gao, Songlin Xiao, Xiaoming Wei, Xiaoqi Ma, et~al.
\newblock {LongCat-Image Technical Report}.
\newblock \emph{arXiv preprint arXiv:2512.07584}, 2025.
\newblock URL \url{https://arxiv.org/abs/2512.07584}.

\bibitem[Meng et~al.(2026)Meng, Bao, Liu, Zhu, Wang, Li, and Pan]{harts2026}
Boyuan Meng, Peihua Bao, Hong Liu, Xiaowei Zhu, Chao Wang, Gen Li, and Zhenxuan
  Pan.
\newblock {HARTS: Efficient Agentic Reinforcement Learning for Hybrid-Attention
  Models over Arbitrary Rollout Trees}.
\newblock \emph{arXiv preprint arXiv:2608.28158}, 2026.
\newblock \doi{10.48550/arXiv.2608.28158}.
\newblock URL \url{https://arxiv.org/abs/2608.28158}.

\bibitem[Merrill et~al.(2026)Merrill, Li, Romero, Svete, Costello, Dasigi,
  Groeneveld, Heineman, Kuehl, Lambert, Li, Lo, Malik, Matusz, Minixhofer,
  Morrison, Soldaini, Timbers, Walsh, Smith, Hajishirzi, and
  Sabharwal]{merrill2026olmohybrid}
William Merrill, Yanhong Li, Tyler Romero, Anej Svete, Caia Costello, Pradeep
  Dasigi, Dirk Groeneveld, David Heineman, Bailey Kuehl, Nathan Lambert, Chuan
  Li, Kyle Lo, Saumya Malik, DJ~Matusz, Benjamin Minixhofer, Jacob Morrison,
  Luca Soldaini, Finbarr Timbers, Pete Walsh, Noah~A. Smith, Hannaneh
  Hajishirzi, and Ashish Sabharwal.
\newblock {Olmo Hybrid: From Theory to Practice and Back}.
\newblock \emph{arXiv preprint arXiv:2604.03444}, 2026.
\newblock \doi{10.48550/arXiv.2604.03444}.
\newblock URL \url{https://arxiv.org/abs/2604.03444}.

\bibitem[{OpenGVLab}(2024)]{opengvlab2024internvl2}
{OpenGVLab}.
\newblock {InternVL2 Model Card}.
\newblock \url{https://huggingface.co/OpenGVLab/InternVL2-8B}, 2024.

\bibitem[Pan et~al.(2025)Pan, Shukla, Singh, Zhao, Mishra, Wang, Xu, Chen, Li,
  Juefei-Xu, et~al.]{pan2025metaquery}
Xichen Pan, Satya~Narayan Shukla, Aashu Singh, Zhuokai Zhao, Shlok~Kumar
  Mishra, Jialiang Wang, Zhiyang Xu, Jiuhai Chen, Kunpeng Li, Felix Juefei-Xu,
  et~al.
\newblock {Transfer between Modalities with MetaQueries}.
\newblock \emph{arXiv preprint arXiv:2504.06256}, 2025.
\newblock URL \url{https://arxiv.org/abs/2504.06256}.

\bibitem[Peng et~al.(2026)Peng, Meng, Cai, Zhuang, Yang, Fang, Wu, Lin, Wu, and
  Bai]{peng2026uniar}
Wujian Peng, Lingchen Meng, Yuxuan Cai, Xianwei Zhuang, Yuhuan Yang, Rongyao
  Fang, Chenfei Wu, Junyang Lin, Zuxuan Wu, and Shuai Bai.
\newblock {Unified Multimodal Autoregressive Modeling with Shared
  Context-Visual Tokenizer is Key to Unification}.
\newblock \emph{arXiv preprint arXiv:2606.18249}, 2026.
\newblock \doi{10.48550/arXiv.2606.18249}.
\newblock URL \url{https://arxiv.org/abs/2606.18249}.

\bibitem[Qu et~al.(2024)Qu, Zhang, Liu, Wang, Jiang, Gao, Ye, Du, Yuan, and
  Wu]{qu2024tokenflow}
Liao Qu, Huichao Zhang, Yiheng Liu, Xu~Wang, Yi~Jiang, Yiming Gao, Hu~Ye,
  Daniel~K. Du, Zehuan Yuan, and Xinglong Wu.
\newblock {TokenFlow: Unified Image Tokenizer for Multimodal Understanding and
  Generation}.
\newblock \emph{arXiv preprint arXiv:2412.03069}, 2024.
\newblock URL \url{https://arxiv.org/abs/2412.03069}.

\bibitem[{Qwen Team}(2025)]{qwen2025imageedit}
{Qwen Team}.
\newblock {Qwen-Image-Edit Model Card}.
\newblock \url{https://huggingface.co/Qwen/Qwen-Image-Edit}, 2025.

\bibitem[{Qwen Team}(2026)]{qwen2026qwen35modelcard}
{Qwen Team}.
\newblock {Qwen3.5-9B} model card.
\newblock Hugging Face, 2026.
\newblock URL \url{https://huggingface.co/Qwen/Qwen3.5-9B}.
\newblock Accessed September 6, 2026.

\bibitem[Rombach et~al.(2022)Rombach, Blattmann, Lorenz, Esser, and
  Ommer]{rombach2022latentdiffusion}
Robin Rombach, Andreas Blattmann, Dominik Lorenz, Patrick Esser, and Bj{\"o}rn
  Ommer.
\newblock High-resolution image synthesis with latent diffusion models.
\newblock In \emph{Proceedings of the IEEE/CVF Conference on Computer Vision
  and Pattern Recognition}, 2022.
\newblock URL \url{https://arxiv.org/abs/2112.10752}.

\bibitem[Shi et~al.(2024)Shi, Han, Zhou, Liang, Lin, Zettlemoyer, and
  Yu]{shi2024llamafusion}
Weijia Shi, Xiaochuang Han, Chunting Zhou, Weixin Liang, Xi~Victoria Lin, Luke
  Zettlemoyer, and Lili Yu.
\newblock {LMFusion: Adapting Pretrained Language Models for Multimodal
  Generation}.
\newblock \emph{arXiv preprint arXiv:2412.15188}, 2024.
\newblock URL \url{https://arxiv.org/abs/2412.15188}.

\bibitem[Tan et~al.(2026)Tan, Chen, Shen, Liu, Shen, Wu, and Ye]{hydrahead2026}
Zhentao Tan, Wei Chen, Jingyi Shen, Yao Liu, Xu~Shen, Yue Wu, and Jieping Ye.
\newblock {HydraHead: From Head-Level Functional Heterogeneity to Specialized
  Attention Hybridization}.
\newblock \emph{arXiv preprint arXiv:2606.20097}, 2026.
\newblock \doi{10.48550/arXiv.2606.20097}.
\newblock URL \url{https://arxiv.org/abs/2606.20097}.

\bibitem[{Tencent Hunyuan Foundation Model Team}(2025)]{cao2025hunyuanimage}
{Tencent Hunyuan Foundation Model Team}.
\newblock {HunyuanImage 3.0 Technical Report}.
\newblock \emph{arXiv preprint arXiv:2509.23951}, 2025.
\newblock URL \url{https://arxiv.org/abs/2509.23951}.

\bibitem[Tian et~al.(2026)Tian, Yang, Chen, Cui, Wang, Duan, Yin, Chen, Yang,
  Liu, et~al.]{tian2026internvl}
Changyao Tian, Danni Yang, Guanzhou Chen, Erfei Cui, Zhaokai Wang, Yuchen Duan,
  Penghao Yin, Sitao Chen, Ganlin Yang, Mingxin Liu, et~al.
\newblock {InternVL-U: Democratizing Unified Multimodal Models for
  Understanding, Reasoning, Generation and Editing}.
\newblock \emph{arXiv preprint arXiv:2603.09877}, 2026.
\newblock URL \url{https://arxiv.org/abs/2603.09877}.

\bibitem[Tong et~al.(2024{\natexlab{a}})Tong, Fan, Zhu, Xiong, Chen, Sinha,
  Rabbat, LeCun, Xie, and Liu]{tong2024metamorph}
Shengbang Tong, David Fan, Jiachen Zhu, Yunyang Xiong, Xinlei Chen, Koustuv
  Sinha, Michael Rabbat, Yann LeCun, Saining Xie, and Zhuang Liu.
\newblock {MetaMorph: Multimodal Understanding and Generation via Instruction
  Tuning}.
\newblock \emph{arXiv preprint arXiv:2412.14164}, 2024{\natexlab{a}}.
\newblock URL \url{https://arxiv.org/abs/2412.14164}.

\bibitem[Tong et~al.(2024{\natexlab{b}})Tong, Liu, Zhai, Ma, LeCun, and
  Xie]{tong2024eyes}
Shengbang Tong, Zhuang Liu, Yuexiang Zhai, Yi~Ma, Yann LeCun, and Saining Xie.
\newblock {Eyes Wide Shut? Exploring the Visual Shortcomings of Multimodal
  LLMs}.
\newblock \emph{arXiv preprint arXiv:2401.06209}, 2024{\natexlab{b}}.
\newblock \doi{10.48550/arXiv.2401.06209}.
\newblock URL \url{https://arxiv.org/abs/2401.06209}.

\bibitem[Vaswani et~al.(2017)Vaswani, Shazeer, Parmar, Uszkoreit, Jones, Gomez,
  Kaiser, and Polosukhin]{vaswani2017attention}
Ashish Vaswani, Noam Shazeer, Niki Parmar, Jakob Uszkoreit, Llion Jones,
  Aidan~N. Gomez, Lukasz Kaiser, and Illia Polosukhin.
\newblock Attention is all you need.
\newblock In \emph{Advances in Neural Information Processing Systems}, 2017.
\newblock URL \url{https://arxiv.org/abs/1706.03762}.

\bibitem[{Wan Team}(2025)]{wan2025wan22}
{Wan Team}.
\newblock {Wan2.2: Open and Advanced Large-Scale Video Generative Models}.
\newblock \url{https://github.com/Wan-Video/Wan2.2}, 2025.

\bibitem[Wang et~al.(2024{\natexlab{a}})Wang, Lu, Yang, Huang, Han, Hou, Zhang,
  and Xu]{wang2024illume}
Chunwei Wang, Guansong Lu, Junwei Yang, Runhui Huang, Jianhua Han, Lu~Hou, Wei
  Zhang, and Hang Xu.
\newblock {ILLUME: Illuminating Your LLMs to See, Draw, and Self-Enhance}.
\newblock \emph{arXiv preprint arXiv:2412.06673}, 2024{\natexlab{a}}.
\newblock URL \url{https://arxiv.org/abs/2412.06673}.

\bibitem[Wang et~al.(2024{\natexlab{b}})Wang, Bai, Tan, Wang, Fan, Bai, Chen,
  Liu, Wang, Ge, et~al.]{wang2024qwen2vl}
Peng Wang, Shuai Bai, Sinan Tan, Shijie Wang, Zhihao Fan, Jinze Bai, Keqin
  Chen, Xuejing Liu, Jialin Wang, Wenbin Ge, et~al.
\newblock {Qwen2-VL: Enhancing Vision-Language Model's Perception of the World
  at Any Resolution}.
\newblock \emph{arXiv preprint arXiv:2409.12191}, 2024{\natexlab{b}}.
\newblock URL \url{https://arxiv.org/abs/2409.12191}.

\bibitem[Wang et~al.(2024{\natexlab{c}})Wang, Zhang, Luo, Sun, Cui, Wang,
  Zhang, Wang, Li, Yu, et~al.]{wang2024emu3}
Xinlong Wang, Xiaosong Zhang, Zhengxiong Luo, Quan Sun, Yufeng Cui, Jinsheng
  Wang, Fan Zhang, Yueze Wang, Zhen Li, Qiying Yu, et~al.
\newblock {Emu3: Next-Token Prediction Is All You Need}.
\newblock \emph{arXiv preprint arXiv:2409.18869}, 2024{\natexlab{c}}.
\newblock URL \url{https://arxiv.org/abs/2409.18869}.

\bibitem[Williams \& Zipser(1989)Williams and Zipser]{williams1989teacher}
Ronald~J. Williams and David Zipser.
\newblock A learning algorithm for continually running fully recurrent neural
  networks.
\newblock \emph{Neural Computation}, 1\penalty0 (2):\penalty0 270--280, 1989.
\newblock \doi{10.1162/neco.1989.1.2.270}.

\bibitem[Wu et~al.(2025{\natexlab{a}})Wu, Li, Zhou, Lin, Gao, Yan, Yin, Bai,
  Xu, Chen, et~al.]{wu2025qwen}
Chenfei Wu, Jiahao Li, Jingren Zhou, Junyang Lin, Kaiyuan Gao, Kun Yan,
  Sheng-ming Yin, Shuai Bai, Xiao Xu, Yilei Chen, et~al.
\newblock {Qwen-Image Technical Report}.
\newblock \emph{arXiv preprint arXiv:2508.02324}, 2025{\natexlab{a}}.
\newblock URL \url{https://arxiv.org/abs/2508.02324}.

\bibitem[Wu et~al.(2024{\natexlab{a}})Wu, Chen, Wu, Ma, Liu, Pan, Liu, Xie, Yu,
  Ruan, et~al.]{wu2024janus}
Chengyue Wu, Xiaokang Chen, Zhiyu Wu, Yiyang Ma, Xingchao Liu, Zizheng Pan, Wen
  Liu, Zhenda Xie, Xingkai Yu, Chong Ruan, et~al.
\newblock {Janus: Decoupling Visual Encoding for Unified Multimodal
  Understanding and Generation}.
\newblock \emph{arXiv preprint arXiv:2410.13848}, 2024{\natexlab{a}}.
\newblock URL \url{https://arxiv.org/abs/2410.13848}.

\bibitem[Wu et~al.(2025{\natexlab{b}})Wu, Zheng, Yan, Xiao, Luo, Wang, Li,
  Jiang, Liu, Zhou, et~al.]{wu2025omnigen2}
Chenyuan Wu, Pengfei Zheng, Ruiran Yan, Shitao Xiao, Xin Luo, Yueze Wang, Wanli
  Li, Xiyan Jiang, Yexin Liu, Junjie Zhou, et~al.
\newblock {OmniGen2: Towards Instruction-Aligned Multimodal Generation}.
\newblock \emph{arXiv preprint arXiv:2506.18871}, 2025{\natexlab{b}}.
\newblock URL \url{https://arxiv.org/abs/2506.18871}.

\bibitem[Wu et~al.(2026)Wu, Diao, Fan, Lu, Lin, and Liu]{wu2026synergy}
Penghao Wu, Haiwen Diao, Weichen Fan, Lewei Lu, Dahua Lin, and Ziwei Liu.
\newblock {Uncovering Understanding-Generation Synergy in Native Unified
  Multimodal Models: From Representation, Task to System}.
\newblock \emph{arXiv preprint arXiv:2609.01607}, 2026.
\newblock \doi{10.48550/arXiv.2609.01607}.
\newblock URL \url{https://arxiv.org/abs/2609.01607}.

\bibitem[Wu et~al.(2024{\natexlab{b}})Wu, Zhang, Chen, Tang, Li, Fang, Zhu,
  Xie, Yin, Yi, et~al.]{wu2024vilau}
Yecheng Wu, Zhuoyang Zhang, Junyu Chen, Haotian Tang, Dacheng Li, Yunhao Fang,
  Ligeng Zhu, Enze Xie, Hongxu Yin, Li~Yi, et~al.
\newblock {VILA-U: a Unified Foundation Model Integrating Visual Understanding
  and Generation}.
\newblock \emph{arXiv preprint arXiv:2409.04429}, 2024{\natexlab{b}}.
\newblock URL \url{https://arxiv.org/abs/2409.04429}.

\bibitem[Wu et~al.(2024{\natexlab{c}})Wu, Chen, Pan, Liu, Liu, Dai, Gao, Ma,
  Wu, Wang, et~al.]{wu2024deepseekvl2}
Zhiyu Wu, Xiaokang Chen, Zizheng Pan, Xingchao Liu, Wen Liu, Damai Dai, Huazuo
  Gao, Yiyang Ma, Chengyue Wu, Bingxuan Wang, et~al.
\newblock {DeepSeek-VL2: Mixture-of-Experts Vision-Language Models for Advanced
  Multimodal Understanding}.
\newblock \emph{arXiv preprint arXiv:2412.10302}, 2024{\natexlab{c}}.
\newblock URL \url{https://arxiv.org/abs/2412.10302}.

\bibitem[Xie et~al.(2024{\natexlab{a}})Xie, Mao, Bai, Zhang, Wang, Lin, Gu,
  Chen, Yang, and Shou]{xie2024showo}
Jinheng Xie, Weijia Mao, Zechen Bai, David~Junhao Zhang, Weihao Wang,
  Kevin~Qinghong Lin, Yuchao Gu, Zhijie Chen, Zhenheng Yang, and Mike~Zheng
  Shou.
\newblock {Show-o}: One single transformer to unify multimodal understanding
  and generation.
\newblock \emph{arXiv preprint arXiv:2408.12528}, 2024{\natexlab{a}}.
\newblock \doi{10.48550/arXiv.2408.12528}.
\newblock URL \url{https://arxiv.org/abs/2408.12528}.

\bibitem[Xie et~al.(2025)Xie, Yang, and Shou]{xie2025showo2}
Jinheng Xie, Zhenheng Yang, and Mike~Zheng Shou.
\newblock {Show-o2}: Improved native unified multimodal models.
\newblock \emph{arXiv preprint arXiv:2506.15564}, 2025.
\newblock \doi{10.48550/arXiv.2506.15564}.
\newblock URL \url{https://arxiv.org/abs/2506.15564}.

\bibitem[Xie et~al.(2024{\natexlab{b}})Xie, Du, Song, and Liu]{xie2024musevl}
Rongchang Xie, Chen Du, Ping Song, and Chang Liu.
\newblock {MUSE-VL: Modeling Unified VLM through Semantic Discrete Encoding}.
\newblock \emph{arXiv preprint arXiv:2411.17762}, 2024{\natexlab{b}}.
\newblock URL \url{https://arxiv.org/abs/2411.17762}.

\bibitem[Xin et~al.(2025{\natexlab{a}})Xin, Qin, Luo, Zhu, Yan, Tai, Lei, Cao,
  Wang, Wang, et~al.]{dimoo}
Yi~Xin, Qi~Qin, Siqi Luo, Kaiwen Zhu, Juncheng Yan, Yan Tai, Jiayi Lei, Yuewen
  Cao, Keqi Wang, Yibin Wang, et~al.
\newblock {Lumina-DiMOO: An Omni Diffusion Large Language Model for Multi-Modal
  Generation and Understanding}.
\newblock \emph{arXiv preprint arXiv:2510.06308}, 2025{\natexlab{a}}.
\newblock URL \url{https://arxiv.org/abs/2510.06308}.

\bibitem[Xin et~al.(2025{\natexlab{b}})Xin, Yan, Qin, Li, Liu, Li, Huang, Zhou,
  Zhang, Zhuo, et~al.]{xin2025lumina}
Yi~Xin, Juncheng Yan, Qi~Qin, Zhen Li, Dongyang Liu, Shicheng Li, Victor
  Shea-Jay Huang, Yupeng Zhou, Renrui Zhang, Le~Zhuo, et~al.
\newblock {Lumina-mGPT 2.0: Stand-Alone AutoRegressive Image Modeling}.
\newblock \emph{arXiv preprint arXiv:2507.17801}, 2025{\natexlab{b}}.
\newblock URL \url{https://arxiv.org/abs/2507.17801}.

\bibitem[Xue et~al.(2024)Xue, Shu, Awadalla, Wang, Yan, Purushwalkam, Zhou,
  Prabhu, Dai, Ryoo, et~al.]{xue2024blip3}
Le~Xue, Manli Shu, Anas Awadalla, Jun Wang, An~Yan, Senthil Purushwalkam,
  Honglu Zhou, Viraj Prabhu, Yutong Dai, Michael~S Ryoo, et~al.
\newblock {xGen-MM (BLIP-3): A Family of Open Large Multimodal Models}.
\newblock \emph{arXiv preprint arXiv:2408.08872}, 2024.
\newblock URL \url{https://arxiv.org/abs/2408.08872}.

\bibitem[Yang et~al.(2026)Yang, Lin, Zhao, Xiao, He, Zhao, Deng, Li, Ding, Guo,
  Wang, Zhu, Nie, Zhu, Lin, Li, Huang, Shi, and Fan]{yang2026contextunrolling}
Ceyuan Yang, Zhijie Lin, Yang Zhao, Fei Xiao, Hao He, Qi~Zhao, Chaorui Deng,
  Kunchang Li, Zihan Ding, Yuwei Guo, Fuyun Wang, Fangqi Zhu, Xiaonan Nie,
  Shenhan Zhu, Shanchuan Lin, Hongsheng Li, Weilin Huang, Guang Shi, and Haoqi
  Fan.
\newblock {Context Unrolling in Omni Models}.
\newblock \emph{arXiv preprint arXiv:2604.21921}, 2026.
\newblock \doi{10.48550/arXiv.2604.21921}.
\newblock URL \url{https://arxiv.org/abs/2604.21921}.

\bibitem[Yang et~al.(2025{\natexlab{a}})Yang, Tian, Li, Zhang, Shen, Tong, and
  Wang]{Mmada}
Ling Yang, Ye~Tian, Bowen Li, Xinchen Zhang, Ke~Shen, Yunhai Tong, and Mengdi
  Wang.
\newblock {MMaDA: Multimodal Large Diffusion Language Models}.
\newblock \emph{arXiv preprint arXiv:2505.15809}, 2025{\natexlab{a}}.
\newblock URL \url{https://arxiv.org/abs/2505.15809}.

\bibitem[Yang et~al.(2025{\natexlab{b}})Yang, Kautz, and
  Hatamizadeh]{yang2025gateddelta}
Songlin Yang, Jan Kautz, and Ali Hatamizadeh.
\newblock Gated delta networks: Improving {Mamba2} with delta rule.
\newblock In \emph{International Conference on Learning Representations},
  2025{\natexlab{b}}.
\newblock \doi{10.48550/arXiv.2412.06464}.
\newblock URL \url{https://arxiv.org/abs/2412.06464}.

\bibitem[Ye et~al.(2025)Ye, He, Li, Lin, Yuan, Yan, Hou, and
  Yuan]{ye2025imgedit}
Yang Ye, Xianyi He, Zongjian Li, Bin Lin, Shenghai Yuan, Zhiyuan Yan, Bohan
  Hou, and Li~Yuan.
\newblock {ImgEdit: A Unified Image Editing Dataset and Benchmark}.
\newblock \emph{arXiv preprint arXiv:2505.20275}, 2025.
\newblock \doi{10.48550/arXiv.2505.20275}.
\newblock URL \url{https://arxiv.org/abs/2505.20275}.

\bibitem[You et~al.(2026)You, Zhang, Zhou, Li, and Wen]{you2026lladao}
Zebin You, Xiaolu Zhang, Jun Zhou, Chongxuan Li, and Ji-Rong Wen.
\newblock {LLaDA-o: An Effective and Length-Adaptive Omni Diffusion Model}.
\newblock \emph{arXiv preprint arXiv:2603.01068}, 2026.
\newblock URL \url{https://arxiv.org/abs/2603.01068}.

\bibitem[Yu et~al.(2024)Yu, Yang, Li, Wang, Lin, Liu, Wang, and
  Wang]{yu2024mmvet}
Weihao Yu, Zhengyuan Yang, Linjie Li, Jianfeng Wang, Kevin Lin, Zicheng Liu,
  Xinchao Wang, and Lijuan Wang.
\newblock {MM-Vet: Evaluating Large Multimodal Models for Integrated
  Capabilities}.
\newblock In \emph{Proceedings of the 41st International Conference on Machine
  Learning}, volume 235 of \emph{Proceedings of Machine Learning Research},
  pp.\  57730--57754. PMLR, 2024.
\newblock URL \url{https://proceedings.mlr.press/v235/yu24o.html}.

\bibitem[Yue et~al.(2024)Yue, Ni, Zhang, Zheng, et~al.]{yue2024mmmu}
Xiang Yue, Yuansheng Ni, Kai Zhang, Tianyu Zheng, et~al.
\newblock {MMMU: A Massive Multi-Discipline Multimodal Understanding and
  Reasoning Benchmark for Expert AGI}.
\newblock In \emph{Proceedings of the IEEE/CVF Conference on Computer Vision
  and Pattern Recognition (CVPR)}, 2024.
\newblock URL
  \url{https://openaccess.thecvf.com/content/CVPR2024/html/Yue_MMMU_A_Massive_Multi-discipline_Multimodal_Understanding_and_Reasoning_Benchmark_for_CVPR_2024_paper.html}.

\bibitem[{Z-Image Team} et~al.(2025){Z-Image Team}, Cai, Cao, Du, Gao, Hoi,
  Hou, Huang, Jiang, Jin, Li, et~al.]{cai2025z}
{Z-Image Team}, Huanqia Cai, Sihan Cao, Ruoyi Du, Peng Gao, Steven Hoi, Zhaohui
  Hou, Shijie Huang, Dengyang Jiang, Xin Jin, Liangchen Li, et~al.
\newblock {Z-Image: An Efficient Image Generation Foundation Model with
  Single-Stream Diffusion Transformer}.
\newblock \emph{arXiv preprint arXiv:2511.22699}, 2025.
\newblock URL \url{https://arxiv.org/abs/2511.22699}.

\bibitem[Zhang et~al.(2026)Zhang, Qu, Liu, Chen, Song, Dong, Sun, Li, Wang,
  Jiang, et~al.]{zhang2026nextflow}
Huichao Zhang, Liao Qu, Yiheng Liu, Hang Chen, Yangyang Song, Yongsheng Dong,
  Shikun Sun, Xian Li, Xu~Wang, Yi~Jiang, et~al.
\newblock {NextFlow: Unified Sequential Modeling Activates Multimodal
  Understanding and Generation}.
\newblock \emph{arXiv preprint arXiv:2601.02204}, 2026.
\newblock URL \url{https://arxiv.org/abs/2601.02204}.

\bibitem[Zhao et~al.(2023)Zhao, Gu, Varma, Luo, Huang, Xu, Wright, Shojanazeri,
  Ott, Shleifer, et~al.]{zhao2023fsdp}
Yanli Zhao, Andrew Gu, Rohan Varma, Liang Luo, Chien-Chin Huang, Min Xu, Less
  Wright, Hamid Shojanazeri, Myle Ott, Sam Shleifer, et~al.
\newblock {PyTorch FSDP: Experiences on Scaling Fully Sharded Data Parallel}.
\newblock \emph{arXiv preprint arXiv:2304.11277}, 2023.
\newblock URL \url{https://arxiv.org/abs/2304.11277}.

\bibitem[Zhou et~al.(2024)Zhou, Yu, Babu, Tirumala, Yasunaga, Shamis, Kahn, Ma,
  Zettlemoyer, and Levy]{zhou2024transfusion}
Chunting Zhou, Lili Yu, Arun Babu, Kushal Tirumala, Michihiro Yasunaga, Leonid
  Shamis, Jacob Kahn, Xuezhe Ma, Luke Zettlemoyer, and Omer Levy.
\newblock {Transfusion}: Predict the next token and diffuse images with one
  multi-modal model.
\newblock \emph{arXiv preprint arXiv:2408.11039}, 2024.
\newblock \doi{10.48550/arXiv.2408.11039}.
\newblock URL \url{https://arxiv.org/abs/2408.11039}.

\bibitem[Zhu et~al.(2026)Zhu, Kan, Yang, Chen, Xing, Zhu, Wang, Zhong, and
  Tong]{zhu2026vgau}
Yubo Zhu, Zhehan Kan, Jingyi Yang, Miaolin Chen, Jinbo Xing, Kai Zhu, Zijian
  Wang, Sheng Zhong, and Wei Tong.
\newblock {When Does Visual Generation Help Visual Understanding in Unified
  Multimodal Models?}
\newblock \emph{arXiv preprint arXiv:2608.22174}, 2026.
\newblock \doi{10.48550/arXiv.2608.22174}.
\newblock URL \url{https://arxiv.org/abs/2608.22174}.

\end{thebibliography}
\end{document}